\documentclass{article} 
\usepackage{collas2024_conference,times}
\usepackage{easyReview}

\usepackage{times}
\usepackage{latexsym}
\usepackage{tikz}
\usepackage{pgfplots}
\usepackage{amsmath}
\usepackage{amsfonts}
\usepackage{graphicx}
\usepackage{subcaption}
\usepackage{multirow}
\usepackage{booktabs}
\usepackage{xcolor}
\usepackage{float}
\usepackage{soul}

\pgfplotsset{compat=1.18}

\usepackage{amsmath,amsfonts,bm}

\def\eqref#1{equation~\ref{#1}}

\def\1{\bm{1}}

\DeclareMathAlphabet{\mathsfit}{\encodingdefault}{\sfdefault}{m}{sl}
\SetMathAlphabet{\mathsfit}{bold}{\encodingdefault}{\sfdefault}{bx}{n}

\usepackage{hyperref}
\hypersetup{
    colorlinks=true,
    linkcolor=red,
    filecolor=magenta,
    urlcolor=blue,
    citecolor=purple,
    pdftitle={Overleaf Example},
    pdfpagemode=FullScreen,
    }

\title{Automatic Pruning of Fine-tuning Datasets for Transformer-based Language Models}

\author{Mohammadreza Tayaranian\thanks{Corresponding author: mohammadreza.tayaranian@mail.mcgill.ca}\quad  Seyyed Hasan Mozafari\quad Brett H. Meyer\quad James J. Clark\quad Warren J. Gross
\\
Department of Electrical and Computer Engineering\\
McGill University\\
Montreal, Canada \\
}

\makeatletter
\AtBeginDocument{\let\hl\@firstofone}
\makeatother

\preprintcopy 

\begin{document}

\maketitle

\begin{abstract}
Transformer-based language models have shown state-of-the-art performance on a variety of natural language understanding tasks.
To achieve this performance, these models are first pre-trained on general corpus and then fine-tuned on downstream tasks.
Previous work studied the effect of pruning the training set of the downstream tasks on the performance of the model on its evaluation set.
In this work, we propose an automatic dataset pruning method for the training set of fine-tuning tasks.
Our method is based on the model's success rate in correctly classifying each training data point.
Unlike previous work which relies on user feedback to determine subset size, our method automatically extracts training subsets that are adapted for each pair of model and fine-tuning task.
Our method provides multiple subsets for use in dataset pruning that navigate the trade-off between subset size and evaluation accuracy.
Our largest subset, which we also refer to as the winning ticket subset, is on average $3 \times$ smaller than the original training set of the fine-tuning task.
Our experiments on 5 downstream tasks and 2 language models show that, on average, fine-tuning on the winning ticket subsets results in a $0.1 \%$ increase in the evaluation performance of the model.

\end{abstract}

\section{Introduction}

Transformer-based language models have shown state-of-the-art performance in various natural language understanding tasks \citep{liu2019roberta, t5paper}.
These models are commonly used in a transfer learning setup in which they are first pre-trained on general textual data and then transferred by fine-tuning their parameters on the training set of each downstream task.
The goal of fine-tuning is to maximise the model's performance on the evaluation set\footnote{In this context, the evaluation set may refer to either the validation or the test set of the fine-tuning task.}.
It relies on the existing information in the pre-trained model in addition to the information gained during fine-tuning \citep{durrani-etal-2022-transformation}.
However, different data points in the fine-tuning dataset have different contributions in achieving this goal \citep{katharopoulos2018not}.

As shown in previous work, the contribution of individual fine-tuning data points on the evaluation performance of deep neural networks is not the same \citep{vodrahalli2018all}.
Based on this observation, dataset pruning methods reduce the complexity of fine-tuning by removing data points from the training set of fine-tuning tasks \citep{toneva2018an, sorscher2022beyond, paul2021deep}.
These methods try to minimize the size of the training set of the fine-tuning task while maintaining the evaluation performance of the model after fine-tuning.
They measure the usefulness of each data point in maximising the evaluation performance using scoring systems as a proxy metric.
The pruning is done by removing data points with the lowest score until a certain pruning percentage is reached.

\begin{figure}
    \vspace{-\baselineskip}
    \centering
    \pgfplotsset{width=0.9\columnwidth, height=6cm}
    \begin{subfigure}[b]{0.45\textwidth}
        \centering
        \input{fig/first_page_right}
    \end{subfigure}
    \hfill
    \begin{subfigure}[b]{0.45\textwidth}
        \centering
        \input{fig/firstpage_left}
    \end{subfigure}
    \caption{
    Results of fine-tuning two transformer-based language models, OPT\textsubscript{\textrm{350M}} \citep{zhang2022opt} and RoBERTa\textsubscript{\textrm{LARGE}} \citep{liu2019roberta}, on the winning ticket subset of various downstream tasks.
    For each task, the size of the winning ticket subset as a percentage of the full dataset is shown as gray bars.
    $\Delta$ Metric Performance, shown using red dots, is the change in the evaluation performance of the model which is fine-tuned on the full dataset, compared to the winning ticket subset.
    A positive $\Delta$ indicates that the winning ticket subset improved the evaluation performance of the model.
    }
    \label{fig:overall}
\end{figure}

\cite{ethayarajh2022understanding} studied the fine-tuning datasets of transformer-based language models and reported that fine-tuning tasks have different difficulties for different pre-trained models.
This finding implies that if dataset pruning is applied to the fine-tuning phase, the pruned subset needs to be created based on the needs of each model and fine-tuning task.
Methods of dataset pruning for the fine-tuning of transformer-based language models address this issue by computing dataset pruning scores for each task and model pair \citep{swayamdipta-etal-2020-dataset, fayyaz2022bert}.
However, existing methods do not adapt the size of the pruned subset for different tasks and models.
This results in using an arbitrary subset size for multiple setups.
The problem with this design choice is that a subset size which is optimal for one task may be sub-optimal for another.
A subset with sub-optimal size will either include too many data points and significantly drop the evaluation performance, or too few data points and miss the opportunity to create a smaller subset.
For instance, as discussed in Section \ref{sec:results}, fine-tuning a pre-trained model on a subset which is $27\%$ of the training set for the MNLI task results in almost no drop in the evaluation.
However, using the same model and subset size for the RACE task results in more than $5\%$ drop in the evaluation accuracy.

A simple solution is using hyperparameter search to find the optimal subset size for each task and model.
This results in multiple fine-tuning runs with different sizes which is an overall costly operation considering the model size and the variety of the tasks.

In this work, we aim at pruning the training set of fine-tuning tasks for transformer-based language models.
To this end, we propose a dataset scoring function, $\mathcal{H}$-score, which is based on the output of the model during fine-tuning.
The $\mathcal{H}$-score of each data point is inversely related with its classification difficulty for the model.
Across multiple runs and epochs, a data point with a score of $0$ has never been correctly classified while a data point with the maximum score has always been correctly classified.
The $\mathcal{H}$-score of the $i$-th training data point, $\mathcal{H}_i$, is an integer which essentially distributes data points into a number of buckets.

For dataset pruning, we remove all the data points with a certain $\mathcal{H}$-score to create training subsets from the main training set.
We remove the most difficult data points as including them in the fine-tuning subset only confuses the model and destabilizes the training procedure.
To further reduce the size of the pruned dataset, we also remove the least difficult data points based on the observation that their exclusion from the training set does not affect the model's evaluation performance.
Note that the list of included data points in each subset is not indicated by the user and is based on the $\mathcal{H}$-score.
Given the small size of this subset and that it helps the model achieve the same evaluation performance as the main training set, we call it the \textbf{Winning Ticket Subset}.
In addition to the winning ticket subset, we create smaller subsets by removing more data point buckets with small $\mathcal{H}$-scores, i.e. data points that are harder to learn.
Overall, our proposed subsets provide a diverse set of options for the trade-off between subset size and evaluation performance where small subsets are as small as $4\%$ of the training set of the fine-tuning task.

We conduct experiments to study the effect of fine-tuning transformer-based language models on the subsets that are created with our method.
For each pair of model and task, we first fine-tune the model on the training set with $6$ different seeds and use the training outputs to compute the $\mathcal{H}$-scores.
We then use the $\mathcal{H}$-scores to create our pruned subsets and use these small subsets to fine-tune the pre-trained model.
Figure \ref{fig:overall} illustrates the change in the evaluation performance of the model when fine-tuned on the winning ticket subset compared to full fine-tuning, along with the size of this subset. 
On average, despite containing only $33\%$ of the training data points, fine-tuning on these subsets results in an overall increase of $0.1\%$ in the evaluation performance of the models.
In cases like fine-tuning OPT\textsubscript{\textrm{350M}} on the SST-2 task we even observe a $1.2\%$ increase in the evaluation accuracy with the winning ticket subset being only $21\%$ of the training set.
As shown in Figures \ref{fig:results_roberta} and \ref{fig:results_opt}, our smaller subsets also maintain the evaluation performance.
For instance, fine-tuning RoBERTa\textsubscript{\textrm{LARGE}} on our smallest subset with $4\%$ of training data points results in just $0.8\%$ drop in the evaluation accuracy compared to full fine-tuning.
Our dataset pruning algorithm incurs a one-time cost to achieve a significantly smaller subset.
As a result, the pruned subsets can be used in future runs to reduce fine-tuning time. 
A prominent use-case of this method is neural architecture search (NAS) which requires thousands of GPU hours of fine-tuning on the same model and dataset to find a target architecture \citep{adabert, nasbert}.
Substituting the full fine-tuning dataset with our subsets can significantly reduce the search time of NAS methods while delivering a similar evaluation performance.

The contributions of this work can be summarized as:

\begin{enumerate}
    \item We propose a dataset scoring function, $\mathcal{H}$-score, which categorizes training data points based on the difficulty of the model in classifying them during fine-tuning.

    \item Using $\mathcal{H}$-score, we obtain smaller subsets of the training set of fine-tuning tasks.
    Our subsets are automatically created to minimize subset size while maximising the evaluation performance of the model fine-tuned on them.
    We call the subset obtained by removing the least and the most difficult data points the \textbf{Winning Ticket Subset}.

    \item
    Our experimental results show that fine-tuning transformer-based language models on these small subsets results in an evaluation accuracy similar to using the full training set of the fine-tuning task.
    
\end{enumerate}

\section{Related Work}

Previous work studied the effect of pruning the training datasets commonly used for image classification tasks.
\cite{paul2021deep} proposed a gradient-based GraNd metric along with an estimated version, EL2N, which only uses the training loss to determine the scores of data points.
Using these scores to identify the easy to learn data points of the CIFAR-10 dataset, they prune away up to $40\%$ of the training set with minimal accuracy loss.
\cite{toneva2018an} introduces the \textit{forgetting score} which measures the rate at which a model misclassifies a data point after correctly classifying it.
\hl{
Unlike the \textit{forgetting score} which removes data points that are least forgotten, our proposed $\mathcal{H}$-score negatively rewards misclassifications across any of the fine-tuning epochs which is different from forgetting.
}
\hl{We further study the effects of this design choice in Section} \ref{sec:compare_forgetting}.
\cite{baldock2021deep} measures the dataset difficulty as the number of layers required to consistently classify a data point's label correctly.
They also propose another score, \textit{iteration learned} which is similar to the forgetting score and show that it is correlated with their other layer-based score.
While the original works where designed for training randomly initialized models, \cite{sorscher2022beyond} prunes the training datasets for fine-tuning of pre-trained vision models.
Their experiments show that, when used for fine-tuning, up to $90\%$ of the CIFAR-10 dataset can be removed with no loss of test accuracy.

\cite{fayyaz2022bert} applies the GraNd and EL2N scores to the fine-tuning tasks of transformer-based language models.
\hl{
Their experimental results show the suboptimal performance of these scores when applied to language models.
}
More specifically, they show that random pruning outperforms GraNd and EL2N for subsets smaller than $70\%$ of the training set.
Moreover, pruning more than $60\%$ of the training set using GraNd and EL2N leads to drops in the evaluation accuracy of over $20\%$.
\cite{swayamdipta-etal-2020-dataset} use model's probability of the correct output to determine a dataset pruning score.
Based on this score, they propose to remove the data points that are most ambiguous for the model.
\hl{They also show that their ambiguous setup is more effective than other dataset pruning scores, including the \textit{forgetting score} when applied to the fine-tuning of language models.}
While their idea of fine-tuning on ambiguous data points is similar to our proposed $\mathcal{H}$-score, these methods differ both in the approach and in the achieved results.
\hl{In contrast to} \cite{swayamdipta-etal-2020-dataset} \hl{which uses one fine-tuning run with $20$ epochs, we use the outputs of $6$ runs, each running for $3$ epochs.}
This enables $\mathcal{H}$-score to take the effect of the order of data points into account.
\hl{Moreover, while} \cite{swayamdipta-etal-2020-dataset} \hl{uses the same pruning percentage for all tasks, our method adapts the subset size for each task and model for optimal evaluation performance.}
\hl{In terms of experimental results,} as discussed in Section \ref{sec:results}, our method creates smaller fine-tuning subsets with a smaller drop in the evaluation performance compared to the \textit{ambiguous} setup of \cite{swayamdipta-etal-2020-dataset}.

\section{Methodology}

\subsection{Dataset Scoring System: $\mathcal{H}$-score}
\label{sec:hscore}

Consider the training set of a fine-tuning task, $\mathcal{D}=\{(x_i,y_i)\}^{N}_{i=1}$.
The impact of each $(x_i,y_i)$ on the evaluation performance of the model varies based on different attributes \citep{vodrahalli2018all, katharopoulos2018not}.
We define a dataset scoring function that quantifies this impact based on the ability of the model in predicting $y_i$ during fine-tuning.

Given a classification problem with $M$ classes and $z_i \in \mathbb{R}^M$ as the model's logits for $x_i$, the model's prediction of the true label is calculated as $\hat{y}_i = \mathrm{argmax}_m (P(z_{im}))$.
In this context, $P(z_{im})$ is model's predicted probability for class $m$ given by the softmax function.
Fine-tuning the model for $S$ runs with different initializations with each run taking $E$ epochs, we store all the predictions as $\hat{Y} \in \mathbb{Z}^{N \times S \times E}$.
We define our proposed dataset scoring function as $f: \hat{Y} \times Y \rightarrow \mathcal{H}$ where $\mathcal{H} \in \mathbb{Z}^{N}$ is the list of scores for the data points.
For $i \in \{1, 2, \ldots, N\}$, the $\mathcal{H}$-score for data point $(x_i, y_i)$, is calculated as
\begin{equation}
    \mathcal{H}_i = \sum_{j=1}^{S} \prod_{k=1}^{E} [\hat{Y}_{ijk} = y_i]
    \label{eq:hscore}
\end{equation}
where $[ \cdot ]$ is the Iverson bracket.

For each data point, Equation \ref{eq:hscore} assigns a reward of $1$ for each epoch and fine-tuning run if the model has correctly classified the true label.
In the next step it multiplies this reward over all the epochs of each run to penalize data points that are misclassified at any epoch during at least one fine-tuning run.
The resulting values, that are either $0$ or $1$, are summed for each run, resulting in $\mathcal{H}$-scores such that $\mathcal{H}_i \in \{0, 1, \ldots, S\}$.

\begin{figure}
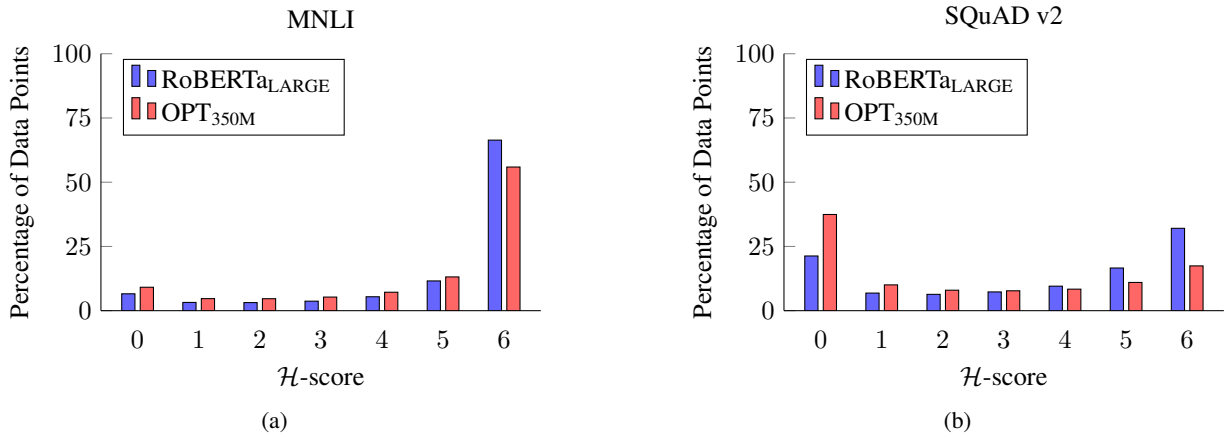

  \centering
  \begin{subfigure}{0.45\textwidth}
    \centering
    \pgfplotsset{width=\columnwidth, height=5cm}
    \input{fig/score_hist/mnli}
    \caption{}
    \label{fig:mnli_hist}
  \end{subfigure}
  \hfill
  \begin{subfigure}{0.45\textwidth}
    \centering
    \pgfplotsset{width=\columnwidth, height=5cm}
    \input{fig/score_hist/squad}
    \caption{}
    \label{fig:squad_hist}
  \end{subfigure}
    \caption{
    Distribution of the $\mathcal{H}$-score for the training set of the MNLI and SQuAD v2 tasks, calculated using two transformer-based models RoBERTa\textsubscript{\textrm{LARGE}} and OPT\textsubscript{\textrm{350M}}.
    }
  \label{fig:score_hist}
\end{figure}

Here we discuss the properties of a given data point, $(x_i, y_i)$, based on its $\mathcal{H}$-score:
\begin{itemize}
    \item $\mathcal{H}_i =0$: Having a $\mathcal{H}$-score of 0 \hl{indicates that in all of the runs this data point experienced at least one epoch where it was incorrectly classified.}

    \item $\mathcal{H}_i =S$: This data point has been correctly classified and received a reward of 1 in all the epochs of all the fine-tuning runs which implies that correctly classifying it was relatively easier for the model.

    \item $\mathcal{H}_i \in\{{1,2,\ldots, S-1}\}$:
    This data point is more ambiguous for the model than previous cases as it was correctly classified in all epochs of some, but not all of the runs.
    More specifically, the farther the $\mathcal{H}$-score of a data point is from $0$ or $S$, the more ambiguous it is for the model to classify.
\end{itemize}

Figure \ref{fig:score_hist} shows the distribution of the $\mathcal{H}$-scores for the MNLI and SQuAD v2 fine-tuning datasets in a setup where $S = 6$.
For the $\mathcal{H}$-score distrbution of other tasks refer to Appendix \ref{sec:app:dist}.

\subsection{Dataset Pruning With $\mathcal{H}$-score}
\label{sec:pruning}

Given $\mathcal{D}=\{(x_i,y_i)\}^{N}_{i=1}$ the training set of a fine-tuning task, and a set of $\mathcal{H}$-score values, $M = \{m \mid m \in \mathbb{Z}, 0 \leq m \leq S \}$, we define a training subset $\mathcal{D}_M$ as
\begin{equation}
    \mathcal{D}_M = \{(x_i, y_i) \mid \mathcal{H}_i \in M, 1 \leq i \leq N \} .
    \label{eq:subset}
\end{equation}
For instance, $\mathcal{D}_{\{0, 2, 5\}}$ is a subset of the training set which contains all the data points that have a $\mathcal{H}$-score of either $0$, $2$, or $5$.

In order to be used for dataset pruning, a training subset needs to have a smaller size than the training set.
Additionally, the evaluation performance of the model that is fine-tuned on this subset needs to be close to the one fine-tuned on the full dataset.
Previous work fine-tuned pre-trained models on a subset that is created by removing hard-to-learn and easy-to-learn data points from the training set \citep{swayamdipta-etal-2020-dataset, ethayarajh2022understanding}.
As discussed in Section \ref{sec:hscore}, the $\mathcal{H}$-score of a data point has an inverse relationship with its diffculty for the model.
Based on these observations, we propose to use the following subsets for dataset pruning:
\begin{itemize}
    \item $\mathcal{D}_{\{1,2,3,4,5\}}$: This subset, which we also refer to as the \textbf{Winning Ticket Subset}, is created by removing all data points $(x_i, y_i)$ where $\mathcal{H}_i=0$ and $\mathcal{H}_i=6$ which correspond to the most and least difficult training data points, respectively.
    The former group is removed as their $\mathcal{H}$-score of 0 means that the model is unlikely to learn them and their presence in the training subset only confuses the model.
    On the other hand, data points with $\mathcal{H}_i = 6$ are too easy to learn and can be avoided without damaging the evaluation performance.
    As shown in Figures \ref{fig:score_hist} and \ref{fig:more_hist}, the majority of data points are associated with either of these $\mathcal{H}$-scores.

    \item $\mathcal{D}_{\{2,3,4,5\}}$, $\mathcal{D}_{\{3,4,5\}}$, $\mathcal{D}_{\{4,5\}}$, and $\mathcal{D}_{\{5\}}$: These subsets enable pruning to even smaller percentages than the winning ticket subset.
    To create these subsets we start from the winning ticket subset and each time remove the data points with the smallest $\mathcal{H}$-score.
    Although the removed data points are not as difficult as those with $\mathcal{H}_i = 0$, they are the most difficult ones in each subset.

    \item $\mathcal{D}_{\{4\}}$ and $\mathcal{D}_{\{2,3,4\}}$:
    As discussed in Section \ref{sec:hscore}, data points that have $\mathcal{H}$-scores that are farthest from $0$ and $6$ are the most ambiguous for the model.
    These two subsets are also dominated by data points with middle scores and thus represent the most ambiguous subsets.
\end{itemize}

Although the data points of each subset are determined based on the $\mathcal{H}$-score, the variety of subsets helps the user decide on the trade-off between subset size and evaluation performance.
The fact that the choice of data points and the subset size is determined by our method ensures that each subset includes the data points that maximise the evaluation performance at each size.
Our dataset pruning method has the advantage of determining the subset size along with the scores for the data points while previous work requires more fine-tuning runs to search for the optimal subset size after the score computation phase.
\hl{We use the model's output during the fine-tuning runs of the score computation phase to determine pruned subsets of the dataset.
After creation, the pruned subsets can be used for faster fine-tuning runs.}

\subsection{Empirical Analysis of Fine-tuning on Training Subsets}
\label{sec:subset}

In this section, we study the effect of fine-tuning pre-trained models on subsets created using Equation \ref{eq:subset}.
We first compute the $\mathcal{H}$-scores of the training set of the fine-tuning task by fine-tuning it with 6 different seeds.
We then create all the possible subsets, $\mathcal{D}_M$, and study the effects of using data points with certain $\mathcal{H}$-scores on the evaluation performance of the model.
Here we use the MNLI fine-tuning task with the OPT\textsubscript{\textrm{350M}} model.
Further examples of other models and tasks are available in Appendix \ref{sec:app:subset}.

\begin{figure}
  \centering
\pgfplotsset{width=\textwidth, height=8cm}
\input{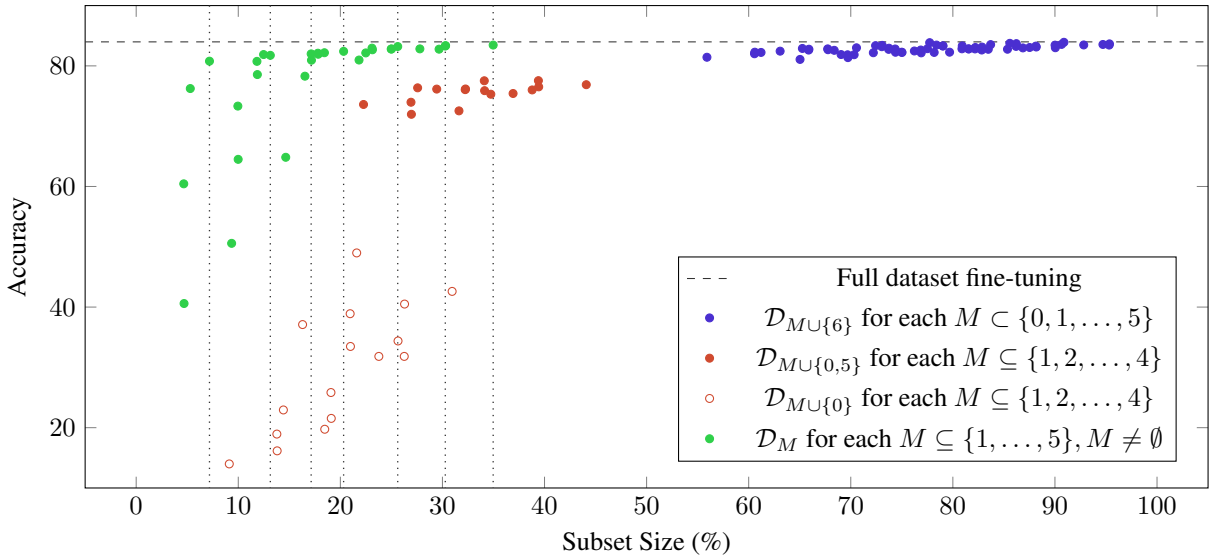}
\caption{
Subset size and evaluation accuracy of different subsets of the MNLI training set based on OPT\textsubscript{\textrm{350M}}.
\hl{Each dot represents a subset.}
Subsets are created using Equation \ref{eq:subset} with all possible values of $M$.
Our proposed subsets are noted with vertical dotted lines.
}
\label{fig:mnli_ablation}
\end{figure}

Figure \ref{fig:mnli_ablation} summarizes the results of these fine-tuning runs.
It plots the accuracy of each run and the size of its training subset shown as a percentage of the full dataset.
Based on these results, we divide the subsets into four groups:

\begin{enumerate}
    \item The first group, shown with blue dots, is the set of training subsets, $\mathcal{D}_{M \cup \{6\} }$, \hl{for all} $M \subset \{ 0,1, \ldots, 5 \}$.
    In addition to other data points, subsets in this group always contain all of the data points with  $\mathcal{H}_i = 6$.
    Given the number of data points with $\mathcal{H}_i = 6$ (Figure \ref{fig:mnli_hist}), the size of these subsets is close to the full dataset.
    They also show consistently small accuracy drops compared to the full fine-tuning.

    \item
    Distinguished with red dots, this group contains the set of training subsets, $\mathcal{D}_{M \cup \{0,5\} }$, \hl{for all $M \subseteq \{ 1,2, \ldots, 4 \}$}.
    Subsets in this group do not include data points with $\mathcal{H}_i = 6$, but contain the ones with $\mathcal{H}_i$ equal to $0$ and $5$.
    The drop in accuracy between the subsets in this group and previous group arises from excluding easy to learn data points, e.g. $\mathcal{H}_i = 6$, while keeping hard to learn ones.
    
    \item
    These are the same subsets from the previous group where their data points with $\mathcal{H}_i = 5$ have also been removed.
    Shown with red circles, subsets in this group have a considerably lower accuracy compared to the ones in group 2.
    This accuracy drop indicates that data points with $\mathcal{H}_i = 5$ have an important role in stabilizing the training and maintaining the evaluation performance.

    \item Subsets in this group, shown with green dots, do not contain any data point that has a $\mathcal{H}$-score of either $0$ or $6$.
    These subsets generally benefit from the high accuracy that's observed in group 1, while also being significantly smaller in terms of subset size.
    The few exceptional subsets of this group, with accuracies below $70\%$, are dominated by data points with a small $\mathcal{H}$-score.
    \label{group4}
    
\end{enumerate}

Throughout our experiments, subsets that included data points with larger $\mathcal{H}$-scores better maintained the evaluation accuracy while data points with smaller $\mathcal{H}$-scores negatively affected the evaluation accuracy of the model.
To illustrate this point, consider the subset $\mathcal{D}_{\{1,2,3,4\}}$.
The evaluation accuracy of the model fine-tuned on this subset is $81\%$.
We then create  $\mathcal{D}_{\{0,1,2,3,4\}}$ by adding all the data points with $\mathcal{H}_i = 0$ to this subset.
Despite having more data points than the former, the evaluation accuracy of the latter subset drops to $43\%$.
On the other hand, fine-tuning on $\mathcal{D}_{\{1,2,3,4,5\}}$ which contains data points with $\mathcal{H}_i = 5$ increases the evaluation accuracy to $83\%$.
This observation is aligned with our decision to remove data points with smaller scores from our dataset pruning subsets.

Indicated with vertical dotted lines, our proposed dataset pruning subsets are found among the green dots of group \ref{group4}.
Since subsets of group \ref{group4} do not contain data points with a score of $6$, our subsets tend to have a smaller size than subsets with similar accuracies.
Also, our subsets have a better accuracy compared to other subsets of similar sizes.
We observe a similar trend when plotting the evaluation performance for subsets of different fine-tuning tasks and models (Appendix \ref{sec:app:subset}).

\section{Dataset Pruning Experiments}
\label{sec:experiments}

\subsection{Experimental Setup}
\label{sec:setup}

We apply our dataset pruning method on the training set of various fine-tuning tasks and use our proposed subsets to fine-tune pre-trained transformer-based language models.
We use OPT\textsubscript{\textrm{350M}} and RoBERTa\textsubscript{\textrm{LARGE}} for our experiments.
\hl{We add a task-specific head with randomly initialized parameters to the model with pre-trained parameters.}
In these experiments, we first create the pruned subsets for each pair of task and model and then use them for fine-tuning.
The reported evaluation results are the average of 3 fine-tuning runs with different initialization seeds.
\hl{The initialization seed controls the order of the data points and the initial parameters of the task-specific head.}
More implementation details and hyper-parameters are included in Appendix \ref{sec:app:setup}.

To obtain our subsets, we fine-tune each of the models on each task using 6 different initialization seeds and use the outputs to compute the $\mathcal{H}$-scores using Equation \ref{eq:hscore}.
The $\mathcal{H}$-scores are then used to extract the training subsets as described in Section \ref{sec:pruning}.

In addition to our subsets, we fine-tune the models on subsets that are obtained with a method similar to \cite{swayamdipta-etal-2020-dataset}.
We first use the outputs of the same 6 fine-tuning runs to calculate their \textit{variability} metric.
Following the \textit{ambiguous} setup of \cite{swayamdipta-etal-2020-dataset} we create subsets by picking data points with the largest \textit{variability} value.
We create the same number of \textit{ambiguous} subsets as ours.
We set the subset sizes such that for each \textit{ambiguous} subset there exists a subset of the same size created with our method.
The \textit{ambiguous} setup described here differs from the original paper because we use 6 runs, instead of 1 run, to calculate the \textit{variability} score.
Similarly, we create the same number of randomly pruned subsets with each having the same number of data points as the \textit{ambiguous} ones.
\hl{To create a random subset of size $x$ we shuffle the dataset and pick the first $x$ data points.}

We measure the evaluation metric performance of the models when fine-tuned on the pruned subsets along with the full training set.
We run our fine-tuning experiments on five downstream tasks that target different natural language understanding abilities:
\begin{itemize}
    \item \textbf{SNLI} and \textbf{MNLI}: The goal of these natural language inference tasks is to determine whether a given sentence entails, contradicts, or is neutral with respect to another sentence. Compared to SNLI \citep{snlipaper}, MNLI is considered more challenging as it uses sentences from more diverse backgrounds \citep{mnlipaper}.

    \item \textbf{SST-2}: The goal of this task is to classify whether each sentence has a negative or a positive sentiment \citep{sst2paper}.

    \item \textbf{SQuAD v2} and \textbf{RACE}: Created for reading comprehension, the task in SQuAD v2 is to find the span of the answer of a question in the given passage.
    The second version of this dataset contains unanswerable questions which make it more challenging \citep{squadpaper}.
    RACE is another reading comprehension task which is made up of articles and multiple choice questions associated with them.
    The goal of the task is to find the correct answer choice for each question.
    The length of the articles and the fact that the questions are from real-world exams make RACE a more challenging reading comprehension task \citep{racepaper}.
\end{itemize}

Except for RACE, the evaluation performance is measured using the publicly available validation sets of each task.
For the RACE task, we use the publicly available test set to measure the evaluation performance.

\begin{figure} [t]
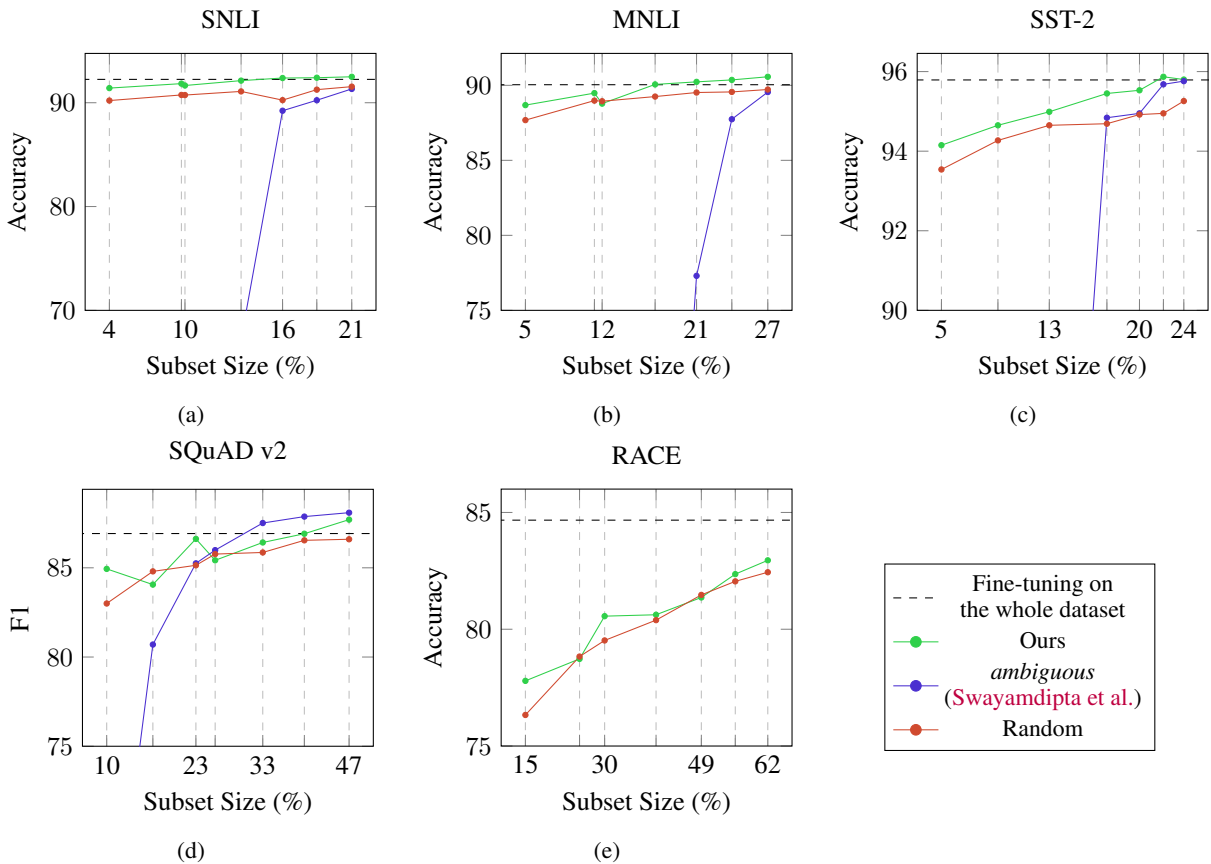

  \centering
  \begin{subfigure}{0.33\textwidth}
    \centering
    \pgfplotsset{width=\columnwidth, height=5cm}
    \input{fig/results/snli_roberta_large}
    \caption{}
    \label{fig:result_snli_roberta}
  \end{subfigure}
  \hfill
  \begin{subfigure}{0.33\textwidth}
    \centering
    \pgfplotsset{width=\columnwidth, height=5cm}
    \input{fig/results/mnli_roberta_large}
    \caption{}
  \end{subfigure}
  \hfill
  \begin{subfigure}{0.33\textwidth}
    \centering
    \pgfplotsset{width=\columnwidth, height=5cm}
    \input{fig/results/sst2_roberta_large}
    \caption{}
  \end{subfigure}
  \vspace{1em}
    \centering
  \begin{subfigure}{0.33\textwidth}
    \centering
    \pgfplotsset{width=\columnwidth, height=5cm}
    \input{fig/results/squad_roberta_large}
    \caption{}
    \label{fig:squad_roberta}
  \end{subfigure}
  \hfill
  \begin{subfigure}{0.33\textwidth}
    \centering
    \pgfplotsset{width=\columnwidth, height=5cm}
    \input{fig/results/race_roberta_large}
    \caption{}
    \label{fig:result_race_roberta}
  \end{subfigure}
  \hfill
  \begin{subfigure}{0.33\textwidth}
    \centering
    \pgfplotsset{width=\columnwidth, height=6cm}
    \input{fig/results/legend}
  \end{subfigure}
    \caption{
    Evaluation accuracy of fine-tuning RoBERTa\textsubscript{\textrm{LARGE}} on different subsets of the training dataset of multiple downstream tasks.
    All the plots share the same provided legend.
    \hl{In }\ref{fig:result_race_roberta}\hl{ the \textit{ambiguous} setup falls outside the range of the vertical axis. Detailed accuracies and subset sizes are provided in Appendix }\ref{sec:app:results}.
    }
  \label{fig:results_roberta}
\end{figure}

To compute the $\mathcal{H}$-score for SQuAD v2 we consider the output as a correct classification if the span of the answer in the passage is exactly the same as the label.

\subsection{Results and Discussion}
\label{sec:results}

\begin{figure} [t]
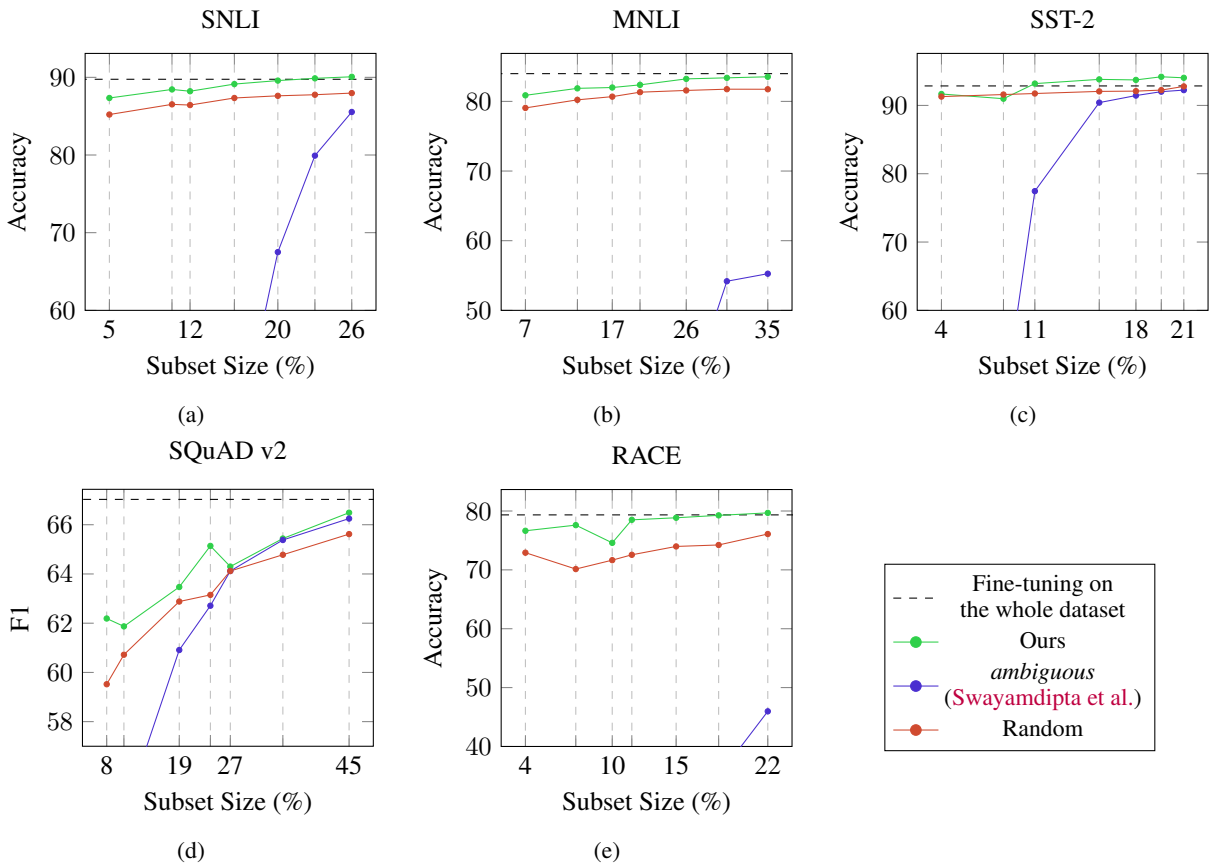

  \centering
  \begin{subfigure}{0.33\textwidth}
    \centering
    \pgfplotsset{width=\columnwidth, height=5cm}
    \input{fig/results/snli_opt}
    \caption{}
  \end{subfigure}
  \hfill
  \begin{subfigure}{0.33\textwidth}
    \centering
    \pgfplotsset{width=\columnwidth, height=5cm}
    \input{fig/results/mnli_opt}
    \caption{}
    \label{fig:result_mnli_opt}
  \end{subfigure}
  \hfill
  \begin{subfigure}{0.33\textwidth}
    \centering
    \pgfplotsset{width=\columnwidth, height=5cm}
    \input{fig/results/sst2_opt}
    \caption{}
  \end{subfigure}
  \vspace{1em}
    \centering
  \begin{subfigure}{0.33\textwidth}
    \centering
    \pgfplotsset{width=\columnwidth, height=5cm}
    \input{fig/results/squad_opt}
    \caption{}
  \end{subfigure}
  \hfill
  \begin{subfigure}{0.33\textwidth}
    \centering
    \pgfplotsset{width=\columnwidth, height=5cm}
    \input{fig/results/race_opt}
    \caption{}
    \label{fig:result_race_opt}
  \end{subfigure}
  \hfill
  \begin{subfigure}{0.33\textwidth}
    \centering
    \pgfplotsset{width=\columnwidth, height=6cm}
    \input{fig/results/legend}
  \end{subfigure}
    \caption{
    Evaluation accuracy of fine-tuning OPT\textsubscript{\textrm{350M}} on different subsets of the training dataset of multiple downstream tasks.
    All the plots share the same provided legend.
    \hl{Detailed accuracies and subset sizes are provided in Appendix }\ref{sec:app:results}.
    }
  \label{fig:results_opt}
\end{figure}

Figures \ref{fig:results_roberta} and \ref{fig:results_opt} illustrate the results of our experiments on RoBERTa\textsubscript{\textrm{LARGE}} and OPT\textsubscript{\textrm{350M}}, respectively.
The evaluation metric performance is shown as  F1 score for SQuAD v2 and as accuracy for other tasks.
For each task, we plot the evaluation performance against the size of the training subset for different dataset pruning setups.
We use vertical dashed lines to indicate each subset size and a horizontal dashed line to indicate the evaluation performance of full fine-tuning.
The range of values in the vertical axis are limited for better visibility which resulted in some points falling outside the range, e.g. \textit{ambiguous} subsets in Figure \ref{fig:result_race_roberta}.
The values for accuracies and subset sizes are provided in Appendix \ref{sec:app:results}.

Our subsets contain significantly fewer data points than the training set of each task.
Despite their small size, fine-tuning the models on these subsets results in a similar evaluation performance compared to full fine-tuning.
The evaluation performance of our subsets shows that removing the hard-to-learn data points from the training set of the fine-tuning task does not negatively affect the model's performance and in some cases increases the evaluation performance.

The number of data points in each subset, specified by our method, varies across different tasks and models.
Our method tries to minimize the subset size for each case while achieving the highest evaluation performance.
As different models have different behavior on the same task, using the same subset size for different cases results in suboptimal performance.
For instance, despite having the same task, subset sizes of Figures \ref{fig:result_race_roberta} and \ref{fig:result_race_opt} greatly vary, showing that subset sizes from one model cannot be used for the other one.
Our results also demonstrate the ability of our method in adapting the subset size for each model based on its properties.

In most cases, our subsets result in a higher evaluation performance than the ones created using the \textit{ambiguous} method.
This trend is reversed for the largest subsets in Figure \ref{fig:squad_roberta} where our subsets have a slightly lower performance.
We attribute this to the fact that our method negatively penalizes a slightly incorrect answer span which can result to unnecessary removal of data points.
Overall, the low evaluation performance of the large \textit{ambiguous} subsets in Figures \ref{fig:result_race_roberta}, \ref{fig:result_mnli_opt}, and \ref{fig:result_race_opt} along with their sudden performance drops for other tasks is indicative of the low reliability of the \textit{ambiguous} method.

\begin{figure} [t]
  \centering
    \begin{subfigure}{0.33\textwidth}
    \centering
    \pgfplotsset{width=\columnwidth, height=5cm}
    \input{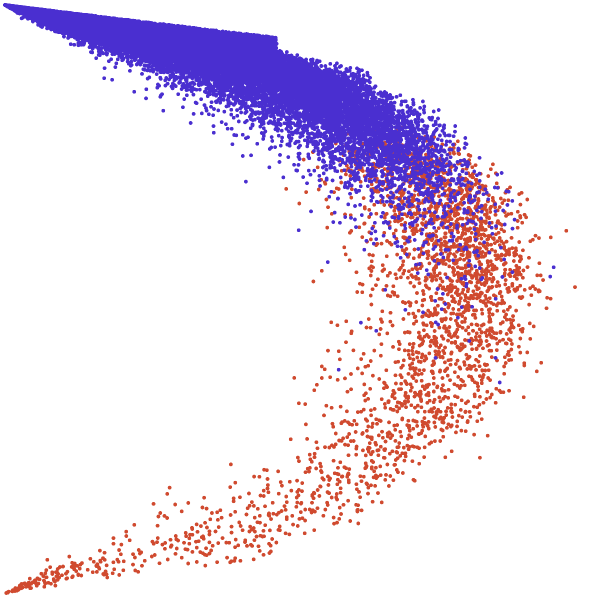}
    \caption{}
    \label{fig:data_map}
  \end{subfigure}
  \hfill
  \begin{subfigure}{0.66\textwidth}
    \centering
    \pgfplotsset{width=0.5\columnwidth, height=5cm}
    \input{fig/mean_score/small_opt}
    \input{fig/mean_score/legend2}
    \caption{}
    \label{fig:meanh}
  \end{subfigure}
    \caption{
    (a) Data map for the data points of the SST-2 train set, based on OPT\textsubscript{\textrm{350M}}. Confidence and variability metrics are defined in \cite{swayamdipta-etal-2020-dataset}.
    Red dots represent data points with $\mathcal{H}_i \leq 1$ while other $\mathcal{H}$-scores are shown as blue dots.
    (b) Mean $\mathcal{H}$-score of training subsets along with the training set for MNLI, SST-2, and SNLI tasks, based on OPT\textsubscript{\textrm{350M}}.
    }
\end{figure}

Figure \ref{fig:data_map} depicts the data map for the training set of the SST-2 task where data points with $\mathcal{H}_i \leq 1$ are distinguished with red dots.
As the \textit{ambiguous} subsets include data points with the largest \textit{variability} value, they tend to be overpopulated with data points where $\mathcal{H}_i \leq 1$.
Figure \ref{fig:meanh} depicts the mean $\mathcal{H}$-score of the data points that are included in each \textit{ambiguous} subset.
A small mean $\mathcal{H}$-score indicates that the subset is dominated by data points with a small $\mathcal{H}$-score.
Given the destabilizing effect of these points, discussed in Section \ref{sec:subset}, their high numbers in a subset leads to low evaluation accuracy.
As the mean $\mathcal{H}$-score increases, the evaluation accuracy of the \textit{ambiguous} subsets in Figure \ref{fig:results_opt} increases as well.
For instance, mean $\mathcal{H}$-score for SST-2 is increased with a higher rate than the MNLI.
This rate difference is also reflected in their respective evaluation accuracies where the evaluation accuracy of smaller SST-2 subsets are closer to their baseline than MNLI subsets.
Data maps for other tasks and models, and the mean $\mathcal{H}$-score plot based on RoBERTa\textsubscript{\textrm{LARGE}} are provided in Appendix \ref{sec:app:datamap}.

\hl{Across tasks and models, randomly created subsets closely follow our subsets in terms of evaluation performance.}
Considering that we utilize a uniform random function, the data points in these subsets have the same $\mathcal{H}$-score distribution, and thus the same mean $\mathcal{H}$-score, as the training set which helps them avoid bad evaluation performances.
Despite this, random subsets are outperformed by our subsets in most cases which highlights the importance of excluding hard to learn data points from training.
For instance, fine-tuning OPT\textsubscript{\textrm{350M}} on the random subsets of the SNLI and RACE tasks results in an average drop in evaluation performance of $2\%$ and $5\%$, respectively.

\subsection{Comparison with Forgetting Score}
\label{sec:compare_forgetting}

To compute the $\mathcal{H}$-score, Equation \ref{eq:hscore} multiplies the reward of each data point across epochs and then adds the resulting value for multiple fine-tuning runs.
As a result of the multiplication, only runs in which the data point is correctly classified in all epochs receive a reward of $1$.
More specifically, if a data point is learned in the middle of the run, i.e. the second epoch, and is never forgotten, it doesn't receive a reward of $1$.
This is in contrast with the \textit{forgetting score} which rewards such data points and only assigns $0$ rewards to forgetting events \citep{toneva2018an}.
\hl{We create $\mathcal{F}$-score to study the effect of the mentioned difference between $\mathcal{H}$-score and forgetting score.}
$\mathcal{F}$-score assigns a reward of $1$ if a data point is learned either in the beginning or in the middle of a run, and not forgotten until the end.

Following Section \ref{sec:pruning}, we use $\mathcal{F}$-score to create subsets of the fine-tuning dataset and use them to fine-tune OPT\textsubscript{\textrm{350M}} and RoBERTa\textsubscript{\textrm{LARGE}}.
Figure \ref{fig:fscore_hist} of Appendix \ref{sec:app:compare_forgetting} provides a comparison between the distribution of scores for both methods.
It shows a distribution shift in $\mathcal{F}$-score subsets which results in more data points falling in the maximum score, leading to the pruning of more data points.
As shown in Tables \ref{tab:fscore_mnli} and \ref{tab:fscore_race}, the removed data points lead to an overall smaller size for subsets based on the $\mathcal{F}$-score.
However, fine-tuning the models on these subsets leads to significant drops in the evaluation performance compared to $\mathcal{H}$-score subsets.
As a result, \hl{using $\mathcal{F}$-score}, and thus following \citet{toneva2018an}, \hl{results in subsets with a lower evaluation performance compared to $\mathcal{H}$-score.}

\subsection{Sensitivity Analysis on $E$ and $S$ Hyperparameters}
\label{sec:ablation_nbruns}

The number of fine-tuning runs that are used to compute the $\mathcal{H}$-score is shown as $S$ in Equation \ref{eq:hscore}.
We study the sensitivity of our dataset pruning method to the value of $S$.
We create winning ticket subsets with $\mathcal{H}$-scores for all possible values of $S \in \{2,3,4,5,6\}$.
As shown in Table \ref{tab:ablation_s} of Appendix \ref{sec:app:ablation_nbruns}, \hl{both the winning ticket subset size and the evaluation performance of the model drops when $S$ is decreased.}
Compared to subsets of similar size with $S = 6$, winning ticket subsets with $S < 6$ sometimes have a slightly higher evaluation performance, e.g. RACE on RoBERTa\textsubscript{\textrm{LARGE}} (Tables \ref{tab:results1} to \ref{tab:results3}).
However, as $\mathcal{H}_i \in \{0, 1, \ldots, S\}$, \hl{a larger $S$ leads to a higher granularity of the $\mathcal{H}$-score and more options for subsets.}
Moreover, extreme cases like $S = 2$ lack sufficient information about the effect of the initialization seed due to the limited number of outputs used for score computation.
A similar trend is observed in the case of $E$, i.e. the number of epochs used to compute the $\mathcal{H}$-score.
Tables \ref{tab:ablation_e1} and \ref{tab:ablation_e2} of Appendix \ref{sec:app:ablation_nbruns} show that \hl{decreasing the hyperparameter $E$ from $3$ to $1$ has an overall negative effect on the evaluation performance of the model.}

Our sensitivity analysis experiments provide information for the trade-off between the overhead of computing the $\mathcal{H}$-score and the evaluation performance of the created subsets.
Although reducing the number of epochs and fine-tuning runs mitigates the overhead of dataset pruning, it can also result in subsets with a worse evaluation performance.

\section{Conclusion and Future Work}

We presented a novel method to prune the training set of fine-tuning tasks of the transformer-based language models.
We propose a dataset difficulty score, $\mathcal{H}$-score, which is based on the predictions of the model for each training data point.
Using the $\mathcal{H}$-score, we create subsets of the training set that are used for fine-tuning the model.
Unlike other dataset pruning methods, our method creates subsets that are optimized for each model.
Such design enables automatic dataset pruning and avoids the costly hyper-parameter search.
Data points in each subset are selected to minimize subset size and maximise the evaluation performance of the model that is fine-tuned on them.
Our experiments show that our largest subset, the winning ticket subset, is on average $33\%$ of the training set but results in a $0.1\%$ increase in the evaluation performance of the fine-tuned model.
\hl{Given their evaluation performance, the pruned subsets can replace the full fine-tuning dataset and save time and resources in applications where multiple fine-tuning runs are needed, e.g. neural architecture search.}

For future work, we will apply our method to a wider variety of natural language understanding tasks.
Specifically, we will focus on question answering tasks similar to SQuAD.
In addition to that, the effect of fine-tuning on the pruned subset on the evaluation performance of the model on adversarial evaluation sets is a topic of interest which is left for future work.

\section{Limitations}

Through removing parts of the training set, our method essentially limits the information that is fed to a pre-trained model during fine-tuning.
Our experiments have shown that the model is able to maintain its performance on the validation or the test set associated with each task.
However, these datasets may not be capable to fully capture the model's ability in learning the fine-tuning task.
As a result, the performance of our fine-tuned models on adversarial tasks can be lower than full fine-tuning.
This topic is left for future work.

Although our method is effective in achieving near baseline evaluation performance, the size of the winning ticket subset varies across different models.
As a result, the size of the pruned subsets, and thus the speedup associated with it, is dependant on the model and the task.
We leave the topic of studying the effectiveness of our proposed method on more models and tasks to future work.

\subsubsection*{Acknowledgments}

We would like to thank our colleagues Amir Ardakani and Hugo Tessier, and the anonymous reviewers for their helpful comments and discussions.
This work was supported by Huawei Technologies Canada.

\bibliography{collas2024_conference}
\bibliographystyle{collas2024_conference}

\appendix
\section{Appendix}

\subsection{$\mathcal{H}$-score Distribution for Other Tasks}
\label{sec:app:dist}

\begin{figure}[h]
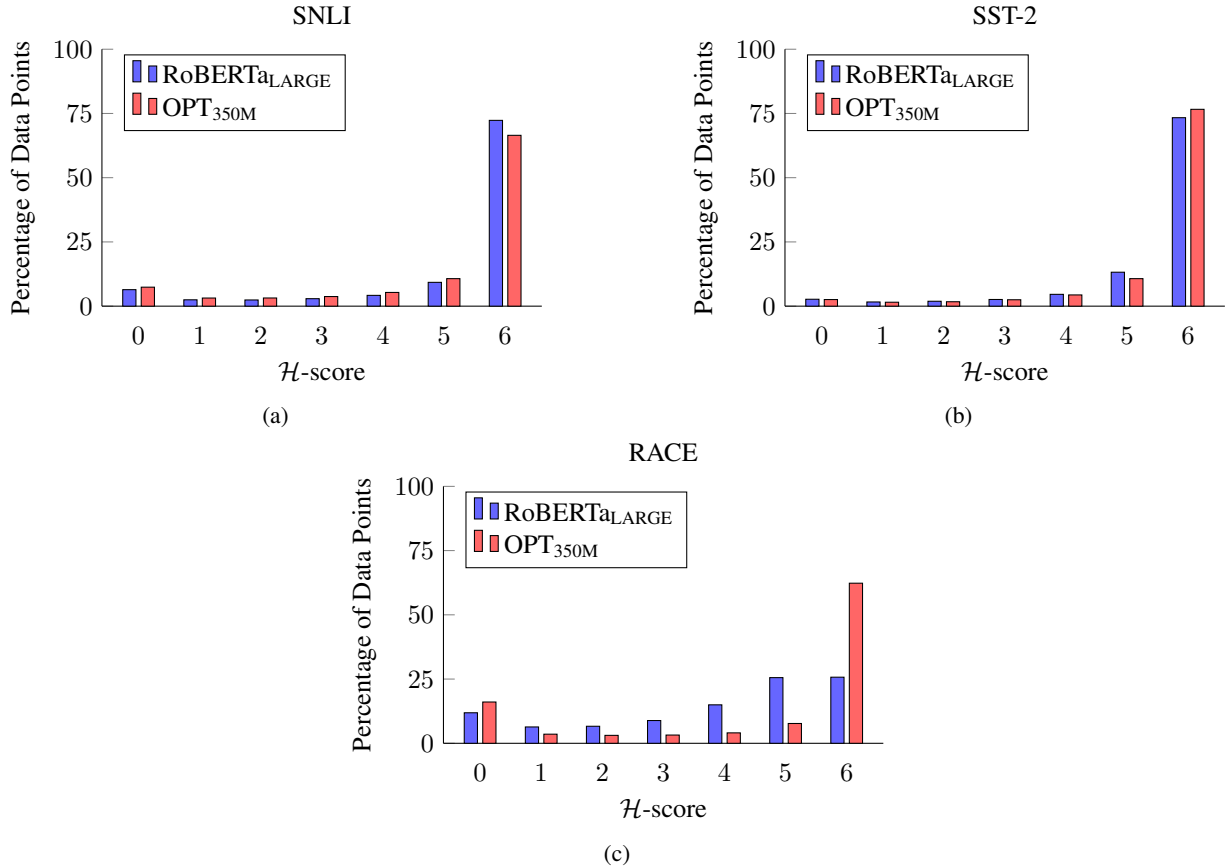

  \centering
  \begin{subfigure}{0.45\textwidth}
    \centering
    \pgfplotsset{width=\columnwidth, height=5cm}
    \input{fig/score_hist/snli}
    \caption{}
    \label{fig:snli_hist}
  \end{subfigure}
  \hfill
  \begin{subfigure}{0.45\textwidth}
    \centering
    \pgfplotsset{width=\columnwidth, height=5cm}
    \input{fig/score_hist/sst2}
    \caption{}
    \label{fig:sst2_hist}
  \end{subfigure}
  \vfill
    \begin{subfigure}{0.45\textwidth}
    \centering
    \pgfplotsset{width=\columnwidth, height=5cm}
    \input{fig/score_hist/race}
    \caption{}
    \label{fig:race_hist}
  \end{subfigure}
    \caption{
    Distribution of the $\mathcal{H}$-score for the training set of the SNLI, SST-2, and RACE tasks, based on RoBERTa\textsubscript{\textrm{LARGE}} and OPT\textsubscript{\textrm{350M}}.
    }
  \label{fig:more_hist}
\end{figure}

The $\mathcal{H}$-score distributions of the SNLI, SST-2, and RACE tasks are provided in Figure \ref{fig:more_hist}.

\subsection{Subsets for other tasks and models}
\label{sec:app:subset}

Figures \ref{fig:mnli_ablation_large}, \ref{fig:mnli_ablation_base}, \ref{fig:snli_ablation_base}, \ref{fig:sst2_ablation_base}, and \ref{fig:squadv2_ablation_large} plot the evaluation accuracy against the subset size for all possible subsets created using Equation \ref{eq:subset} for various tasks and models.

\begin{figure}[H]
  \centering
\pgfplotsset{width=\textwidth, height=8cm}
\input{fig/ablation/mnli_roberta_large}
\caption{
Subset size and evaluation accuracy of different subsets of the MNLI training set based on RoBERTa\textsubscript{\textrm{LARGE}}. Our proposed subsets are noted with vertical dotted lines.
}
\label{fig:mnli_ablation_large}
\end{figure}

\begin{figure}[H]
  \centering
\pgfplotsset{width=\textwidth, height=8cm}
\input{fig/ablation/mnli_roberta_base}
\caption{
Subset size and evaluation accuracy of different subsets of the MNLI training set based on RoBERTa\textsubscript{\textrm{BASE}}. Our proposed subsets are noted with vertical dotted lines.
}
\label{fig:mnli_ablation_base}
\end{figure}

\begin{figure}[H]
  \centering
\pgfplotsset{width=\textwidth, height=8cm}
\input{fig/ablation/snli_roberta_base}
\caption{
Subset size and evaluation accuracy of different subsets of the SNLI training set based on RoBERTa\textsubscript{\textrm{BASE}}. Our proposed subsets are noted with vertical dotted lines.
}
\label{fig:snli_ablation_base}
\end{figure}

\begin{figure}[H]
  \centering
\pgfplotsset{width=\textwidth, height=8cm}
\input{fig/ablation/sst2_roberta_base}
\caption{
Subset size and evaluation accuracy of different subsets of the SST-2 training set based on RoBERTa\textsubscript{\textrm{BASE}}. Our proposed subsets are noted with vertical dotted lines.
}
\label{fig:sst2_ablation_base}
\end{figure}

\begin{figure}[H]
  \centering
\pgfplotsset{width=\textwidth, height=8cm}
\input{fig/ablation/squadv2_roberta_large}
\caption{
Subset size and evaluation accuracy of different subsets of the SQuAD v2 training set based on RoBERTa\textsubscript{\textrm{LARGE}}. Our proposed subsets are noted with vertical dotted lines.
}
\label{fig:squadv2_ablation_large}
\end{figure}

\subsection{Details of Experimental Setup}
\label{sec:app:setup}

Our experiments are created on top of the existing fine-tuning codes in version $4.31.0$ of the transformers library\footnote{https://github.com/huggingface/transformers/}.
Specifically, for the MNLI, SNLI, and SST-2 tasks we use \texttt{run\_glue.py}, for the RACE task we use \texttt{run\_swag.py}, and for SQuAD v2 we use \texttt{run\_qa.py}.
All our experiments are done on an NVIDIA RTX 4090 GPU with version $12.1$ of the CUDA framework and version $2.1.0$ of PyTorch.

We use the \texttt{RobertaForMultipleChoice} class for fine-tuning RoBERTa\textsubscript{\textrm{LARGE}} on RACE.
For OPT\textsubscript{\textrm{350M}}, we use the \texttt{OPTForSequenceClassification} class and treat RACE as a sequence classification task.
To do this, we concatenate each choice to its corresponding article and question with the label set to true for the correct choice.

Table \ref{tab:hyperparams} contains the list of hyper-parameters used in the fine-tuning experiments.

\begin{table}[H]
    \centering
    \begin{tabular}{cc|ccccc}
     
     & & \begin{tabular}[c]{@{}c@{}}Learning Rate\end{tabular} & \begin{tabular}[c]{@{}c@{}}Batch Size\end{tabular} & \begin{tabular}[c]{@{}c@{}}Gradient\\Accumulation Steps\end{tabular} & \begin{tabular}[c]{@{}c@{}}Max Sequence\\Length\end{tabular} & \begin{tabular}[c]{@{}c@{}}Fine-tuning Epochs\end{tabular}
      \\
    \cmidrule[0.2pt](l{\tabcolsep}r{\tabcolsep}){1-7}
    
    \parbox[t]{5mm}{\multirow{6.5}{*}{\rotatebox[origin=c]{90}{\begin{tabular}[c]{@{}c@{}}RoBERTa\\Large\end{tabular}}}} & 
    MNLI & $10^{-5}$ & 32 & 1 & 128 & 3
    \\[5pt]
    & SNLI & $10^{-5}$ & 32 & 1 & 128 & 3
    \\[5pt]
    & SST-2 & $10^{-5}$ & 32 & 1 & 128 & 3
    \\[5pt]
    & RACE & $10^{-5}$ & 2 & 8 & 512 & 3
    \\[5pt]
    & SQuAD v2 & $3 \times 10^{-5}$ & 12 & 1 & 384 & 2
    \\

    \cmidrule[1pt](l{\abovetopsep}r{\belowbottomsep}){1-7}

    \parbox[t]{5mm}{\multirow{6.5}{*}{\rotatebox[origin=c]{90}{\begin{tabular}[c]{@{}c@{}}OPT\\350M\end{tabular}}}} & 
    MNLI &  $10^{-5}$ & 32 & 1 & 128 & 3
    \\[5pt]
    & SNLI &  $10^{-5}$ & 32 & 1 & 128 & 3
    \\[5pt]
    & SST-2 &  $10^{-5}$ & 32 & 1 & 128 & 3
    \\[5pt]
    & RACE & $10^{-5}$ & 8 & 2 & 512 & 3
    \\[5pt]
    & SQuAD v2 & $3 \times 10^{-5}$ & 12 & 1 & 384 & 2
    \\

\end{tabular}

    \caption{
    Values for hyper-parameters used in the fine-tuning experiments.
    Any hyper-parameter which is not mentioned is set to the default value of the framework.
    }
    \label{tab:hyperparams}
\end{table}

\subsection{Experimental Results}
\label{sec:app:results}

Tables \ref{tab:results1}, \ref{tab:results2}, and \ref{tab:results3} demonstrate the evaluation performance of our fine-tuning experiments described in Section \ref{sec:experiments}.
The subsets are distinguished by their size. 
Table \ref{tab:subset_size} provides the size of each subset discussed in Section \ref{sec:pruning}.

\begin{table}[H]
    \centering
    \begin{tabular}{cc|ccccccc|c}
     \multicolumn{10}{c}{\textbf{MNLI Matched} (Accuracy)} \\
     
    \cmidrule[1.5pt](l{\abovetopsep}r{\belowbottomsep}){1-10}
    
     & & \multicolumn{8}{c}{Subset Size} \\
     & & $5.42\%$ & $11.58\%$ & $12.27\%$ & $17.0\%$ & $20.7\%$ & $23.85\%$ & $27.06\%$ & $100\%$
      \\
    \cmidrule[0.2pt](l{\tabcolsep}r{\tabcolsep}){1-10}
    
    \parbox[t]{5mm}{\multirow{4}{*}{\rotatebox[origin=c]{90}{\begin{tabular}[c]{@{}c@{}}RoBERTa\\Large\end{tabular}}}} & 
    Ours & 
    $\mathbf{88.68}_{\hspace{1pt}\hspace{1pt}0.50}$ & $\mathbf{89.48}_{\hspace{1pt}0.09}$ & $\mathbf{89.91}_{\hspace{1pt}0.24}$ & $\underline{\mathbf{90.06}}_{\hspace{1pt}0.13}$ & $\underline{\mathbf{90.23}}_{\hspace{1pt}0.11}$ & $\underline{\mathbf{90.36}}_{\hspace{1pt}0.09}$ & $\underline{\mathbf{90.57}}_{\hspace{1pt}0.12}$
    & \multirow{4}{*}{$90.04_{\hspace{1pt}0.08}$}
    \\[5pt]
    & \textit{ambiguous} & 
    $26.03_{\hspace{1pt}6.14}$ & $31.23_{\hspace{1pt}0.85}$ & $31.59_{\hspace{1pt}1.43}$ & $37.56_{\hspace{1pt}9.94}$ & $77.30_{\hspace{1pt}1.86}$ & $87.74_{\hspace{1pt}0.39}$ & $89.55_{\hspace{1pt}0.05}$
    \\[5pt]
    & Random & 
    $87.68_{\hspace{1pt}0.36}$ & $88.98_{\hspace{1pt}0.12}$ & $88.94_{\hspace{1pt}0.28}$ & $89.25_{\hspace{1pt}0.22}$ & $89.52_{\hspace{1pt}0.08}$ & $89.56_{\hspace{1pt}0.31}$ & $89.72_{\hspace{1pt}0.34}$
    \\

    \cmidrule[1pt](l{\abovetopsep}r{\belowbottomsep}){1-10}

    & & \multicolumn{8}{c}{Subset Size} \\
     & & $7.18\%$ & $13.14\%$ & $17.14\%$ & $20.32\%$ & $25.62\%$ & $30.28\%$ & $34.97\%$ & $100\%$
      \\
    \cmidrule[0.2pt](l{\tabcolsep}r{\tabcolsep}){1-10}
    
    \parbox[t]{5mm}{\multirow{4}{*}{\rotatebox[origin=c]{90}{\begin{tabular}[c]{@{}c@{}}OPT\\350M\end{tabular}}}} & 
    Ours &
    $\mathbf{80.87}_{\hspace{1pt}0.39}$ & $\mathbf{81.87}_{\hspace{1pt}0.14}$ & $\mathbf{81.99}_{\hspace{1pt}0.44}$ & $\mathbf{82.37}_{\hspace{1pt}0.19}$ & $\mathbf{83.23}_{\hspace{1pt}0.14}$ & $\mathbf{83.38}_{\hspace{1pt}0.11}$ & $\mathbf{83.53}_{\hspace{1pt}0.37}$
    & \multirow{4}{*}{$83.98_{\hspace{1pt}0.18}$}
    \\[5pt]
    & \textit{ambiguous} & 
    $19.12_{\hspace{1pt}0.56}$ & $22.17_{\hspace{1pt}0.19}$ & $23.90_{\hspace{1pt}0.38}$ & $27.74_{\hspace{1pt}0.24}$ & $36.65_{\hspace{1pt}4.28}$ & $54.18_{\hspace{1pt}0.93}$ & $55.26_{\hspace{1pt}20.34}$
    \\[5pt]
    & Random & 
    $79.06_{\hspace{1pt}0.48}$ & $80.22_{\hspace{1pt}0.37}$ & $80.68_{\hspace{1pt}0.31}$ & $81.32_{\hspace{1pt}0.36}$ & $81.57_{\hspace{1pt}0.14}$ & $81.75_{\hspace{1pt}0.19}$ & $81.74_{\hspace{1pt}0.23}$
    \\
    
    \cmidrule[1.5pt](l{\abovetopsep}r{\belowbottomsep}){1-10}
     \multicolumn{10}{c}{\textbf{MNLI Mismatched} (Accuracy)} \\
    \cmidrule[1.5pt](l{\abovetopsep}r{\belowbottomsep}){1-10}

    & & \multicolumn{8}{c}{Subset Size} \\
     & & $5.42\%$ & $11.58\%$ & $12.27\%$ & $17.0\%$ & $20.7\%$ & $23.85\%$ & $27.06\%$ & $100\%$
      \\
    \cmidrule[0.2pt](l{\tabcolsep}r{\tabcolsep}){1-10}
    
    \parbox[t]{5mm}{\multirow{4}{*}{\rotatebox[origin=c]{90}{\begin{tabular}[c]{@{}c@{}}RoBERTa\\Large\end{tabular}}}} & 
    Ours & 
    $\mathbf{88.79}_{\hspace{1pt}0.35}$ & $\mathbf{89.26}_{\hspace{1pt}0.25}$ & $\mathbf{89.61}_{\hspace{1pt}0.18}$ & $\mathbf{89.81}_{\hspace{1pt}0.14}$ & $\underline{\mathbf{90.05}}_{\hspace{1pt}0.04}$ & $\underline{\mathbf{90.12}}_{\hspace{1pt}0.31}$ & $\underline{\mathbf{90.22}}_{\hspace{1pt}0.05}$
    & \multirow{4}{*}{$89.99_{\hspace{1pt}0.06}$}
    \\[5pt]
    & \textit{ambiguous} & 
    $26.28_{\hspace{1pt}6.14}$ & $30.22_{\hspace{1pt}1.47}$ & $30.22_{\hspace{1pt}1.87}$ & $37.76_{\hspace{1pt}10.29}$ & $76.53_{\hspace{1pt}2.12}$ & $87.74_{\hspace{1pt}0.39}$ & $89.27_{\hspace{1pt}0.17}$
    \\[5pt]
    & Random & 
    $87.79_{\hspace{1pt}0.14}$ & $88.61_{\hspace{1pt}0.14}$ & $88.74_{\hspace{1pt}0.15}$ & $89.16_{\hspace{1pt}0.37}$ & $89.12_{\hspace{1pt}0.30}$ & $89.20_{\hspace{1pt}0.31}$ & $89.27_{\hspace{1pt}0.34}$
    \\

    \cmidrule[1pt](l{\abovetopsep}r{\belowbottomsep}){1-10}

    & & \multicolumn{8}{c}{Subset Size} \\
     & & $7.18\%$ & $13.14\%$ & $17.14\%$ & $20.32\%$ & $25.62\%$ & $30.28\%$ & $34.97\%$ & $100\%$
      \\
    \cmidrule[0.2pt](l{\tabcolsep}r{\tabcolsep}){1-10}
    
    \parbox[t]{5mm}{\multirow{4}{*}{\rotatebox[origin=c]{90}{\begin{tabular}[c]{@{}c@{}}OPT\\350M\end{tabular}}}} & 
    Ours &
    $\mathbf{81.83}_{\hspace{1pt}0.35}$ & $\mathbf{82.56}_{\hspace{1pt}0.25}$ & $\mathbf{82.85}_{\hspace{1pt}0.18}$ & $\mathbf{83.55}_{\hspace{1pt}0.14}$ & $\mathbf{83.70}_{\hspace{1pt}0.04}$ & $\underline{\mathbf{84.15}}_{\hspace{1pt}0.31}$ & $\underline{\mathbf{84.29}}_{\hspace{1pt}0.05}$
    & \multirow{4}{*}{$83.98_{\hspace{1pt}0.18}$}
    \\[5pt]
    & \textit{ambiguous} & 
    $17.68_{\hspace{1pt}1.11}$ & $20.47_{\hspace{1pt}0.34}$ & $22.28_{\hspace{1pt}0.30}$ & $25.63_{\hspace{1pt}0.74}$ & $35.66_{\hspace{1pt}3.34}$ & $52.17_{\hspace{1pt}0.82}$ & $54.31_{\hspace{1pt}19.52}$
    \\[5pt]
    & Random & 
    $79.81_{\hspace{1pt}0.10}$ & $81.11_{\hspace{1pt}0.16}$ & $81.52_{\hspace{1pt}0.15}$ & $81.68_{\hspace{1pt}0.58}$ & $82.08_{\hspace{1pt}0.36}$ & $82.37_{\hspace{1pt}0.44}$ & $82.53_{\hspace{1pt}0.13}$
    \\

    \cmidrule[1.5pt](l{\abovetopsep}r{\belowbottomsep}){1-10}
     \multicolumn{10}{c}{\textbf{SNLI} (Accuracy)} \\
    \cmidrule[1.5pt](l{\abovetopsep}r{\belowbottomsep}){1-10}

   & & \multicolumn{8}{c}{Subset Size} \\
     & & 
     $4.2\%$ & $9.27\%$ & $9.52\%$ & $13.47\%$ & $16.39\%$ & $18.8\%$ & $21.26\%$
     & $100\%$ \\
    \cmidrule[0.2pt](l{\tabcolsep}r{\tabcolsep}){1-10}
    
    \parbox[t]{5mm}{\multirow{4}{*}{\rotatebox[origin=c]{90}{\begin{tabular}[c]{@{}c@{}}RoBERTa\\Large\end{tabular}}}} & 
    Ours & 
    $\mathbf{91.41}_{\hspace{1pt}0.31}$ & $\mathbf{91.84}_{\hspace{1pt}0.15}$ & $\mathbf{91.66}_{\hspace{1pt}0.16}$ & $\mathbf{92.13}_{\hspace{1pt}0.22}$ & $\underline{\mathbf{92.38}}_{\hspace{1pt}0.03}$ & $\underline{\mathbf{92.41}}_{\hspace{1pt}0.03}$ & $\underline{\mathbf{92.50}}_{\hspace{1pt}0.14}$
    & \multirow{4}{*}{$92.25_{\hspace{1pt}0.12}$}
    \\[5pt]
    & \textit{ambiguous} & 
    $30.89_{\hspace{1pt}3.57}$ & $58.45_{\hspace{1pt}4.81}$ & $61.80_{\hspace{1pt}5.09}$ & $67.18_{\hspace{1pt}29.80}$ & $89.23_{\hspace{1pt}0.25}$ & $90.24_{\hspace{1pt}1.05}$ & $91.33_{\hspace{1pt}0.13}$
    \\[5pt]
    & Random & 
    $90.21_{\hspace{1pt}0.32}$ & $90.75_{\hspace{1pt}0.28}$ & $90.74_{\hspace{1pt}0.06}$ & $91.09_{\hspace{1pt}0.35}$ & $90.25_{\hspace{1pt}2.17}$ & $91.26_{\hspace{1pt}0.22}$ & $91.55_{\hspace{1pt}0.05}$
    \\

    \cmidrule[1pt](l{\abovetopsep}r{\belowbottomsep}){1-10}

    & & \multicolumn{8}{c}{Subset Size} \\
     & & 
     $5.34\%$ & $10.71\%$ & $12.26\%$ & $16.06\%$ & $19.79\%$ & $22.97\%$ & $26.13\%$
      & $100\%$ \\
    \cmidrule[0.2pt](l{\tabcolsep}r{\tabcolsep}){1-10}
    
    \parbox[t]{5mm}{\multirow{4}{*}{\rotatebox[origin=c]{90}{\begin{tabular}[c]{@{}c@{}}OPT\\350M\end{tabular}}}} & 
    Ours &
    $\mathbf{87.34}_{\hspace{1pt}0.12}$ & $\mathbf{88.44}_{\hspace{1pt}0.03}$ & $\mathbf{88.21}_{\hspace{1pt}0.32}$ & $\mathbf{89.13}_{\hspace{1pt}0.13}$ & $\mathbf{89.58}_{\hspace{1pt}0.11}$ & $\underline{\mathbf{89.87}}_{\hspace{1pt}0.14}$ & $\underline{\mathbf{90.07}}_{\hspace{1pt}0.29}$
    & \multirow{4}{*}{$89.75_{\hspace{1pt}0.21}$}
    \\[5pt]
    & \textit{ambiguous} & 
    $17.16_{\hspace{1pt}0.16}$ & $21.57_{\hspace{1pt}0.48}$ & $24.22_{\hspace{1pt}1.07}$ & $41.35_{\hspace{1pt}3.83}$ & $67.50_{\hspace{1pt}1.84}$ & $79.92_{\hspace{1pt}1.01}$ & $85.54_{\hspace{1pt}0.85}$
    \\[5pt]
    & Random & 
    $85.21_{\hspace{1pt}0.29}$ & $86.52_{\hspace{1pt}0.09}$ & $86.43_{\hspace{1pt}0.06}$ & $87.34_{\hspace{1pt}0.19}$ & $87.62_{\hspace{1pt}0.10}$ & $87.76_{\hspace{1pt}0.09}$ & $87.97_{\hspace{1pt}0.21}$
    \\
    \cmidrule[1.5pt](l{\abovetopsep}r{\belowbottomsep}){1-10}

\end{tabular}

    \caption{
    Evaluation accuracy of fine-tuning RoBERTa\textsubscript{\textrm{LARGE}} and OPT\textsubscript{\textrm{350M}} on various training subsets.
    For each subset size, the best result is indicated with a bold font.
    Results that outperform the baseline are indicated with underline.
    Each reported accuracy is the average of 3 runs with different intialization seeds.
    The standard deviation of the runs is also reported as the subscript.
    Last column reports the accuracy of fine-tuning on the entire dataset (baseline).
    }
    \label{tab:results1}
\end{table}

\begin{table}[H]
    \centering
    \begin{tabular}{cc|ccccccc|c}
     \multicolumn{10}{c}{\textbf{SST-2} (Accuracy)} \\
     
    \cmidrule[1.5pt](l{\abovetopsep}r{\belowbottomsep}){1-10}
    
     & & \multicolumn{8}{c}{Subset Size} \\
     & &
     $4.6\%$ & $9.11\%$ & $13.21\%$ & $17.81\%$ & $20.41\%$ & $22.32\%$ & $23.95\%$
     & $100\%$
      \\
    \cmidrule[0.2pt](l{\tabcolsep}r{\tabcolsep}){1-10}
    
    \parbox[t]{5mm}{\multirow{4}{*}{\rotatebox[origin=c]{90}{\begin{tabular}[c]{@{}c@{}}RoBERTa\\Large\end{tabular}}}} & 
    Ours & 
    $\mathbf{94.15}_{\hspace{1pt}0.35}$ & $\mathbf{94.65}_{\hspace{1pt}0.46}$ & $\mathbf{94.99}_{\hspace{1pt}0.17}$ & $\mathbf{95.45}_{\hspace{1pt}0.24}$ & $\mathbf{95.53}_{\hspace{1pt}0.42}$ & $\mathbf{95.87}_{\hspace{1pt}0.23}$ & $\underline{\mathbf{95.80}}_{\hspace{1pt}0.37}$
    & \multirow{4}{*}{$95.79_{\hspace{1pt}0.58}$}
    \\[5pt]
    & \textit{ambiguous} & 
    $50.69_{\hspace{1pt}0.40}$ & $50.54_{\hspace{1pt}1.48}$ & $74.62_{\hspace{1pt}20.33}$ & $94.84_{\hspace{1pt}0.50}$ & $94.95_{\hspace{1pt}1.05}$ & $95.68_{\hspace{1pt}0.40}$ & $95.76_{\hspace{1pt}0.42}$
    \\[5pt]
    & Random & 
    $93.54_{\hspace{1pt}0.52}$ & $94.27_{\hspace{1pt}0.46}$ & $94.65_{\hspace{1pt}0.76}$ & $94.69_{\hspace{1pt}0.37}$ & $94.92_{\hspace{1pt}0.58}$ & $94.95_{\hspace{1pt}0.35}$ & $95.26_{\hspace{1pt}0.63}$
    \\

    \cmidrule[1pt](l{\abovetopsep}r{\belowbottomsep}){1-10}

    & & \multicolumn{8}{c}{Subset Size} \\
     & &
     $4.37\%$ & $8.58\%$ & $10.71\%$ & $15.08\%$ & $17.57\%$ & $19.29\%$ & $20.82\%$
     & $100\%$
      \\
    \cmidrule[0.2pt](l{\tabcolsep}r{\tabcolsep}){1-10}
    
    \parbox[t]{5mm}{\multirow{4}{*}{\rotatebox[origin=c]{90}{\begin{tabular}[c]{@{}c@{}}OPT\\350M\end{tabular}}}} & 
    Ours &
    $\mathbf{91.67}_{\hspace{1pt}0.33}$ & $90.98_{\hspace{1pt}0.35}$ & $\underline{\mathbf{93.19}}_{\hspace{1pt}0.18}$ & $\underline{\mathbf{93.81}}_{\hspace{1pt}0.12}$ & $\underline{\mathbf{93.73}}_{\hspace{1pt}0.07}$ & $\underline{\mathbf{94.19}}_{\hspace{1pt}0.29}$ & $\underline{\mathbf{94.04}}_{\hspace{1pt}0.12}$
    & \multirow{4}{*}{$92.85_{\hspace{1pt}0.52}$}
    \\[5pt]
    & \textit{ambiguous} & 
    $16.63_{\hspace{1pt}4.54}$ & $45.68_{\hspace{1pt}18.05}$ & $77.45_{\hspace{1pt}3.39}$ & $90.41_{\hspace{1pt}0.37}$ & $91.44_{\hspace{1pt}1.07}$ & $92.01_{\hspace{1pt}0.35}$ & $92.24_{\hspace{1pt}0.24}$
    \\[5pt]
    & Random & 
    $91.29_{\hspace{1pt}0.50}$ & $\mathbf{91.59}_{\hspace{1pt}0.87}$ & $91.74_{\hspace{1pt}0.59}$ & $92.05_{\hspace{1pt}0.48}$ & $92.09_{\hspace{1pt}0.20}$ & $92.24_{\hspace{1pt}0.95}$ & $92.78_{\hspace{1pt}0.52}$
    \\
    
    \cmidrule[1.5pt](l{\abovetopsep}r{\belowbottomsep}){1-10}
     \multicolumn{10}{c}{\textbf{RACE} (Accuracy)} \\
    \cmidrule[1.5pt](l{\abovetopsep}r{\belowbottomsep}){1-10}

    & & \multicolumn{8}{c}{Subset Size} \\
     & & 
     $14.97\%$ & $25.56\%$ & $30.47\%$ & $40.53\%$ & $49.39\%$ & $56.03\%$ & $62.39\%$
     & $100\%$
      \\
    \cmidrule[0.2pt](l{\tabcolsep}r{\tabcolsep}){1-10}
    
    \parbox[t]{5mm}{\multirow{4}{*}{\rotatebox[origin=c]{90}{\begin{tabular}[c]{@{}c@{}}RoBERTa\\Large\end{tabular}}}} & 
    Ours & 
    $\mathbf{77.79}_{\hspace{1pt}0.39}$ & $78.73_{\hspace{1pt}0.25}$ & $\mathbf{80.56}_{\hspace{1pt}0.73}$ & $\mathbf{80.62}_{\hspace{1pt}0.12}$ & $81.36_{\hspace{1pt}0.43}$ & $\mathbf{82.36}_{\hspace{1pt}0.54}$ & $\mathbf{82.95}_{\hspace{1pt}0.13}$
    & \multirow{4}{*}{$84.67_{\hspace{1pt}0.17}$}
    \\[5pt]
    & \textit{ambiguous} & 
    $24.02_{\hspace{1pt}1.25}$ & $23.56_{\hspace{1pt}0.95}$ & $23.68_{\hspace{1pt}1.43}$ & $43.58_{\hspace{1pt}33.44}$ & $62.47_{\hspace{1pt}34.48}$ & $24.36_{\hspace{1pt}0.96}$ & $43.21_{\hspace{1pt}34.81}$
    \\[5pt]
    & Random & 
    $76.33_{\hspace{1pt}0.84}$ & $\mathbf{78.83}_{\hspace{1pt}0.44}$ & $79.52_{\hspace{1pt}0.64}$ & $80.39_{\hspace{1pt}0.44}$ & $\mathbf{81.47}_{\hspace{1pt}0.19}$ & $82.05_{\hspace{1pt}0.44}$ & $82.44_{\hspace{1pt}0.34}$
    \\

    \cmidrule[1pt](l{\abovetopsep}r{\belowbottomsep}){1-10}

    & & \multicolumn{8}{c}{Subset Size} \\
     & & 
     $4.06\%$ & $7.71\%$ & $10.35\%$ & $11.77\%$ & $14.98\%$ & $18.06\%$ & $21.62\%$
     & $100\%$
      \\
    \cmidrule[0.2pt](l{\tabcolsep}r{\tabcolsep}){1-10}
    
    \parbox[t]{5mm}{\multirow{4}{*}{\rotatebox[origin=c]{90}{\begin{tabular}[c]{@{}c@{}}OPT\\350M\end{tabular}}}} & 
    Ours &
    $\mathbf{76.64}_{\hspace{1pt}0.42}$ & $\mathbf{77.60}_{\hspace{1pt}0.25}$ & $\mathbf{74.59}_{\hspace{1pt}0.20}$ & $\mathbf{78.50}_{\hspace{1pt}0.10}$ & $\mathbf{78.85}_{\hspace{1pt}0.06}$ & $\mathbf{79.26}_{\hspace{1pt}0.25}$ & $\underline{\mathbf{79.68}}_{\hspace{1pt}0.09}$
    & \multirow{4}{*}{$79.34_{\hspace{1pt}0.20}$}
    \\[5pt]
    & \textit{ambiguous} & 
    $33.30_{\hspace{1pt}4.03}$ & $34.60_{\hspace{1pt}4.05}$ & $34.34_{\hspace{1pt}5.99}$ & $36.07_{\hspace{1pt}8.37}$ & $32.99_{\hspace{1pt}0.30}$ & $35.00_{\hspace{1pt}2.55}$ & $45.97_{\hspace{1pt}4.04}$
    \\[5pt]
    & Random & 
    $72.92_{\hspace{1pt}2.54}$ & $70.14_{\hspace{1pt}1.45}$ & $71.65_{\hspace{1pt}0.86}$ & $72.56_{\hspace{1pt}2.78}$ & $73.98_{\hspace{1pt}2.01}$ & $74.23_{\hspace{1pt}0.87}$ & $76.09_{\hspace{1pt}0.46}$
    \\

    \cmidrule[1.5pt](l{\abovetopsep}r{\belowbottomsep}){1-10}
    
\end{tabular}

    \caption{
    Evaluation accuracy of fine-tuning RoBERTa\textsubscript{\textrm{LARGE}} and OPT\textsubscript{\textrm{350M}} on various training subsets.
    For each subset size, the best result is indicated with a bold font.
    Results that outperform the baseline are indicated with underline.
    Each reported accuracy is the average of 3 runs with different initialization seeds.
    The standard deviation of the runs is also reported as the subscript.
    Last column reports the accuracy of fine-tuning on the entire dataset (baseline).
    }
    \label{tab:results2}
\end{table}

\begin{table}[H]
    \centering
    \begin{tabular}{cc|ccccccc|c}
     \multicolumn{10}{c}{\textbf{SQuAD v2} (\begin{tabular}[c]{@{}c@{}}Exact Match\\F1 Score\end{tabular})} \\
     
    \cmidrule[1.5pt](l{\abovetopsep}r{\belowbottomsep}){1-10}
    
     & & \multicolumn{8}{c}{Subset Size} \\
     & &
     $9.54\%$ & $16.61\%$ & $23.2\%$ & $26.15\%$ & $33.44\%$ & $39.81\%$ & $46.66\%$
     & $100\%$
      \\
    \cmidrule[0.2pt](l{\tabcolsep}r{\tabcolsep}){1-10}
    
    \parbox[t]{5mm}{\multirow{7}{*}{\rotatebox[origin=c]{90}{\begin{tabular}[c]{@{}c@{}}RoBERTa\\Large\end{tabular}}}} & 
    \multirow{2}{*}{Ours} & 
    $\mathbf{82.13}_{\hspace{1pt}0.20}$ & $81.40_{\hspace{1pt}0.20}$ & $\underline{\mathbf{84.03}}_{\hspace{1pt}0.34}$ & $82.84_{\hspace{1pt}0.41}$ & $83.80_{\hspace{1pt}0.22}$ & $\underline{84.39}_{\hspace{1pt}0.25}$ & $\underline{85.09}_{\hspace{1pt}0.14}$
    & \multirow{7}{*}{
    \begin{tabular}[c]{@{}c@{}}$83.94_{\hspace{1pt}0.62}$\\$86.92_{\hspace{1pt}0.66}$\end{tabular}
    }
    \\
    & &
    $\mathbf{84.94}_{\hspace{1pt}0.07}$ & $84.06_{\hspace{1pt}0.28}$ & $\mathbf{86.62}_{\hspace{1pt}0.26}$ & $85.42_{\hspace{1pt}0.45}$ & $86.42_{\hspace{1pt}0.30}$ & $86.91_{\hspace{1pt}0.28}$ & $\underline{87.70}_{\hspace{1pt}0.17}$
    \\[5pt]
    & \multirow{2}{*}{\textit{ambiguous}} & 
    $60.11_{\hspace{1pt}1.30}$ & $78.87_{\hspace{1pt}0.67}$ & $82.90_{\hspace{1pt}0.40}$ & $\mathbf{83.57}_{\hspace{1pt}0.38}$ & $\underline{\mathbf{84.93}}_{\hspace{1pt}0.07}$ & $\underline{\mathbf{85.19}}_{\hspace{1pt}0.38}$ & $\underline{\mathbf{85.32}}_{\hspace{1pt}0.15}$
    \\
    &&
    $60.51_{\hspace{1pt}1.39}$ & $80.70_{\hspace{1pt}0.65}$ & $85.25_{\hspace{1pt}0.40}$ & $\mathbf{85.99}_{\hspace{1pt}0.37}$ & $\underline{\mathbf{87.51}}_{\hspace{1pt}0.06}$ & $\underline{\mathbf{87.87}}_{\hspace{1pt}0.36}$ & $\underline{\mathbf{88.09}}_{\hspace{1pt}0.10}$
    \\[5pt]
    & \multirow{2}{*}{Random} & 
    $79.65_{\hspace{1pt}0.92}$ & $\mathbf{81.60}_{\hspace{1pt}0.44}$ & $81.96_{\hspace{1pt}0.60}$ & $82.70_{\hspace{1pt}0.90}$ & $82.71_{\hspace{1pt}0.07}$ & $83.44_{\hspace{1pt}0.28}$ & $83.52_{\hspace{1pt}0.10}$
    \\
    &&
    $83.00_{\hspace{1pt}0.83}$ & $\mathbf{84.80}_{\hspace{1pt}0.61}$ & $85.14_{\hspace{1pt}0.59}$ & $85.77_{\hspace{1pt}0.90}$ & $85.86_{\hspace{1pt}0.10}$ & $86.54_{\hspace{1pt}0.27}$ & $86.60_{\hspace{1pt}0.11}$
    \\

    \cmidrule[1pt](l{\abovetopsep}r{\belowbottomsep}){1-10}

    & & \multicolumn{8}{c}{Subset Size} \\
     & &
     $8.38\%$ & $10.99\%$ & $19.38\%$ & $24.10\%$ & $27.12\%$ & $35.09\%$ & $45.13\%$
     & $100\%$
      \\
    \cmidrule[0.2pt](l{\tabcolsep}r{\tabcolsep}){1-10}
    
    \parbox[t]{5mm}{\multirow{7}{*}{\rotatebox[origin=c]{90}{\begin{tabular}[c]{@{}c@{}}OPT\\350M\end{tabular}}}} & 
    \multirow{2}{*}{Ours} & 
    $\mathbf{59.75}_{\hspace{1pt}0.34}$ & $\mathbf{59.54}_{\hspace{1pt}0.17}$ & $\mathbf{61.00}_{\hspace{1pt}0.14}$ & $\mathbf{62.50}_{\hspace{1pt}0.45}$ & $\mathbf{61.77}_{\hspace{1pt}0.21}$ & $\mathbf{62.86}_{\hspace{1pt}0.12}$ & $\underline{\mathbf{63.72}}_{\hspace{1pt}0.38}$
    & \multirow{7}{*}{
    \begin{tabular}[c]{@{}c@{}}$63.26_{\hspace{1pt}0.31}$\\$67.03_{\hspace{1pt}0.23}$\end{tabular}
    }
    \\
    & &
    $\mathbf{62.19}_{\hspace{1pt}0.19}$ & $\mathbf{61.87}_{\hspace{1pt}0.20}$ & $\mathbf{63.47}_{\hspace{1pt}0.10}$ & $\mathbf{65.14}_{\hspace{1pt}0.34}$ & $\mathbf{64.30}_{\hspace{1pt}0.31}$ & $\mathbf{65.44}_{\hspace{1pt}0.13}$ & $\mathbf{66.49}_{\hspace{1pt}0.22}$
    \\[5pt]
    & \multirow{2}{*}{\textit{ambiguous}} & 
    $48.97_{\hspace{1pt}0.35}$ & $51.59_{\hspace{1pt}1.06}$ & $57.90_{\hspace{1pt}0.54}$ & $59.66_{\hspace{1pt}0.68}$ & $61.02_{\hspace{1pt}0.70}$ & $62.30_{\hspace{1pt}0.36}$ & $62.98_{\hspace{1pt}0.41}$
    \\
    &&
    $51.30_{\hspace{1pt}0.50}$ & $54.25_{\hspace{1pt}1.00}$ & $60.91_{\hspace{1pt}0.22}$ & $62.71_{\hspace{1pt}0.47}$ & $64.12_{\hspace{1pt}0.52}$ & $65.38_{\hspace{1pt}0.28}$ & $66.25_{\hspace{1pt}0.43}$
    \\[5pt]
    & \multirow{2}{*}{Random} & 
    $55.19_{\hspace{1pt}0.50}$ & $56.37_{\hspace{1pt}0.74}$ & $58.87_{\hspace{1pt}0.88}$ & $59.17_{\hspace{1pt}0.38}$ & $60.05_{\hspace{1pt}0.15}$ & $60.76_{\hspace{1pt}0.08}$ & $61.66_{\hspace{1pt}0.27}$
    \\
    &&
    $59.52_{\hspace{1pt}0.28}$ & $60.72_{\hspace{1pt}0.68}$ & $62.88_{\hspace{1pt}0.73}$ & $63.15_{\hspace{1pt}0.28}$ & $64.12_{\hspace{1pt}0.18}$ & $64.78_{\hspace{1pt}0.17}$ & $65.62_{\hspace{1pt}0.13}$
    \\
    
    \cmidrule[1.5pt](l{\abovetopsep}r{\belowbottomsep}){1-10}

\end{tabular}

    \caption{
    Evaluation accuracy of fine-tuning RoBERTa\textsubscript{\textrm{LARGE}} and OPT\textsubscript{\textrm{350M}} on training subsets of SQuAD v2 task.
    For each subset size, the best result is indicated with a bold font.
    Results that outperform the baseline are indicated with underline.
    Each reported metric is the average of 3 runs with different initialization seeds.
    The standard deviation of the runs is also reported as the subscript.
    Last column reports the accuracy of fine-tuning on the entire dataset (baseline).
    }
    \label{tab:results3}
\end{table}

\begin{table}[H]
    \centering
    \begin{tabular}{cc|ccccccc}
    
     & & \multicolumn{7}{c}{Subset} \\
     & & $\mathcal{D}_{\{2,3,4\}}$ & $\mathcal{D}_{\{4\}}$ & $\mathcal{D}_{\{5\}}$ & $\mathcal{D}_{\{4,5\}}$ & $\mathcal{D}_{\{3,4,5\}}$ & $\mathcal{D}_{\{2,3,4,5\}}$ & $\mathcal{D}_{\{1,2,3,4,5\}}$
      \\
    \cmidrule[0.2pt](l{\tabcolsep}r{\tabcolsep}){1-9}
    
    \parbox[t]{5mm}{\multirow{6.5}{*}{\rotatebox[origin=c]{90}{\begin{tabular}[c]{@{}c@{}}RoBERTa\\Large\end{tabular}}}} & 
    MNLI & 
    $12.27\%$ & $5.42\%$ & $11.58\%$ & $17.00\%$ & $20.70\%$ & $23.85\%$ & $27.06\%$
    \\[5pt]
    & SNLI & 
    $9.52\%$ & $4.2\%$ & $9.27\%$ & $13.47\%$ & $16.39\%$ & $18.8\%$ & $21.26\%$
    \\[5pt]
    & SST-2 & 
    $9.11\%$ & $4.6\%$ & $13.21\%$ & $17.81\%$ & $20.41\%$ & $22.32\%$ & $23.95\%$
    \\[5pt]
    & RACE & 
    $30.47\%$ & $14.97\%$ & $25.26\%$ & $40.53\%$ & $49.39\%$ & $56.03\%$ & $62.39\%$
    \\[5pt]
    & SQuAD v2 & 
    $23.2\%$ & $9.54\%$ & $16.61\%$ & $26.15\%$ & $33.44\%$ & $39.81\%$ & $46.66\%$
    \\

    \cmidrule[1pt](l{\abovetopsep}r{\belowbottomsep}){1-9}

    \parbox[t]{5mm}{\multirow{6.5}{*}{\rotatebox[origin=c]{90}{\begin{tabular}[c]{@{}c@{}}OPT\\350M\end{tabular}}}} & 
    MNLI & 
    $17.14\%$ & $7.18\%$ & $13.14\%$ & $20.32\%$ & $25.62\%$ & $30.28\%$ & $34.97\%$
    \\[5pt]
    & SNLI & 
    $12.26\%$ & $5.34\%$ & $10.71\%$ & $16.06\%$ & $19.79\%$ & $22.97\%$ & $26.13\%$
    \\[5pt]
    & SST-2 & 
    $8.58\%$ & $4.37\%$ & $10.71\%$ & $15.08\%$ & $17.57\%$ & $19.29\%$ & $20.82\%$
    \\[5pt]
    & RACE & 
    $10.35\%$ & $4.06\%$ & $7.71\%$ & $11.77\%$ & $14.98\%$ & $18.06\%$ & $21.62\%$
    \\[5pt]
    & SQuAD v2 & 
    $19.38\%$ & $8.38\%$ & $10.99\%$ & $24.1\%$ & $27.12\%$ & $35.09\%$ & $45.13\%$
    \\
    
    \cmidrule[1.5pt](l{\abovetopsep}r{\belowbottomsep}){1-9}

\end{tabular}

    \caption{
    Relative size for each of our proposed subsets used for dataset pruning compared to the training set.
    $\mathcal{D}_{\{1,2,3,4,5\}}$ is also referred to as the winning ticket subset.
    }
    \label{tab:subset_size}
\end{table}

\subsection{Further Comparison with \textit{ambiguous} Subsets}
\label{sec:app:datamap}

\begin{figure}[H]
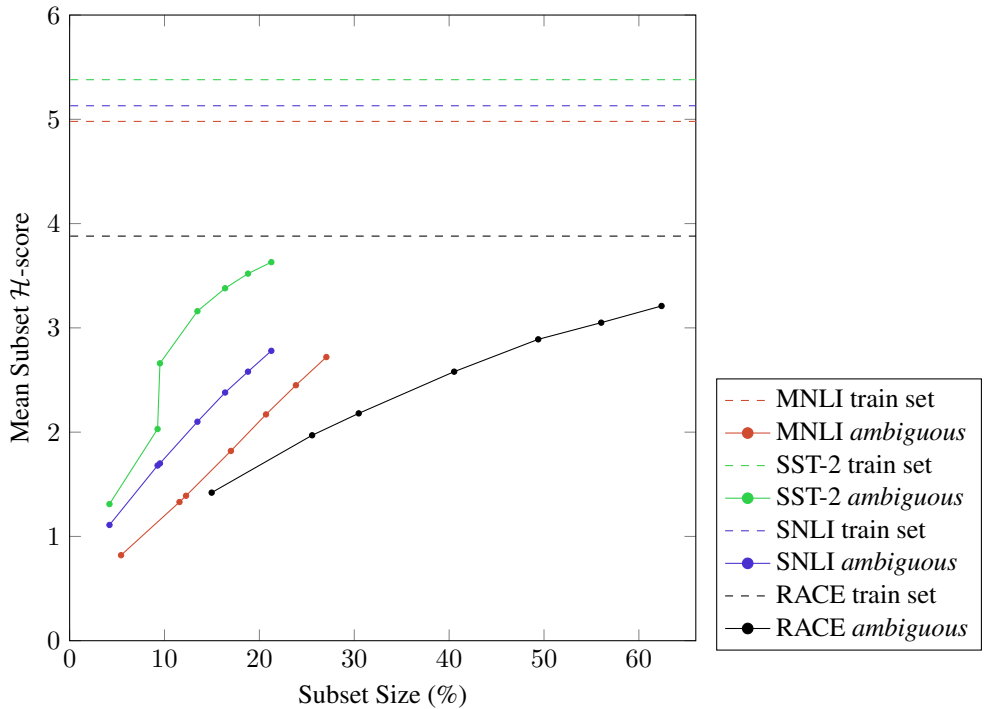

    \centering
    \pgfplotsset{width=0.6\columnwidth, height=0.6\columnwidth}
    \input{fig/mean_score/large_roberta}
    \pgfplotsset{width=0.3\columnwidth, height=0.3\columnwidth}
    \input{fig/mean_score/legend2}
    \caption{Mean $\mathcal{H}$-score of training subsets along with the training set for MNLI, SST-2, and SNLI tasks, based on RoBERTa\textsubscript{\textrm{LARGE}}}
    \label{fig:more_meanh}
\end{figure}

\begin{figure}[H]
  \centering
  \begin{subfigure}{0.48\textwidth}
    \centering
    \pgfplotsset{width=\columnwidth, height=\columnwidth}
    \input{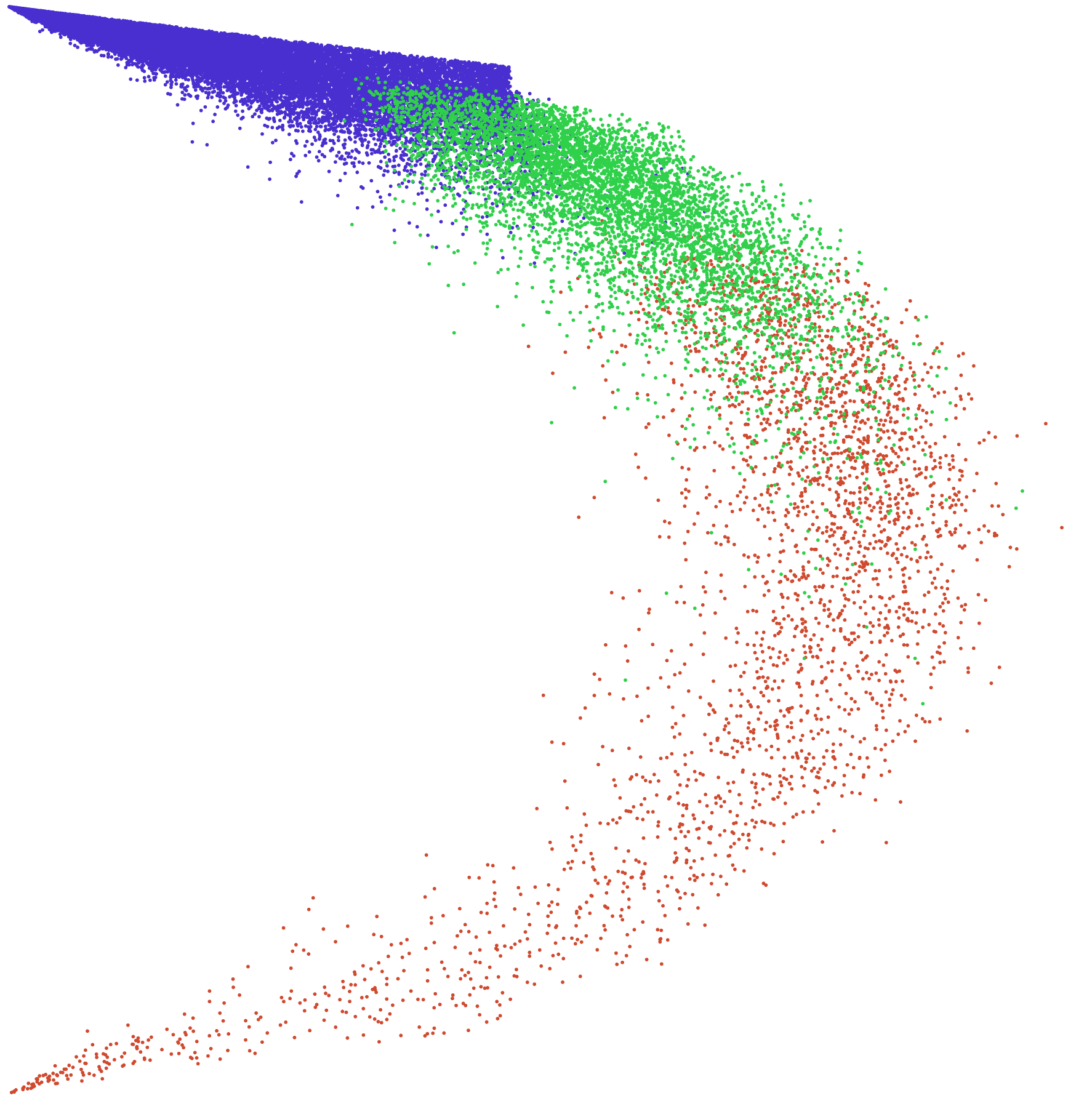}
    \caption{}
    \label{fig:carto_sst2_opt}
  \end{subfigure}
  \hfill
  \begin{subfigure}{0.48\textwidth}
    \centering
    \pgfplotsset{width=\columnwidth, height=\columnwidth}
    \input{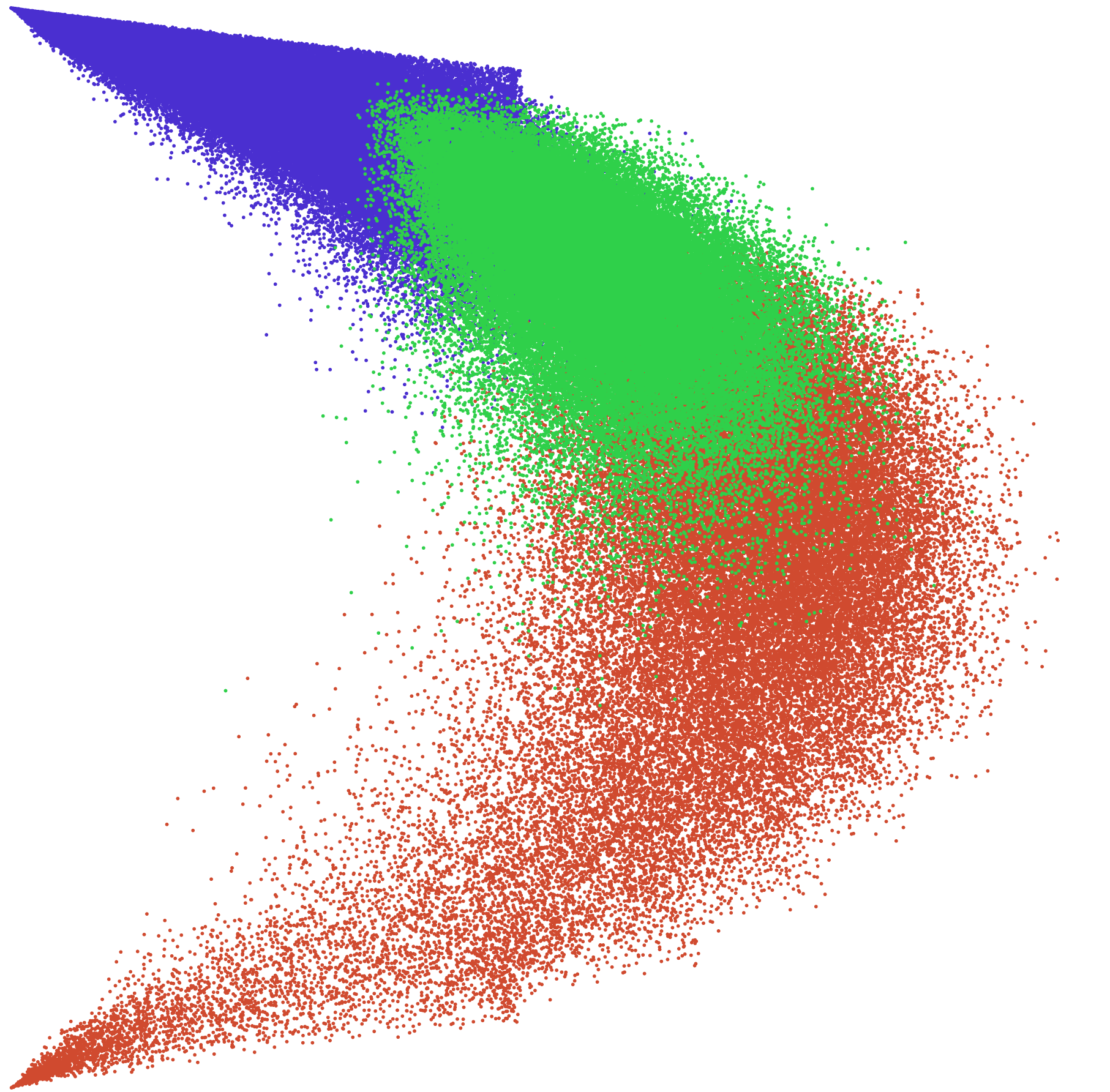}
    \caption{}
    \label{fig:carto_mnli_opt}
  \end{subfigure}
  \vfill
  \centering
  \begin{subfigure}{0.48\textwidth}
    \centering
    \pgfplotsset{width=\columnwidth, height=\columnwidth}
    \input{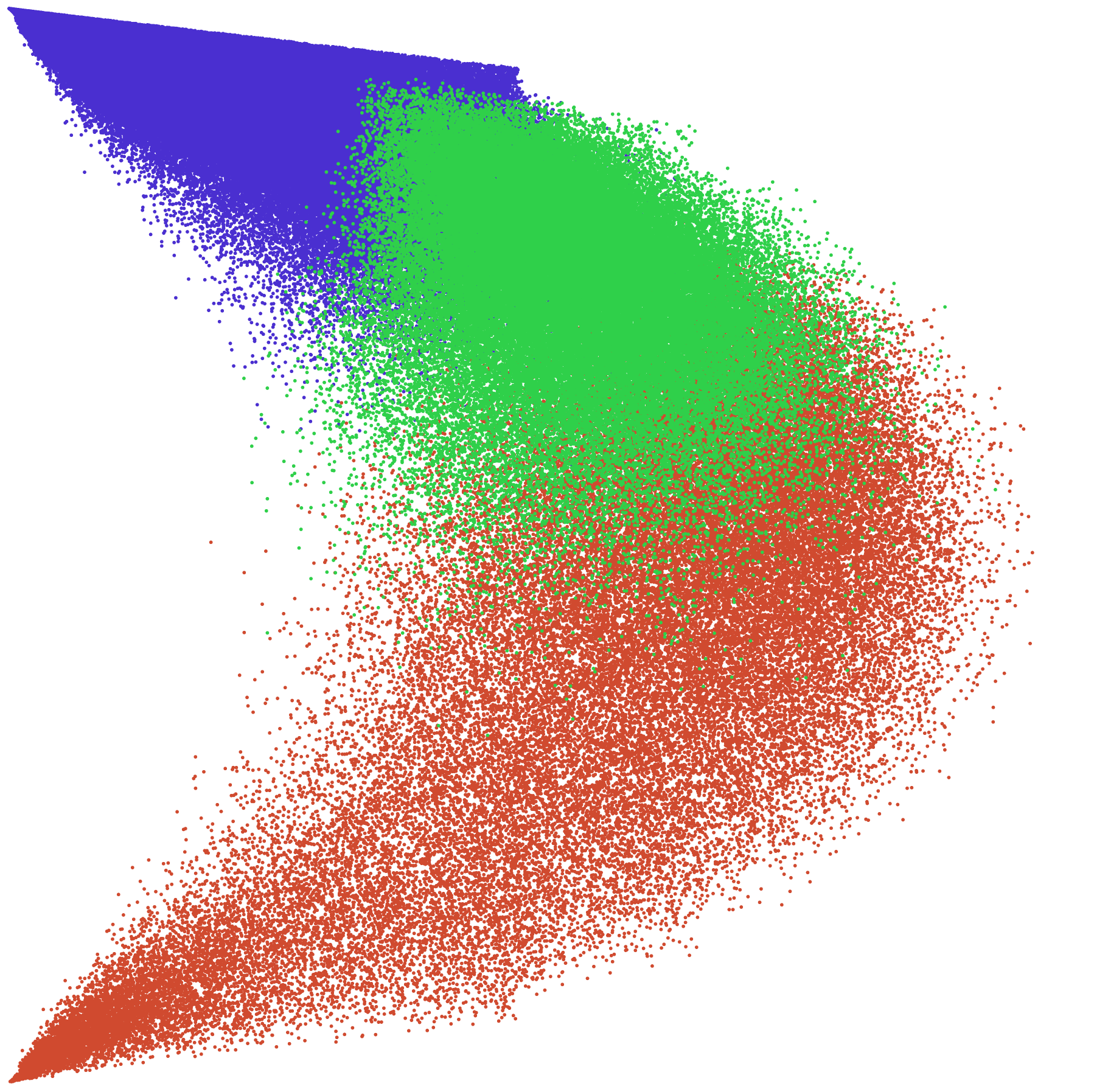}
    \caption{}
    \label{fig:carto_snli_opt}
  \end{subfigure}
  \hfill
  \begin{subfigure}{0.48\textwidth}
    \centering
    \pgfplotsset{width=\columnwidth, height=\columnwidth}
    \input{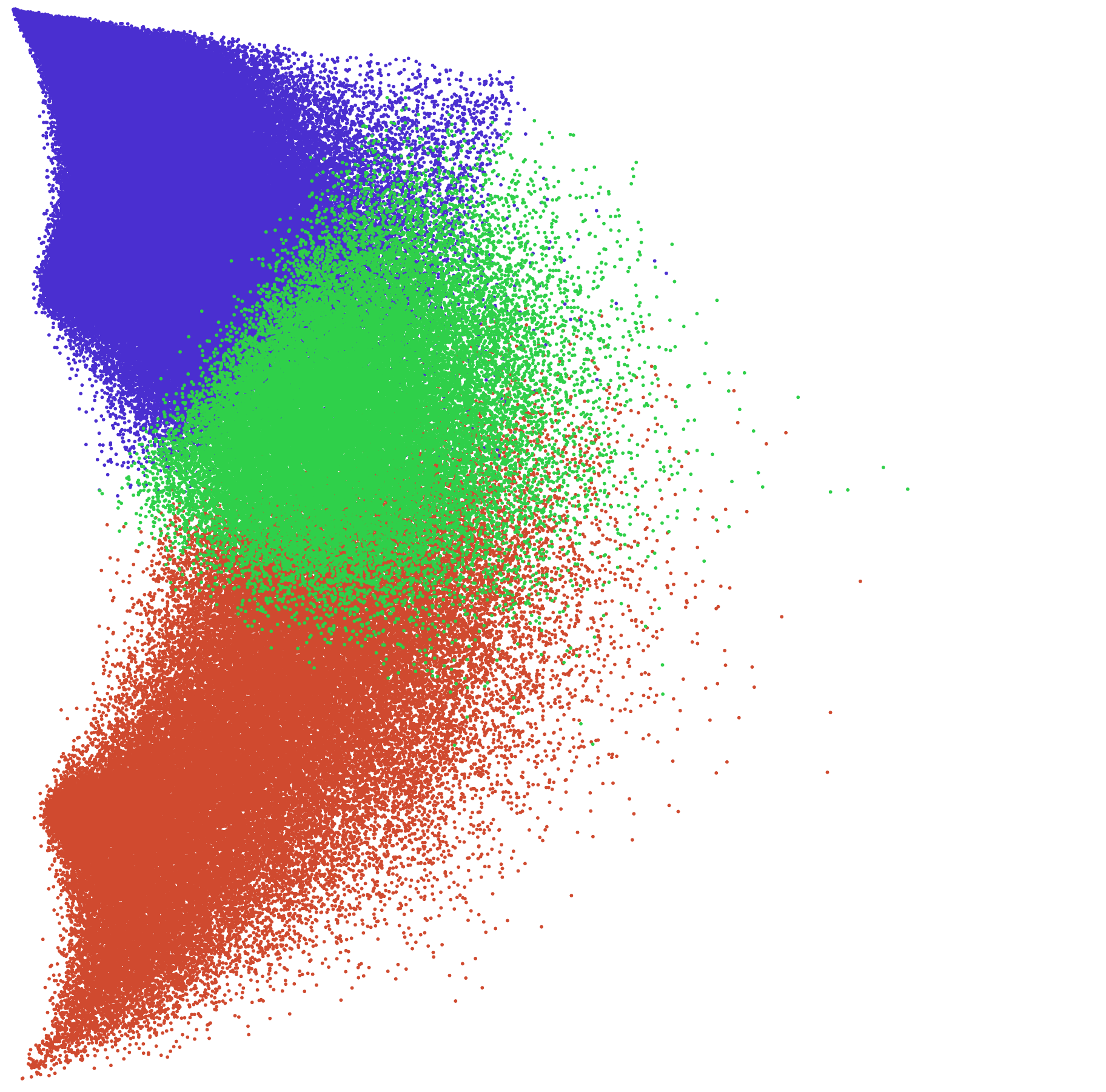}
    \caption{}
    \label{fig:carto_race_opt}
  \end{subfigure}
    \caption{
    Data map for data points of the training set of various tasks, based on OPT\textsubscript{\textrm{350M}}.
    Red dots represent data points with $\mathcal{H}_i \in \{0,1\}$, green dots represent data points with $\mathcal{H}_i \in \{2,3,4\}$, and blue dots represent data points with $\mathcal{H}_i \in \{5,6\}$.
    }
  \label{fig:opt_data_map}
\end{figure}

\begin{figure}[H]
  \centering
  \begin{subfigure}{0.48\textwidth}
    \centering
    \pgfplotsset{width=\columnwidth, height=\columnwidth}
    \input{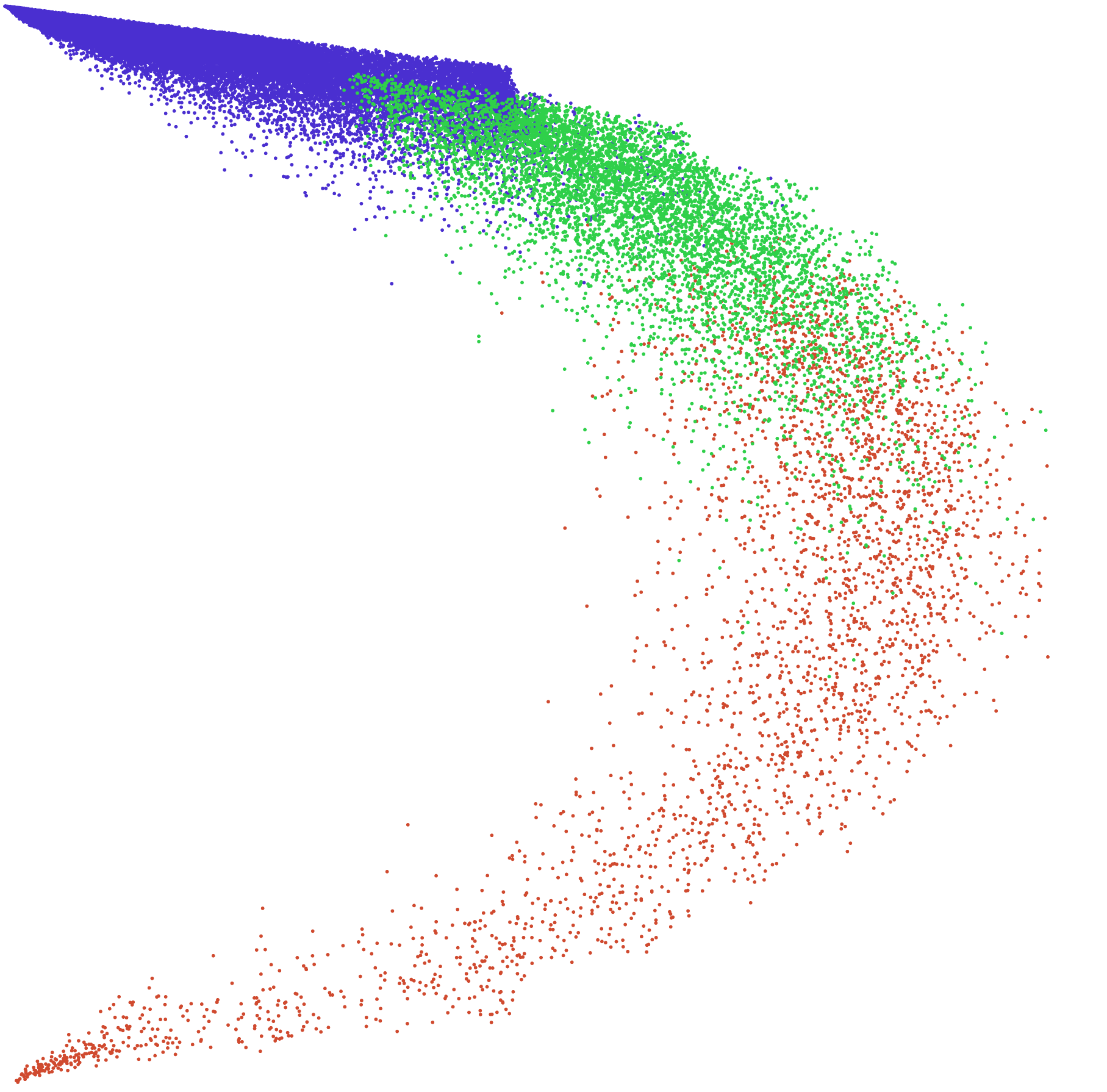}
    \caption{}
    \label{fig:carto_sst2_roberta_large}
  \end{subfigure}
  \hfill
  \begin{subfigure}{0.48\textwidth}
    \centering
    \pgfplotsset{width=\columnwidth, height=\columnwidth}
    \input{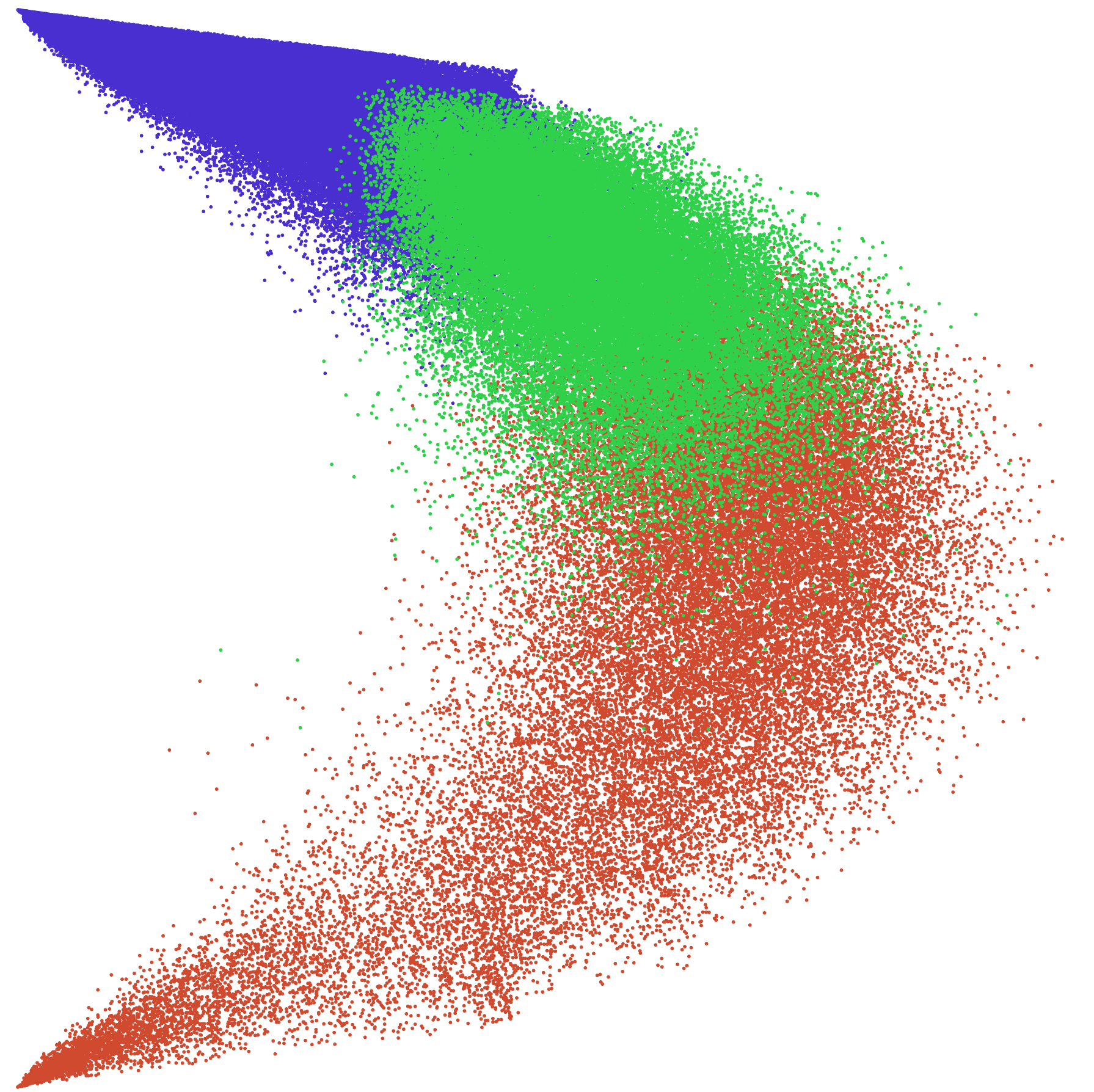}
    \caption{}
    \label{fig:carto_mnli_roberta_large}
  \end{subfigure}
  \vfill
  \centering
  \begin{subfigure}{0.48\textwidth}
    \centering
    \pgfplotsset{width=\columnwidth, height=\columnwidth}
    \input{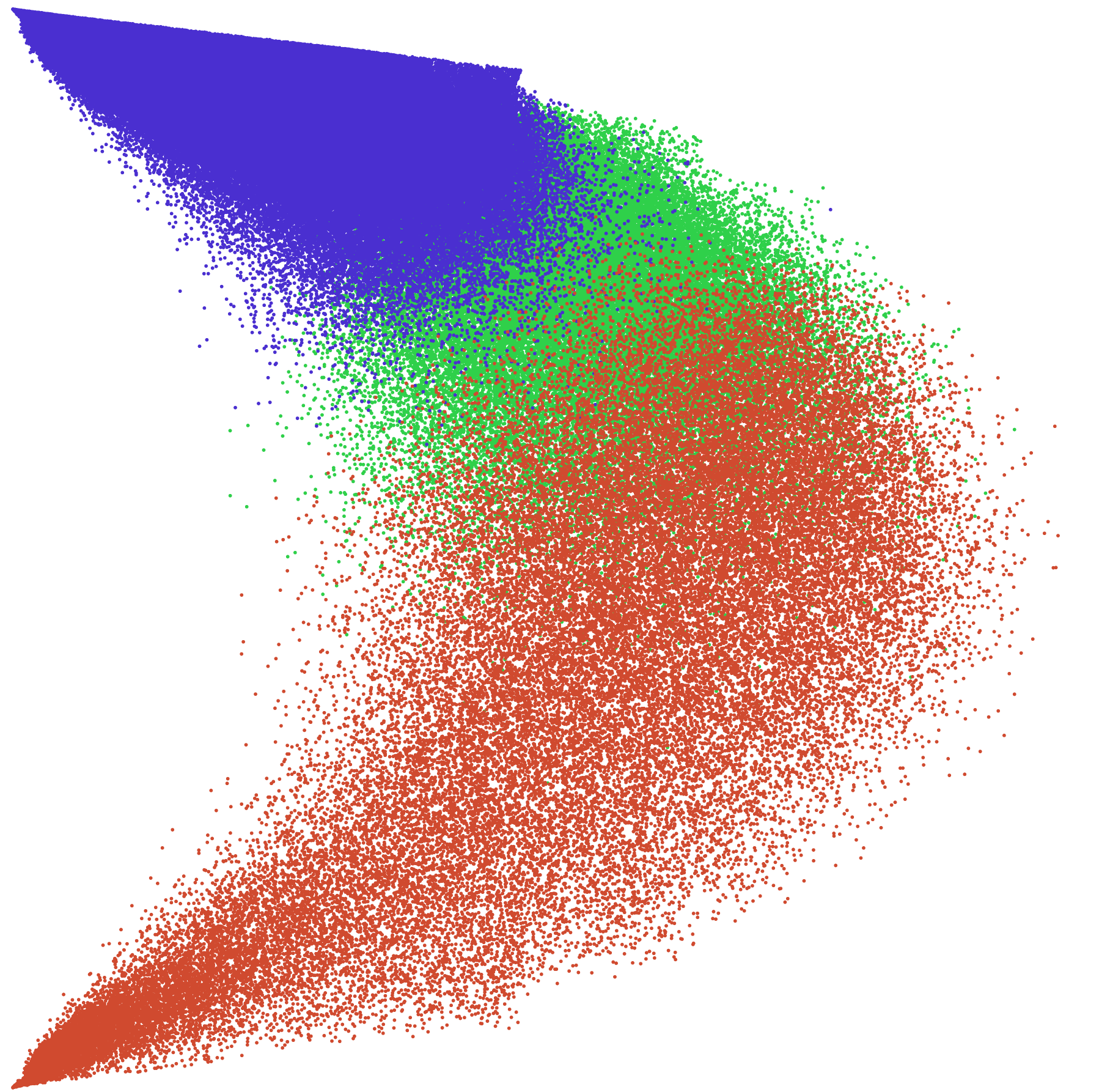}
    \caption{}
    \label{fig:carto_snli_roberta_large}
  \end{subfigure}
  \hfill
  \begin{subfigure}{0.48\textwidth}
    \centering
    \pgfplotsset{width=\columnwidth, height=\columnwidth}
    \input{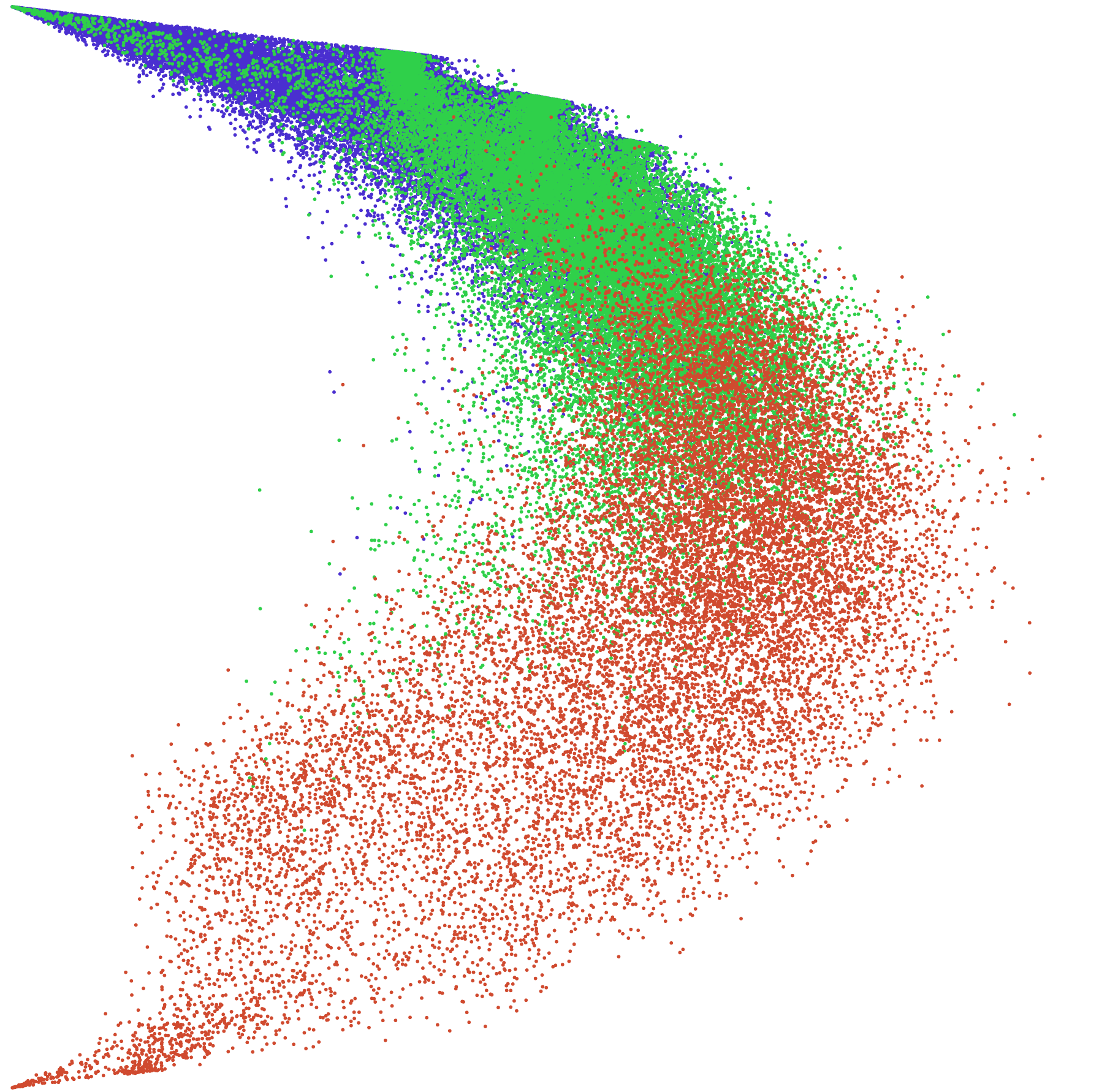}
    \caption{}
    \label{fig:carto_race_roberta_large}
  \end{subfigure}
    \caption{
    Data map for data points of the training set of various tasks, based on RoBERTa\textsubscript{\textrm{LARGE}}.
    Red dots represent data points with $\mathcal{H}_i \in \{0,1\}$, green dots represent data points with $\mathcal{H}_i \in \{2,3,4\}$, and blue dots represent data points with $\mathcal{H}_i \in \{5,6\}$.
    }
  \label{fig:roberta_large_data_map}
\end{figure}

\begin{figure}[H]
  \centering
  \begin{subfigure}{0.48\textwidth}
    \centering
    \pgfplotsset{width=\columnwidth, height=\columnwidth}
    \input{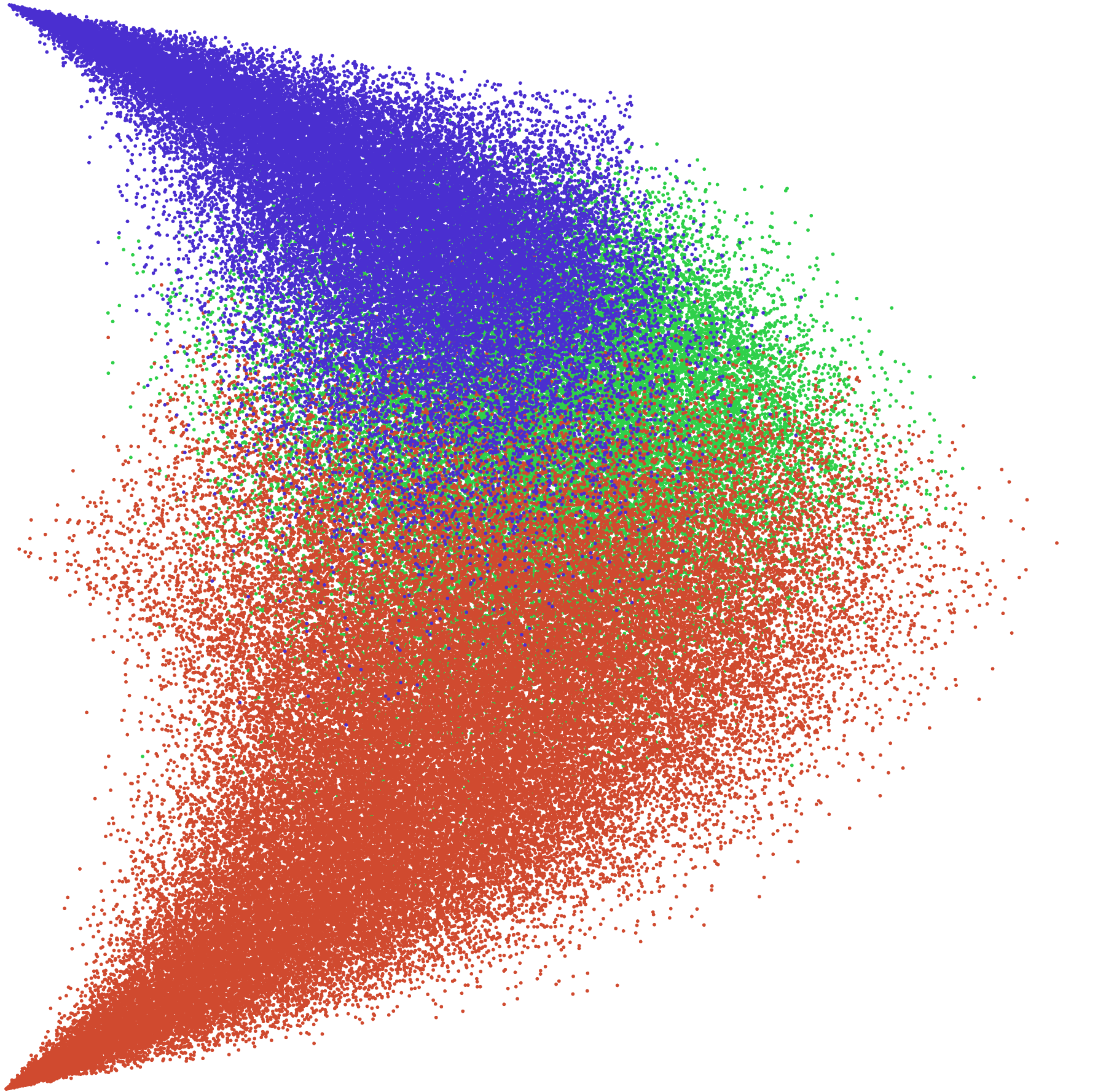}
    \caption{}
    \label{fig:carto_squad_v2_opt}
  \end{subfigure}
  \hfill
  \begin{subfigure}{0.48\textwidth}
    \centering
    \pgfplotsset{width=\columnwidth, height=\columnwidth}
    \input{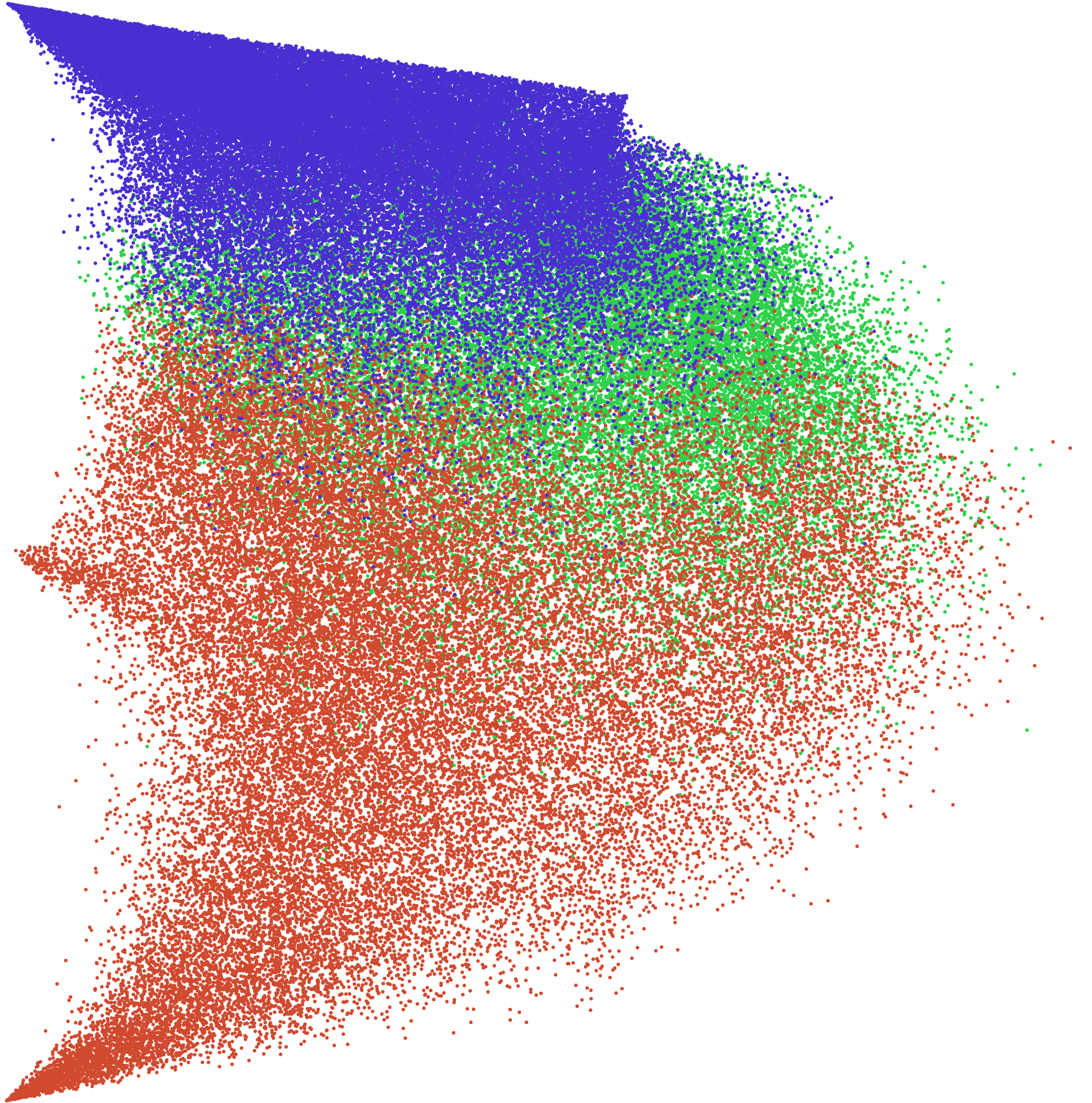}
    \caption{}
    \label{fig:carto_squad_v2_roberta_large}
  \end{subfigure}
    \caption{
    Data map for data points of the training set of SQuAD v2 task, based on OPT\textsubscript{\textrm{350M}} (left) and  RoBERTa\textsubscript{\textrm{LARGE}} (right).
    Red dots represent data points with $\mathcal{H}_i \in \{0,1\}$, green dots represent data points with $\mathcal{H}_i \in \{2,3,4\}$, and blue dots represent data points with $\mathcal{H}_i \in \{5,6\}$.
    }
  \label{fig:squad_dmaps}
\end{figure}

Similar to Figure \ref{fig:meanh}, Figure \ref{fig:more_meanh} depicts the mean $\mathcal{H}$-score against the subset size for various tasks based on RoBERTa\textsubscript{\textrm{LARGE}}.
Figures \ref{fig:opt_data_map}, \ref{fig:roberta_large_data_map}, and \ref{fig:squad_dmaps} depict the data maps for the training sets of all our tasks.
We compute the variability and confidence metric on the outputs of our fine-tuning runs, following \cite{swayamdipta-etal-2020-dataset}.
Variability is calculated as the standard deviation of the probability of the model for the golden label and confidence is its average.

In each plot, red dots represent data points with $\mathcal{H}_i \in \{0,1\}$, green dots represent data points with $\mathcal{H}_i \in \{2,3,4\}$, and blue dots represent data points with $\mathcal{H}_i \in \{5,6\}$.
\cite{swayamdipta-etal-2020-dataset} considers data points with a large variability metric as the most ambiguous.
However, we believe data points with $\mathcal{H}_i \in \{2,3,4\}$ are the most ambiguous.
Figures \ref{fig:opt_data_map}, \ref{fig:roberta_large_data_map}, and \ref{fig:squad_dmaps} clearly show the difference between our two methods.

\subsection{Comparison with Forgetting Score}
\label{sec:app:compare_forgetting}

\begin{figure}[h]
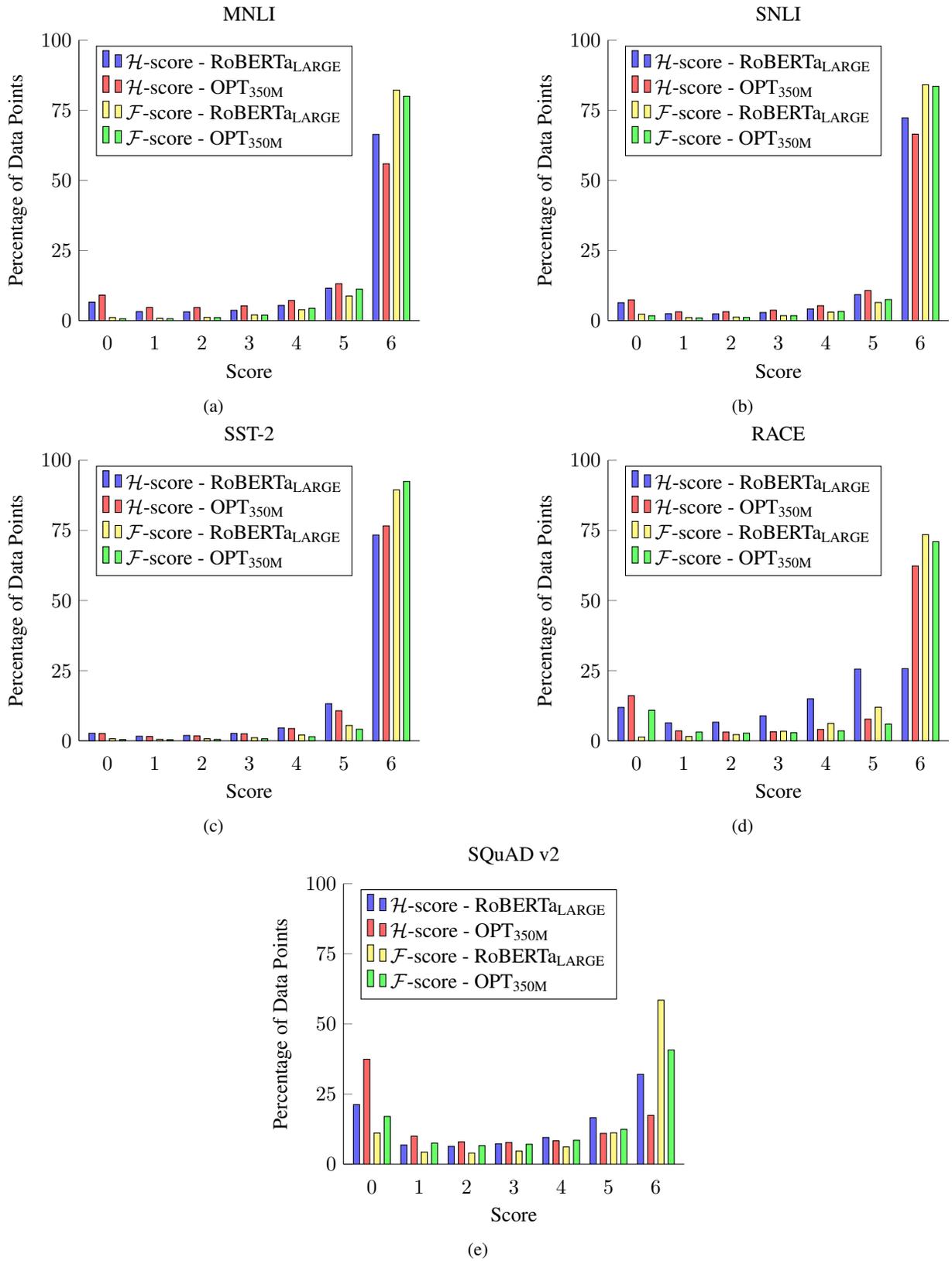

  \centering
  \begin{subfigure}{0.45\textwidth}
    \centering
    \pgfplotsset{width=\columnwidth, height=6.4cm}
    \input{fig/fscore_hist/mnli}
    \caption{}
    \label{fig:fscore_mnli_hist}
  \end{subfigure}
  \hfill
  \begin{subfigure}{0.45\textwidth}
    \centering
    \pgfplotsset{width=\columnwidth, height=6.4cm}
    \input{fig/fscore_hist/snli}
    \caption{}
    \label{fig:fscore_snli_hist}
  \end{subfigure}
  \vfill
    \begin{subfigure}{0.45\textwidth}
    \centering
    \pgfplotsset{width=\columnwidth, height=6.4cm}
    \input{fig/fscore_hist/sst2}
    \caption{}
    \label{fig:fscore_sst2_hist}
  \end{subfigure}
  \hfill
  \begin{subfigure}{0.45\textwidth}
    \centering
    \pgfplotsset{width=\columnwidth, height=6.4cm}
    \input{fig/fscore_hist/race}
    \caption{}
    \label{fig:fscore_race_hist}
  \end{subfigure}
  \vfill
  \begin{subfigure}{0.45\textwidth}
    \centering
    \pgfplotsset{width=\columnwidth, height=6.4cm}
    \input{fig/fscore_hist/squad}
    \caption{}
    \label{fig:fscore_squad_hist}
  \end{subfigure}
    \caption{
    Comparison between the distribution of the $\mathcal{H}$-score and $\mathcal{F}$-score.
    Scores are reported for the training set of the MNLI, SNLI, SST-2, RACE, and SQuAD v2 tasks, based on RoBERTa\textsubscript{\textrm{LARGE}} and OPT\textsubscript{\textrm{350M}} models.
    }
  \label{fig:fscore_hist}
\end{figure}

We introduced $\mathcal{F}$-score in Section \ref{sec:compare_forgetting} to study the main difference between our $\mathcal{H}$-score and the forgetting score \cite{toneva2018an}.
Figure \ref{fig:fscore_hist} shows the distribution of both scores side by side.
Across all tasks, $\mathcal{F}$-score reduces the number of data points with the score smaller than $6$.
This demonstrates that many of the data points that where learned at a later epoch in one of the runs, were correctly classified in all epochs in other runs.
While under $\mathcal{H}$-score these data points receive a reward of $1$ only in the latter type of runs, $\mathcal{F}$-score rewards all of them and thus leads to a score of $6$.

We use $\mathcal{F}$-score to create subsets similar to the experiments in Section \ref{sec:setup}.
Tables \ref{tab:fscore_mnli} and \ref{tab:fscore_race} compare the evaluation performance of RoBERTa\textsubscript{\textrm{LARGE}} and OPT\textsubscript{\textrm{350M}} on subsets made using $\mathcal{F}$-score with subsets made based on $\mathcal{H}$-score.
The experimental results demonstrate that using $\mathcal{F}$-score results in smaller subsets and also significant drops in the evaluation performance.

\clearpage

\begin{table}[H]
    \centering
    \begin{tabular}{cc|ccccccc}
     \multicolumn{9}{c}{\textbf{MNLI Matched} (Accuracy)} \\
     
    \cmidrule[1.5pt](l{\abovetopsep}r{\belowbottomsep}){1-9}
    
     & & $\mathcal{D}_{\{4\}}$ & $\mathcal{D}_{\{5\}}$ & $\mathcal{D}_{\{2,3,4\}}$ & $\mathcal{D}_{\{4,5\}}$ & $\mathcal{D}_{\{3,4,5\}}$ & $\mathcal{D}_{\{2,3,4,5\}}$ & $\mathcal{D}_{\{1,2,3,4,5\}}$
      \\
    \cmidrule[0.2pt](l{\tabcolsep}r{\tabcolsep}){1-9}
    
    \parbox[t]{5mm}{\multirow{5}{*}{\rotatebox[origin=c]{90}{\begin{tabular}[c]{@{}c@{}}RoBERTa\\Large\end{tabular}}}} & 
    \multirow{2}{*}{$\mathcal{H}$-score} & 
    $88.68_{\hspace{1pt}0.50}$ & $89.48_{\hspace{1pt}0.09}$ & $89.91_{\hspace{1pt}0.24}$ & $90.06_{\hspace{1pt}0.13}$ & $90.23_{\hspace{1pt}0.11}$ & $90.36_{\hspace{1pt}0.09}$ & $90.57_{\hspace{1pt}0.12}$
    \\[1pt]
    & &
    $(5.42\%)$ & $(11.58\%)$ & $(12.27\%)$ & $(17.0\%)$ & $(20.7\%)$ & $(23.85\%)$ & $(27.06\%)$
    \\[5pt]
    &
    \multirow{2}{*}{$\mathcal{F}$-score} & 
    $79.77_{\hspace{1pt}0.47}$ & $88.34_{\hspace{1pt}0.14}$ & $43.53_{\hspace{1pt}11.39}$ & $87.54_{\hspace{1pt}1.54}$ & $87.86_{\hspace{1pt}0.43}$ & $50.49_{\hspace{1pt}32.31}$ & $76.45_{\hspace{1pt}12.16}$
    \\[1pt]
    & &
    $(3.92\%)$ & $(8.8\%)$ & $(7.14\%)$ & $(12.72\%)$ & $(14.77\%)$ & $(15.94\%)$ & $(16.73\%)$
    \\

    \cmidrule[1pt](l{\abovetopsep}r{\belowbottomsep}){1-9}

    \parbox[t]{5mm}{\multirow{5}{*}{\rotatebox[origin=c]{90}{\begin{tabular}[c]{@{}c@{}}OPT\\350M\end{tabular}}}} & 
    \multirow{2}{*}{$\mathcal{H}$-score} & 
    $80.87_{\hspace{1pt}0.39}$ & $81.87_{\hspace{1pt}0.14}$ & $81.99_{\hspace{1pt}0.44}$ & $82.37_{\hspace{1pt}0.19}$ & $83.23_{\hspace{1pt}0.14}$ & $83.38_{\hspace{1pt}0.11}$ & $83.53_{\hspace{1pt}0.37}$
    \\[1pt]
    &&
    $(7.18\%)$ & $(13.14\%)$ & $(17.14\%)$ & $(20.32\%)$ & $(25.62\%)$ & $(30.28\%)$ & $(34.97\%)$
    \\[5pt]
    &
    \multirow{2}{*}{$\mathcal{F}$-score} & 
    $60.09_{\hspace{1pt}2.01}$ & $77.65_{\hspace{1pt}0.26}$ & $51.27_{\hspace{1pt}1.81}$ & $76.20_{\hspace{1pt}0.29}$ & $75.15_{\hspace{1pt}0.67}$ & $73.30_{\hspace{1pt}0.31}$ & $69.95_{\hspace{1pt}1.59}$
    \\[1pt]
    &&
    $(4.38\%)$ & $(11.22\%)$ & $(7.43\%)$ & $(15.61\%)$ & $(17.59\%)$ & $(18.65\%)$ & $(19.34\%)$
    \\
    
    \cmidrule[1.5pt](l{\abovetopsep}r{\belowbottomsep}){1-9}
     \multicolumn{9}{c}{\textbf{MNLI Mismatched} (Accuracy)} \\
    \cmidrule[1.5pt](l{\abovetopsep}r{\belowbottomsep}){1-9}

     & & $\mathcal{D}_{\{4\}}$ & $\mathcal{D}_{\{5\}}$ & $\mathcal{D}_{\{2,3,4\}}$ & $\mathcal{D}_{\{4,5\}}$ & $\mathcal{D}_{\{3,4,5\}}$ & $\mathcal{D}_{\{2,3,4,5\}}$ & $\mathcal{D}_{\{1,2,3,4,5\}}$
      \\
    \cmidrule[0.2pt](l{\tabcolsep}r{\tabcolsep}){1-9}
    
    \parbox[t]{5mm}{\multirow{5}{*}{\rotatebox[origin=c]{90}{\begin{tabular}[c]{@{}c@{}}RoBERTa\\Large\end{tabular}}}} & 
    \multirow{2}{*}{$\mathcal{H}$-score} & 
    $88.79_{\hspace{1pt}0.35}$ & $89.26_{\hspace{1pt}0.25}$ & $89.61_{\hspace{1pt}0.18}$ & $89.81_{\hspace{1pt}0.14}$ & $90.05_{\hspace{1pt}0.04}$ & $90.12_{\hspace{1pt}0.31}$ & $90.22_{\hspace{1pt}0.05}$
    \\[1pt]
    & &
    $(5.42\%)$ & $(11.58\%)$ & $(12.27\%)$ & $(17.0\%)$ & $(20.7\%)$ & $(23.85\%)$ & $(27.06\%)$
    \\[5pt]
    &
    \multirow{2}{*}{$\mathcal{F}$-score} & 
    $80.19_{\hspace{1pt}0.47}$ & $88.19_{\hspace{1pt}0.14}$ & $43.95_{\hspace{1pt}11.39}$ & $87.53_{\hspace{1pt}1.54}$ & $87.93_{\hspace{1pt}0.43}$ & $50.48_{\hspace{1pt}32.31}$ & $77.50_{\hspace{1pt}12.16}$
    \\[1pt]
    & &
    $(3.92\%)$ & $(8.8\%)$ & $(7.14\%)$ & $(12.72\%)$ & $(14.77\%)$ & $(15.94\%)$ & $(16.73\%)$
    \\

    \cmidrule[1pt](l{\abovetopsep}r{\belowbottomsep}){1-9}

    \parbox[t]{5mm}{\multirow{5}{*}{\rotatebox[origin=c]{90}{\begin{tabular}[c]{@{}c@{}}OPT\\350M\end{tabular}}}} & 
    \multirow{2}{*}{$\mathcal{H}$-score} & 
    $81.83_{\hspace{1pt}0.35}$ & $82.56_{\hspace{1pt}0.25}$ & $82.85_{\hspace{1pt}0.18}$ & $83.55_{\hspace{1pt}0.14}$ & $83.70_{\hspace{1pt}0.04}$ & $84.15_{\hspace{1pt}0.31}$ & $84.29_{\hspace{1pt}0.05}$
    \\[1pt]
    &&
    $(7.18\%)$ & $(13.14\%)$ & $(17.14\%)$ & $(20.32\%)$ & $(25.62\%)$ & $(30.28\%)$ & $(34.97\%)$
    \\[5pt]
    &
    \multirow{2}{*}{$\mathcal{F}$-score} & 
    $61.85_{\hspace{1pt}1.63}$ & $77.99_{\hspace{1pt}0.25}$ & $52.52_{\hspace{1pt}2.13}$ & $77.26_{\hspace{1pt}0.22}$ & $75.68_{\hspace{1pt}0.53}$ & $73.60_{\hspace{1pt}0.25}$ & $70.91_{\hspace{1pt}1.20}$
    \\[1pt]
    &&
    $(4.38\%)$ & $(11.22\%)$ & $(7.43\%)$ & $(15.61\%)$ & $(17.59\%)$ & $(18.65\%)$ & $(19.34\%)$
    \\
    \cmidrule[1.5pt](l{\abovetopsep}r{\belowbottomsep}){1-9}

    \multicolumn{9}{c}{\textbf{SNLI} (Accuracy)} \\
     
    \cmidrule[1.5pt](l{\abovetopsep}r{\belowbottomsep}){1-9}
    
     & & $\mathcal{D}_{\{4\}}$ & $\mathcal{D}_{\{5\}}$ & $\mathcal{D}_{\{2,3,4\}}$ & $\mathcal{D}_{\{4,5\}}$ & $\mathcal{D}_{\{3,4,5\}}$ & $\mathcal{D}_{\{2,3,4,5\}}$ & $\mathcal{D}_{\{1,2,3,4,5\}}$
      \\
    \cmidrule[0.2pt](l{\tabcolsep}r{\tabcolsep}){1-9}
    
    \parbox[t]{5mm}{\multirow{5}{*}{\rotatebox[origin=c]{90}{\begin{tabular}[c]{@{}c@{}}RoBERTa\\Large\end{tabular}}}} & 
    \multirow{2}{*}{$\mathcal{H}$-score} & 
    $91.41_{\hspace{1pt}0.31}$ & $91.84_{\hspace{1pt}0.15}$ & $91.66_{\hspace{1pt}0.16}$ & $92.13_{\hspace{1pt}0.22}$ & $92.38_{\hspace{1pt}0.03}$ & $92.41_{\hspace{1pt}0.03}$ & $92.50_{\hspace{1pt}0.14}$
    \\[1pt]
    & &
    $(4.2\%)$ & $(9.27\%)$ & $(9.52\%)$ & $(13.47\%)$ & $(16.39\%)$ & $(18.8\%)$ & $(21.26\%)$
    \\[5pt]
    &
    \multirow{2}{*}{$\mathcal{F}$-score} & 
    $32.77_{\hspace{1pt}0.00}$ & $71.70_{\hspace{1pt}33.71}$ & $46.82_{\hspace{1pt}24.34}$ & $32.77_{\hspace{1pt}0.00}$ & $52.14_{\hspace{1pt}33.56}$ & $42.51_{\hspace{1pt}16.86}$ & $51.83_{\hspace{1pt}33.02}$
    \\[1pt]
    & &
    $(3.04\%)$ & $(6.48\%)$ & $(6.08\%)$ & $(9.52\%)$ & $(11.3\%)$ & $(12.55\%)$ & $(13.62\%)$
    \\

    \cmidrule[1pt](l{\abovetopsep}r{\belowbottomsep}){1-9}

    \parbox[t]{5mm}{\multirow{5}{*}{\rotatebox[origin=c]{90}{\begin{tabular}[c]{@{}c@{}}OPT\\350M\end{tabular}}}} & 
    \multirow{2}{*}{$\mathcal{H}$-score} & 
    $87.34_{\hspace{1pt}0.12}$ & $88.44_{\hspace{1pt}0.03}$ & $88.21_{\hspace{1pt}0.32}$ & $89.13_{\hspace{1pt}0.13}$ & $89.58_{\hspace{1pt}0.11}$ & $89.87_{\hspace{1pt}0.14}$ & $90.07_{\hspace{1pt}0.29}$
    \\[1pt]
    &&
    $(5.34\%)$ & $(10.71\%)$ & $(12.26\%)$ & $(16.06\%)$ & $(19.79\%)$ & $(22.97\%)$ & $(26.13\%)$
    \\[5pt]
    &
    \multirow{2}{*}{$\mathcal{F}$-score} & 
    $70.56_{\hspace{1pt}0.71}$ & $84.32_{\hspace{1pt}0.07}$ & $53.86_{\hspace{1pt}1.32}$ & $84.07_{\hspace{1pt}0.21}$ & $82.86_{\hspace{1pt}0.44}$ & $80.86_{\hspace{1pt}0.17}$ & $77.03_{\hspace{1pt}0.33}$
    \\[1pt]
    &&
    $(3.25\%)$ & $(7.54\%)$ & $(6.19\%)$ & $(10.79\%)$ & $(12.57\%)$ & $(13.73\%)$ & $(14.7\%)$
    \\
    
    \cmidrule[1.5pt](l{\abovetopsep}r{\belowbottomsep}){1-9}
     \multicolumn{9}{c}{\textbf{SST-2} (Accuracy)} \\
    \cmidrule[1.5pt](l{\abovetopsep}r{\belowbottomsep}){1-9}

     & & $\mathcal{D}_{\{4\}}$ & $\mathcal{D}_{\{2,3,4\}}$ & $\mathcal{D}_{\{5\}}$ & $\mathcal{D}_{\{4,5\}}$ & $\mathcal{D}_{\{3,4,5\}}$ & $\mathcal{D}_{\{2,3,4,5\}}$ & $\mathcal{D}_{\{1,2,3,4,5\}}$
      \\
    \cmidrule[0.2pt](l{\tabcolsep}r{\tabcolsep}){1-9}
    
    \parbox[t]{5mm}{\multirow{5}{*}{\rotatebox[origin=c]{90}{\begin{tabular}[c]{@{}c@{}}RoBERTa\\Large\end{tabular}}}} & 
    \multirow{2}{*}{$\mathcal{H}$-score} & 
    $94.15_{\hspace{1pt}0.35}$ & $94.65_{\hspace{1pt}0.46}$ & $94.99_{\hspace{1pt}0.17}$ & $95.45_{\hspace{1pt}0.24}$ & $95.53_{\hspace{1pt}0.42}$ & $95.87_{\hspace{1pt}0.23}$ & $95.80_{\hspace{1pt}0.37}$
    \\[1pt]
    & &
    $(4.6\%)$ & $(9.11\%)$ & $(13.21\%)$ & $(17.81\%)$ & $(20.41\%)$ & $(22.32\%)$ & $(23.95\%)$
    \\[5pt]
    &
    \multirow{2}{*}{$\mathcal{F}$-score} & 
    $55.08_{\hspace{1pt}5.22}$ & $50.92_{\hspace{1pt}0.00}$ & $63.72_{\hspace{1pt}22.18}$ & $93.31_{\hspace{1pt}0.82}$ & $63.53_{\hspace{1pt}21.85}$ & $54.89_{\hspace{1pt}6.88}$ & $64.76_{\hspace{1pt}23.97}$
    \\[1pt]
    & &
    $(2.04\%)$ & $(3.86\%)$ & $(5.47\%)$ & $(7.51\%)$ & $(8.61\%)$ & $(9.33\%)$ & $(9.84\%)$
    \\

    \cmidrule[1pt](l{\abovetopsep}r{\belowbottomsep}){1-9}

    \parbox[t]{5mm}{\multirow{5}{*}{\rotatebox[origin=c]{90}{\begin{tabular}[c]{@{}c@{}}OPT\\350M\end{tabular}}}} & 
    \multirow{2}{*}{$\mathcal{H}$-score} & 
    $91.67_{\hspace{1pt}0.33}$ & $90.98_{\hspace{1pt}0.35}$ & $93.19_{\hspace{1pt}0.18}$ & $93.81_{\hspace{1pt}0.12}$ & $93.73_{\hspace{1pt}0.07}$ & $94.19_{\hspace{1pt}0.29}$ & $94.04_{\hspace{1pt}0.12}$
    \\[1pt]
    &&
    $(4.37\%)$ & $(8.58\%)$ & $(10.71\%)$ & $(15.08\%)$ & $(17.57\%)$ & $(19.29\%)$ & $(20.82\%)$
    \\[5pt]
    &
    \multirow{2}{*}{$\mathcal{F}$-score} & 
    $68.77_{\hspace{1pt}9.20}$ & $49.16_{\hspace{1pt}12.11}$ & $88.61_{\hspace{1pt}1.40}$ & $88.34_{\hspace{1pt}0.70}$ & $87.35_{\hspace{1pt}1.62}$ & $84.82_{\hspace{1pt}1.34}$ & $80.70_{\hspace{1pt}2.20}$
    \\[1pt]
    &&
    $(1.43\%)$ & $(2.64\%)$ & $(4.12\%)$ & $(5.55\%)$ & $(6.26\%)$ & $(6.76\%)$ & $(7.13\%)$
    \\
    
\end{tabular}

    \caption{
    Evaluation accuracy of fine-tuning RoBERTa\textsubscript{\textrm{LARGE}} and OPT\textsubscript{\textrm{350M}} on subsets created with $\mathcal{H}$-score and $\mathcal{F}$-score.
    The size of each subset is provided in parenthesis as a percentage of the original dataset size.
    Each reported accuracy is the average of 3 runs with different intialization seeds.
    The standard deviation of the runs is also reported as the subscript.
    }
    \label{tab:fscore_mnli}
\end{table}

\begin{table}[H]
    \centering
    \begin{tabular}{cc|ccccccc}
     \multicolumn{9}{c}{\textbf{RACE} (Accuracy)} \\
     
    \cmidrule[1.5pt](l{\abovetopsep}r{\belowbottomsep}){1-9}
    
     & & $\mathcal{D}_{\{4\}}$ & $\mathcal{D}_{\{5\}}$ & $\mathcal{D}_{\{2,3,4\}}$ & $\mathcal{D}_{\{4,5\}}$ & $\mathcal{D}_{\{3,4,5\}}$ & $\mathcal{D}_{\{2,3,4,5\}}$ & $\mathcal{D}_{\{1,2,3,4,5\}}$
      \\
    \cmidrule[0.2pt](l{\tabcolsep}r{\tabcolsep}){1-9}
    
    \parbox[t]{5mm}{\multirow{5}{*}{\rotatebox[origin=c]{90}{\begin{tabular}[c]{@{}c@{}}RoBERTa\\Large\end{tabular}}}} & 
    \multirow{2}{*}{$\mathcal{H}$-score} & 
    $77.79_{\hspace{1pt}0.39}$ & $78.73_{\hspace{1pt}0.25}$ & $80.56_{\hspace{1pt}0.73}$ & $80.62_{\hspace{1pt}0.12}$ & $81.36_{\hspace{1pt}0.43}$ & $82.36_{\hspace{1pt}0.54}$ & $82.95_{\hspace{1pt}0.13}$
    \\[1pt]
    & &
    $(14.97)\%$ & $(25.56)\%$ & $(30.47)\%$ & $(40.53)\%$ & $(49.39)\%$ & $(56.03)\%$ & $(62.39)\%$
    \\[5pt]
    &
    \multirow{2}{*}{$\mathcal{F}$-score} & 
    $28.54_{\hspace{1pt}2.00}$ & $23.75_{\hspace{1pt}1.22}$ & $23.64_{\hspace{1pt}0.15}$ & $40.22_{\hspace{1pt}29.11}$ & $24.22_{\hspace{1pt}1.43}$ & $24.94_{\hspace{1pt}0.08}$ & $24.54_{\hspace{1pt}1.18}$
    \\[1pt]
    & &
    $(6.15)\%$ & $(11.94)\%$ & $(11.73)\%$ & $(18.1)\%$ & $(21.47)\%$ & $(23.67)\%$ & $(25.18)\%$
    \\

    \cmidrule[1pt](l{\abovetopsep}r{\belowbottomsep}){1-9}

    \parbox[t]{5mm}{\multirow{5}{*}{\rotatebox[origin=c]{90}{\begin{tabular}[c]{@{}c@{}}OPT\\350M\end{tabular}}}} & 
    \multirow{2}{*}{$\mathcal{H}$-score} & 
    $76.64_{\hspace{1pt}0.42}$ & $77.60_{\hspace{1pt}0.25}$ & $74.59_{\hspace{1pt}0.20}$ & $78.50_{\hspace{1pt}0.10}$ & $78.85_{\hspace{1pt}0.06}$ & $79.26_{\hspace{1pt}0.25}$ & $79.68_{\hspace{1pt}0.09}$
    \\[1pt]
    & &
    $(4.06)\%$ & $(7.71)\%$ & $(10.35)\%$ & $(11.77)\%$ & $(14.98)\%$ & $(18.06)\%$ & $(21.62)\%$
    \\[5pt]
    &
    \multirow{2}{*}{$\mathcal{F}$-score} & 
    $56.91_{\hspace{1pt}13.41}$ & $74.07_{\hspace{1pt}1.66}$ & $32.53_{\hspace{1pt}0.74}$ & $71.39_{\hspace{1pt}2.08}$ & $62.16_{\hspace{1pt}8.99}$ & $48.22_{\hspace{1pt}10.33}$ & $37.81_{\hspace{1pt}1.89}$
    \\[1pt]
    & &
    $(3.55)\%$ & $(5.95)\%$ & $(9.14)\%$ & $(9.5)\%$ & $(12.36)\%$ & $(15.09)\%$ & $(18.18)\%$
    \\
    
    \cmidrule[1.5pt](l{\abovetopsep}r{\belowbottomsep}){1-9}
     \multicolumn{9}{c}{\textbf{SQuAD v2} (Exact Match)} \\
    \cmidrule[1.5pt](l{\abovetopsep}r{\belowbottomsep}){1-9}

     & & $\mathcal{D}_{\{4\}}$ & $\mathcal{D}_{\{5\}}$ & $\mathcal{D}_{\{2,3,4\}}$ & $\mathcal{D}_{\{4,5\}}$ & $\mathcal{D}_{\{3,4,5\}}$ & $\mathcal{D}_{\{2,3,4,5\}}$ & $\mathcal{D}_{\{1,2,3,4,5\}}$
      \\
    \cmidrule[0.2pt](l{\tabcolsep}r{\tabcolsep}){1-9}
    
    \parbox[t]{5mm}{\multirow{5}{*}{\rotatebox[origin=c]{90}{\begin{tabular}[c]{@{}c@{}}RoBERTa\\Large\end{tabular}}}} & 
    \multirow{2}{*}{$\mathcal{H}$-score} & 
    $82.13_{\hspace{1pt}0.20}$ & $81.40_{\hspace{1pt}0.20}$ & $84.03_{\hspace{1pt}0.34}$ & $82.84_{\hspace{1pt}0.41}$ & $83.80_{\hspace{1pt}0.22}$ & $84.39_{\hspace{1pt}0.25}$ & $85.09_{\hspace{1pt}0.14}$
    \\[1pt]
    & &
    $(9.54)\%$ & $(16.61)\%$ & $(23.2)\%$ & $(26.15)\%$ & $(33.44)\%$ & $(39.81)\%$ & $(46.66)\%$
    \\[5pt]
    &
    \multirow{2}{*}{$\mathcal{F}$-score} & 
    $68.05_{\hspace{1pt}7.13}$ & $81.76_{\hspace{1pt}0.28}$ & $68.60_{\hspace{1pt}2.39}$ & $82.74_{\hspace{1pt}0.52}$ & $82.97_{\hspace{1pt}0.21}$ & $82.78_{\hspace{1pt}0.49}$ & $82.22_{\hspace{1pt}0.31}$
    \\[1pt]
    & &
    $(6.18)\%$ & $(11.2)\%$ & $(14.85)\%$ & $(17.38)\%$ & $(22.05)\%$ & $(26.05)\%$ & $(30.38)\%$
    \\

    \cmidrule[1pt](l{\abovetopsep}r{\belowbottomsep}){1-9}

    \parbox[t]{5mm}{\multirow{5}{*}{\rotatebox[origin=c]{90}{\begin{tabular}[c]{@{}c@{}}OPT\\350M\end{tabular}}}} & 
    \multirow{2}{*}{$\mathcal{H}$-score} & 
    $59.75_{\hspace{1pt}0.34}$ & $59.54_{\hspace{1pt}0.17}$ & $61.00_{\hspace{1pt}0.14}$ & $62.50_{\hspace{1pt}0.45}$ & $61.77_{\hspace{1pt}0.21}$ & $62.86_{\hspace{1pt}0.12}$ & $63.72_{\hspace{1pt}0.38}$
    \\[1pt]
    & &
    $(8.38)\%$ & $(10.99)\%$ & $(19.38)\%$ & $(24.10)\%$ & $(27.12)\%$ & $(35.09)\%$ & $(45.13)\%$
    \\[5pt]
    &
    \multirow{2}{*}{$\mathcal{F}$-score} & 
    $54.73_{\hspace{1pt}1.23}$ & $59.59_{\hspace{1pt}0.57}$ & $54.84_{\hspace{1pt}0.07}$ & $61.32_{\hspace{1pt}0.31}$ & $61.76_{\hspace{1pt}0.22}$ & $61.21_{\hspace{1pt}0.60}$ & $60.55_{\hspace{1pt}0.20}$
    \\[1pt]
    & &
    $(8.53)\%$ & $(12.42)\%$ & $(22.29)\%$ & $(20.95)\%$ & $(28.08)\%$ & $(34.71)\%$ & $(42.26)\%$
    \\
    \cmidrule[1.5pt](l{\abovetopsep}r{\belowbottomsep}){1-9}

    \multicolumn{9}{c}{\textbf{SQuAD v2} (F1 Score)} \\
     
    \cmidrule[1.5pt](l{\abovetopsep}r{\belowbottomsep}){1-9}
    
     & & $\mathcal{D}_{\{4\}}$ & $\mathcal{D}_{\{5\}}$ & $\mathcal{D}_{\{2,3,4\}}$ & $\mathcal{D}_{\{4,5\}}$ & $\mathcal{D}_{\{3,4,5\}}$ & $\mathcal{D}_{\{2,3,4,5\}}$ & $\mathcal{D}_{\{1,2,3,4,5\}}$
      \\
    \cmidrule[0.2pt](l{\tabcolsep}r{\tabcolsep}){1-9}
    
    \parbox[t]{5mm}{\multirow{5}{*}{\rotatebox[origin=c]{90}{\begin{tabular}[c]{@{}c@{}}RoBERTa\\Large\end{tabular}}}} & 
    \multirow{2}{*}{$\mathcal{H}$-score} & 
    $84.94_{\hspace{1pt}0.07}$ & $84.06_{\hspace{1pt}0.28}$ & $86.62_{\hspace{1pt}0.26}$ & $85.42_{\hspace{1pt}0.45}$ & $86.42_{\hspace{1pt}0.30}$ & $86.91_{\hspace{1pt}0.28}$ & $87.70_{\hspace{1pt}0.17}$
    \\[1pt]
    & &
    $(9.54)\%$ & $(16.61)\%$ & $(23.2)\%$ & $(26.15)\%$ & $(33.44)\%$ & $(39.81)\%$ & $(46.66)\%$
    \\[5pt]
    &
    \multirow{2}{*}{$\mathcal{F}$-score} & 
    $71.11_{\hspace{1pt}6.90}$ & $84.59_{\hspace{1pt}0.31}$ & $71.69_{\hspace{1pt}2.32}$ & $85.40_{\hspace{1pt}0.48}$ & $85.70_{\hspace{1pt}0.27}$ & $85.53_{\hspace{1pt}0.51}$ & $85.17_{\hspace{1pt}0.41}$
    \\[1pt]
    & &
    $(6.18)\%$ & $(11.2)\%$ & $(14.85)\%$ & $(17.38)\%$ & $(22.05)\%$ & $(26.05)\%$ & $(30.38)\%$
    \\

    \cmidrule[1pt](l{\abovetopsep}r{\belowbottomsep}){1-9}

    \parbox[t]{5mm}{\multirow{5}{*}{\rotatebox[origin=c]{90}{\begin{tabular}[c]{@{}c@{}}OPT\\350M\end{tabular}}}} & 
    \multirow{2}{*}{$\mathcal{H}$-score} & 
    $62.19_{\hspace{1pt}0.19}$ & $61.87_{\hspace{1pt}0.20}$ & $63.47_{\hspace{1pt}0.10}$ & $65.14_{\hspace{1pt}0.34}$ & $64.30_{\hspace{1pt}0.31}$ & $65.44_{\hspace{1pt}0.13}$ & $66.49_{\hspace{1pt}0.22}$
    \\[1pt]
    &&
    $(8.38)\%$ & $(10.99)\%$ & $(19.38)\%$ & $(24.10)\%$ & $(27.12)\%$ & $(35.09)\%$ & $(45.13)\%$
    \\[5pt]
    &
    \multirow{2}{*}{$\mathcal{F}$-score} & 
    $58.10_{\hspace{1pt}1.17}$ & $62.56_{\hspace{1pt}0.45}$ & $58.67_{\hspace{1pt}0.09}$ & $64.31_{\hspace{1pt}0.29}$ & $64.90_{\hspace{1pt}0.24}$ & $64.56_{\hspace{1pt}0.64}$ & $63.93_{\hspace{1pt}0.12}$
    \\[1pt]
    & &
    $(8.53)\%$ & $(12.42)\%$ & $(22.29)\%$ & $(20.95)\%$ & $(28.08)\%$ & $(34.71)\%$ & $(42.26)\%$
    \\
           
\end{tabular}

    \caption{
    Evaluation accuracy of fine-tuning RoBERTa\textsubscript{\textrm{LARGE}} and OPT\textsubscript{\textrm{350M}} on subsets created with $\mathcal{H}$-score and $\mathcal{F}$-score.
    The size of each subset is provided in parenthesis as a percentage of the original dataset size.
    Each reported accuracy is the average of 3 runs with different intialization seeds.
    The standard deviation of the runs is also reported as the subscript.
    }
    \label{tab:fscore_race}
\end{table}

\subsection{Sensitivity Analysis on $S$ and $E$ Hyperparameters}
\label{sec:app:ablation_nbruns}

We study the effect of number of runs, $S$, and number of epochs, $E$, on our pruned subsets.
We run fine-tuning experiments on subsets created with $S \in \{2,3,4,5,6\}$ and $E \in \{1,2,3\}$.
Note that our main experimental setup in Section \ref{sec:setup} is done with $S = 6$ and $E = 3$\footnote{For SQuAD v2, $E = 2$}.
Table \ref{tab:ablation_s} reports the analysis on $S$ and Tables \ref{tab:ablation_e1} and \ref{tab:ablation_e2} report the analysis on $E$.

\begin{table}[H]
    \centering
    \begin{tabular}{cc|ccccc}
    
    Task & Model & $S=2$ & $S=3$ & $S=4$ & $S=5$ & $S=6$
      \\
    \cmidrule[0.2pt](l{\tabcolsep}r{\tabcolsep}){1-7}
    
    \multirow{5}{*}{\begin{tabular}[c]{@{}c@{}}\textbf{MNLI Matched}\\Accuracy\end{tabular}} &
    \multirow{2}{*}{RoBERTa\textsubscript{\textrm{LARGE}}} & 
    $90.04_{\hspace{1pt}0.24}$ & $89.50_{\hspace{1pt}0.96}$ & $83.52_{\hspace{1pt}11.75}$ & $90.26_{\hspace{1pt}0.11}$ & $90.57_{\hspace{1pt}0.12}$
    \\[1pt]
    & &
    $(11.61\%)$ & $(17.62\%)$ & $(21.56\%)$ & $(24.61\%)$ & $(27.06\%)$
    \\[6pt]

    & \multirow{2}{*}{OPT\textsubscript{\textrm{350M}}} & 
    $82.06_{\hspace{1pt}0.06}$ & $82.82_{\hspace{1pt}0.17}$ & $83.46_{\hspace{1pt}0.25}$ & $83.56_{\hspace{1pt}0.32}$ & $83.53_{\hspace{1pt}0.37}$
    \\[1pt]
    & &
    $(15.5\%)$ & $(23.21\%)$ & $(28.26\%)$ & $(31.96\%)$ & $(34.97\%)$
    \\
    \cmidrule[1.5pt](l{\abovetopsep}r{\belowbottomsep}){1-7}

    \multirow{5}{*}{\begin{tabular}[c]{@{}c@{}}\textbf{MNLI Mismatched}\\Accuracy\end{tabular}} &
    \multirow{2}{*}{RoBERTa\textsubscript{\textrm{LARGE}}} & 
    $89.48_{\hspace{1pt}0.19}$ & $89.22_{\hspace{1pt}0.96}$ & $83.85_{\hspace{1pt}10.81}$ & $90.16_{\hspace{1pt}0.25}$ & $90.22_{\hspace{1pt}0.05}$
    \\[1pt]
    & &
    $(11.61\%)$ & $(17.62\%)$ & $(21.56\%)$ & $(24.61\%)$ & $(27.06\%)$
    \\[6pt]

    & \multirow{2}{*}{OPT\textsubscript{\textrm{350M}}} & 
    $82.90_{\hspace{1pt}0.20}$ & $83.72_{\hspace{1pt}0.14}$ & $84.14_{\hspace{1pt}0.11}$ & $84.39_{\hspace{1pt}0.30}$ & $84.29_{\hspace{1pt}0.05}$
    \\[1pt]
    & &
    $(15.5\%)$ & $(23.21\%)$ & $(28.26\%)$ & $(31.96\%)$ & $(34.97\%)$
    \\
    \cmidrule[1.5pt](l{\abovetopsep}r{\belowbottomsep}){1-7}

    \multirow{5}{*}{\begin{tabular}[c]{@{}c@{}}\textbf{SNLI}\\Accuracy\end{tabular}} &
    \multirow{2}{*}{RoBERTa\textsubscript{\textrm{LARGE}}} & 
    $52.40_{\hspace{1pt}34.01}$ & $92.33_{\hspace{1pt}0.03}$ & $92.33_{\hspace{1pt}0.07}$ & $92.46_{\hspace{1pt}0.06}$ & $92.50_{\hspace{1pt}0.14}$
    \\[1pt]
    & &
    $(9.23\%)$ & $(13.79\%)$ & $(16.91\%)$ & $(19.29\%)$ & $(21.26\%)$
    \\[6pt]

    & \multirow{2}{*}{OPT\textsubscript{\textrm{350M}}} & 
    $88.72_{\hspace{1pt}0.08}$ & $89.33_{\hspace{1pt}0.02}$ & $89.58_{\hspace{1pt}0.16}$ & $89.90_{\hspace{1pt}0.08}$ & $90.07_{\hspace{1pt}0.29}$
    \\[1pt]
    & &
    $(11.41\%)$ & $(17.14\%)$ & $(20.95\%)$ & $(23.83\%)$ & $(26.13\%)$
    \\
    \cmidrule[1.5pt](l{\abovetopsep}r{\belowbottomsep}){1-7}

    \multirow{5}{*}{\begin{tabular}[c]{@{}c@{}}\textbf{SST-2}\\Accuracy\end{tabular}} &
    \multirow{2}{*}{RoBERTa\textsubscript{\textrm{LARGE}}} & 
    $95.07_{\hspace{1pt}0.12}$ & $95.22_{\hspace{1pt}0.17}$ & $95.80_{\hspace{1pt}0.18}$ & $95.91_{\hspace{1pt}0.13}$ & $95.80_{\hspace{1pt}0.37}$
    \\[1pt]
    & &
    $(9.91\%)$ & $(14.7\%)$ & $(17.98\%)$ & $(20.32\%)$ & $(23.95\%)$
    \\[6pt]

    & \multirow{2}{*}{OPT\textsubscript{\textrm{350M}}} & 
    $92.66_{\hspace{1pt}0.41}$ & $93.16_{\hspace{1pt}0.29}$ & $92.85_{\hspace{1pt}0.95}$ & $92.97_{\hspace{1pt}0.59}$ & $94.04_{\hspace{1pt}0.12}$
    \\[1pt]
    & &
    $(9.02\%)$ & $(13.33\%)$ & $(16.43\%)$ & $(18.83\%)$ & $(20.82\%)$
    \\
    \cmidrule[1.5pt](l{\abovetopsep}r{\belowbottomsep}){1-7}

    \multirow{5}{*}{\begin{tabular}[c]{@{}c@{}}\textbf{RACE}\\Accuracy\end{tabular}} &
    \multirow{2}{*}{RoBERTa\textsubscript{\textrm{LARGE}}} & 
    $61.91_{\hspace{1pt}31.87}$ & $81.77_{\hspace{1pt}0.23}$ & $82.29_{\hspace{1pt}0.18}$ & $82.89_{\hspace{1pt}0.40}$ & $82.95_{\hspace{1pt}0.13}$
    \\[1pt]
    & &
    $(28.03\%)$ & $(41.12\%)$ & $(50.33\%)$ & $(56.97\%)$ & $(62.39\%)$
    \\[6pt]

    & \multirow{2}{*}{OPT\textsubscript{\textrm{350M}}} & 
    $76.16_{\hspace{1pt}0.38}$ & $78.55_{\hspace{1pt}0.38}$ & $79.36_{\hspace{1pt}0.11}$ & $79.35_{\hspace{1pt}0.33}$ & $79.68_{\hspace{1pt}0.09}$
    \\[1pt]
    & &
    $(9.49\%)$ & $(14.05\%)$ & $(17.31\%)$ & $(19.68\%)$ & $(21.62\%)$
    \\
    \cmidrule[1.5pt](l{\abovetopsep}r{\belowbottomsep}){1-7}

    \multirow{5}{*}{\begin{tabular}[c]{@{}c@{}}\textbf{SQuAD v2}\\Exact Match\end{tabular}} &
    \multirow{2}{*}{RoBERTa\textsubscript{\textrm{LARGE}}} & 
    $83.64_{\hspace{1pt}0.32}$ & $84.24_{\hspace{1pt}0.38}$ & $84.88_{\hspace{1pt}0.22}$ & $84.92_{\hspace{1pt}0.31}$ & $85.09_{\hspace{1pt}0.14}$
    \\[1pt]
    & &
    $(19.44\%)$ & $(29.15\%)$ & $(35.67\%)$ & $(43.24\%)$ & $(46.66\%)$
    \\[6pt]

    & \multirow{2}{*}{OPT\textsubscript{\textrm{350M}}} & 
    $62.15_{\hspace{1pt}0.17}$ & $62.97_{\hspace{1pt}0.15}$ & $63.23_{\hspace{1pt}0.24}$ & $63.61_{\hspace{1pt}0.18}$ & $63.72_{\hspace{1pt}0.38}$
    \\[1pt]
    & &
    $(20.57\%)$ & $(30.74\%)$ & $(37.12\%)$ & $(41.63\%)$ & $(45.13\%)$
    \\
    \cmidrule[1pt](l{\abovetopsep}r{\belowbottomsep}){1-7}

    \multirow{5}{*}{\begin{tabular}[c]{@{}c@{}}\textbf{SQuAD v2}\\F1 Score\end{tabular}} &
    \multirow{2}{*}{RoBERTa\textsubscript{\textrm{LARGE}}} & 
    $86.35_{\hspace{1pt}0.34}$ & $86.98_{\hspace{1pt}0.36}$ & $87.43_{\hspace{1pt}0.20}$ & $87.53_{\hspace{1pt}0.29}$ & $87.70_{\hspace{1pt}0.14}$
    \\[1pt]
    & &
    $(19.44\%)$ & $(29.15\%)$ & $(35.67\%)$ & $(43.24\%)$ & $(46.66\%)$
    \\[6pt]

    & \multirow{2}{*}{OPT\textsubscript{\textrm{350M}}} & 
    $64.80_{\hspace{1pt}0.17}$ & $65.61_{\hspace{1pt}0.18}$ & $65.80_{\hspace{1pt}0.23}$ & $66.25_{\hspace{1pt}0.22}$ & $66.49_{\hspace{1pt}0.22}$
    \\[1pt]
    & &
    $(20.57\%)$ & $(30.74\%)$ & $(37.12\%)$ & $(41.63\%)$ & $(45.13\%)$
    \\
    \cmidrule[1.5pt](l{\abovetopsep}r{\belowbottomsep}){1-7}
    
\end{tabular}

    \caption{
    Evaluation accuracy of fine-tuning RoBERTa\textsubscript{\textrm{LARGE}} and OPT\textsubscript{\textrm{350M}} on the winning ticket subset for different values of $S$.
    The size of each subset is provided in parenthesis as a percentage of the original dataset size.
    Each reported accuracy is the average of 3 runs with different intialization seeds.
    The standard deviation of the runs is also reported as the subscript.
    }
    \label{tab:ablation_s}
\end{table}

\begin{table}[H]
    \centering
    \begin{tabular}{cc|ccccccc}
     \multicolumn{9}{c}{\textbf{MNLI Matched} (Accuracy)} \\
     
    \cmidrule[1.5pt](l{\abovetopsep}r{\belowbottomsep}){1-9}
    
     & & $\mathcal{D}_{\{4\}}$ & $\mathcal{D}_{\{5\}}$ & $\mathcal{D}_{\{2,3,4\}}$ & $\mathcal{D}_{\{4,5\}}$ & $\mathcal{D}_{\{3,4,5\}}$ & $\mathcal{D}_{\{2,3,4,5\}}$ & $\mathcal{D}_{\{1,2,3,4,5\}}$
      \\
    \cmidrule[0.2pt](l{\tabcolsep}r{\tabcolsep}){1-9}
    
    \parbox[t]{5mm}{\multirow{7}{*}{\rotatebox[origin=c]{90}{\begin{tabular}[c]{@{}c@{}}RoBERTa\\Large\end{tabular}}}} & 
    \multirow{2}{*}{$E = 3$} & 
    $88.68_{\hspace{1pt}0.50}$ & $89.48_{\hspace{1pt}0.09}$ & $89.91_{\hspace{1pt}0.24}$ & $90.06_{\hspace{1pt}0.13}$ & $90.23_{\hspace{1pt}0.11}$ & $90.36_{\hspace{1pt}0.09}$ & $90.57_{\hspace{1pt}0.12}$
    \\[1pt]
    & &
    $(5.42\%)$ & $(11.58\%)$ & $(12.27\%)$ & $(17.0\%)$ & $(20.7\%)$ & $(23.85\%)$ & $(27.06\%)$
    \\[5pt]
    &
    \multirow{2}{*}{$E = 2$} & 
    $88.58_{\hspace{1pt}0.40}$ & $89.35_{\hspace{1pt}0.11}$ & $89.91_{\hspace{1pt}0.09}$ & $89.95_{\hspace{1pt}0.16}$ & $90.06_{\hspace{1pt}0.19}$ & $90.25_{\hspace{1pt}0.09}$ & $90.46_{\hspace{1pt}0.12}$
    \\[1pt]
    & &
    $(5.41\%)$ & $(11.64\%)$ & $(12.18\%)$ & $(17.04\%)$ & $(20.7\%)$ & $(23.81\%)$ & $(26.88\%)$
    \\[5pt]
    &
    \multirow{2}{*}{$E = 1$} & 
    $69.11_{\hspace{1pt}32.30}$ & $69.98_{\hspace{1pt}33.05}$ & $89.42_{\hspace{1pt}0.20}$ & $69.82_{\hspace{1pt}32.92}$ & $90.03_{\hspace{1pt}0.22}$ & $90.41_{\hspace{1pt}0.10}$ & $83.83_{\hspace{1pt}10.71}$
    \\[1pt]
    & &
    $(5.33\%)$ & $(11.8\%)$ & $(11.86\%)$ & $(17.18\%)$ & $(20.75\%)$ & $(23.66\%)$ & $(26.42\%)$
    \\

    \cmidrule[1pt](l{\abovetopsep}r{\belowbottomsep}){1-9}

    \parbox[t]{5mm}{\multirow{7}{*}{\rotatebox[origin=c]{90}{\begin{tabular}[c]{@{}c@{}}OPT\\350M\end{tabular}}}} & 
    \multirow{2}{*}{$E = 3$} & 
    $80.87_{\hspace{1pt}0.39}$ & $81.87_{\hspace{1pt}0.14}$ & $81.99_{\hspace{1pt}0.44}$ & $82.37_{\hspace{1pt}0.19}$ & $83.23_{\hspace{1pt}0.14}$ & $83.38_{\hspace{1pt}0.11}$ & $83.53_{\hspace{1pt}0.37}$
    \\[1pt]
    &&
    $(7.18\%)$ & $(13.14\%)$ & $(17.14\%)$ & $(20.32\%)$ & $(25.62\%)$ & $(30.28\%)$ & $(34.97\%)$
    \\[5pt]
    &
    \multirow{2}{*}{$E = 2$} & 
    $80.98_{\hspace{1pt}0.17}$ & $81.83_{\hspace{1pt}0.09}$ & $81.75_{\hspace{1pt}0.52}$ & $82.51_{\hspace{1pt}0.11}$ & $83.17_{\hspace{1pt}0.13}$ & $83.52_{\hspace{1pt}0.12}$ & $83.70_{\hspace{1pt}0.29}$
    \\[1pt]
    &&
    $(7.14\%)$ & $(13.13\%)$ & $(16.98\%)$ & $(20.27\%)$ & $(25.54\%)$ & $(30.11\%)$ & $(34.7\%)$
    \\[5pt]
    &
    \multirow{2}{*}{$E = 1$} & 
    $80.36_{\hspace{1pt}0.27}$ & $81.81_{\hspace{1pt}0.24}$ & $79.86_{\hspace{1pt}0.36}$ & $82.56_{\hspace{1pt}0.18}$ & $83.13_{\hspace{1pt}0.18}$ & $83.46_{\hspace{1pt}0.29}$ & $83.56_{\hspace{1pt}0.22}$
    \\[1pt]
    &&
    $(7.11\%)$ & $(13.28\%)$ & $(16.75\%)$ & $(20.39\%)$ & $(25.62\%)$ & $(30.03\%)$ & $(34.35\%)$
    \\
    
    \cmidrule[1.5pt](l{\abovetopsep}r{\belowbottomsep}){1-9}
     \multicolumn{9}{c}{\textbf{MNLI Mismatched} (Accuracy)} \\
    \cmidrule[1.5pt](l{\abovetopsep}r{\belowbottomsep}){1-9}

     & & $\mathcal{D}_{\{4\}}$ & $\mathcal{D}_{\{5\}}$ & $\mathcal{D}_{\{2,3,4\}}$ & $\mathcal{D}_{\{4,5\}}$ & $\mathcal{D}_{\{3,4,5\}}$ & $\mathcal{D}_{\{2,3,4,5\}}$ & $\mathcal{D}_{\{1,2,3,4,5\}}$
      \\
    \cmidrule[0.2pt](l{\tabcolsep}r{\tabcolsep}){1-9}
    
    \parbox[t]{5mm}{\multirow{7}{*}{\rotatebox[origin=c]{90}{\begin{tabular}[c]{@{}c@{}}RoBERTa\\Large\end{tabular}}}} & 
    \multirow{2}{*}{$E = 3$} & 
    $88.79_{\hspace{1pt}0.35}$ & $89.26_{\hspace{1pt}0.25}$ & $89.61_{\hspace{1pt}0.18}$ & $89.81_{\hspace{1pt}0.14}$ & $90.05_{\hspace{1pt}0.04}$ & $90.12_{\hspace{1pt}0.31}$ & $90.22_{\hspace{1pt}0.05}$
    \\[1pt]
    & &
    $(5.42\%)$ & $(11.58\%)$ & $(12.27\%)$ & $(17.0\%)$ & $(20.7\%)$ & $(23.85\%)$ & $(27.06\%)$
    \\[5pt]
    &
    \multirow{2}{*}{$E = 2$} & 
    $88.40_{\hspace{1pt}0.43}$ & $89.20_{\hspace{1pt}0.12}$ & $89.80_{\hspace{1pt}0.13}$ & $89.57_{\hspace{1pt}0.05}$ & $89.95_{\hspace{1pt}0.05}$ & $90.15_{\hspace{1pt}0.06}$ & $90.27_{\hspace{1pt}0.03}$
    \\[1pt]
    & &
    $(5.41\%)$ & $(11.64\%)$ & $(12.18\%)$ & $(17.04\%)$ & $(20.7\%)$ & $(23.81\%)$ & $(26.88\%)$
    \\[5pt]
    &
    \multirow{2}{*}{$E = 1$} & 
    $69.10_{\hspace{1pt}32.29}$ & $70.04_{\hspace{1pt}33.10}$ & $89.34_{\hspace{1pt}0.12}$ & $69.70_{\hspace{1pt}32.82}$ & $90.17_{\hspace{1pt}0.19}$ & $90.35_{\hspace{1pt}0.12}$ & $84.26_{\hspace{1pt}9.98}$
    \\[1pt]
    & &
    $(5.33\%)$ & $(11.8\%)$ & $(11.86\%)$ & $(17.18\%)$ & $(20.75\%)$ & $(23.66\%)$ & $(26.42\%)$
    \\

    \cmidrule[1pt](l{\abovetopsep}r{\belowbottomsep}){1-9}

    \parbox[t]{5mm}{\multirow{7}{*}{\rotatebox[origin=c]{90}{\begin{tabular}[c]{@{}c@{}}OPT\\350M\end{tabular}}}} & 
    \multirow{2}{*}{$E = 3$} & 
    $81.83_{\hspace{1pt}0.35}$ & $82.56_{\hspace{1pt}0.25}$ & $82.85_{\hspace{1pt}0.18}$ & $83.55_{\hspace{1pt}0.14}$ & $83.70_{\hspace{1pt}0.04}$ & $84.15_{\hspace{1pt}0.31}$ & $84.29_{\hspace{1pt}0.05}$
    \\[1pt]
    &&
    $(7.18\%)$ & $(13.14\%)$ & $(17.14\%)$ & $(20.32\%)$ & $(25.62\%)$ & $(30.28\%)$ & $(34.97\%)$
    \\[5pt]
    &
    \multirow{2}{*}{$E = 2$} & 
    $81.77_{\hspace{1pt}0.15}$ & $82.69_{\hspace{1pt}0.28}$ & $82.73_{\hspace{1pt}0.03}$ & $83.45_{\hspace{1pt}0.09}$ & $83.86_{\hspace{1pt}0.08}$ & $84.22_{\hspace{1pt}0.35}$ & $84.36_{\hspace{1pt}0.13}$
    \\[1pt]
    &&
    $(7.14\%)$ & $(13.13\%)$ & $(16.98\%)$ & $(20.27\%)$ & $(25.54\%)$ & $(30.11\%)$ & $(34.7\%)$
    \\[5pt]
    &
    \multirow{2}{*}{$E = 1$} & 
    $81.23_{\hspace{1pt}0.11}$ & $82.65_{\hspace{1pt}0.20}$ & $80.57_{\hspace{1pt}0.34}$ & $83.48_{\hspace{1pt}0.02}$ & $83.93_{\hspace{1pt}0.15}$ & $84.23_{\hspace{1pt}0.27}$ & $83.82_{\hspace{1pt}0.38}$
    \\[1pt]
    &&
    $(7.11\%)$ & $(13.28\%)$ & $(16.75\%)$ & $(20.39\%)$ & $(25.62\%)$ & $(30.03\%)$ & $(34.35\%)$
    \\

\end{tabular}

    \caption{
    Evaluation accuracy of fine-tuning RoBERTa\textsubscript{\textrm{LARGE}} and OPT\textsubscript{\textrm{350M}} on subsets acquired with different values of $E$.
    The size of each subset is provided in parenthesis as a percentage of the original dataset size.
    Each reported accuracy is the average of 3 runs with different intialization seeds.
    The standard deviation of the runs is also reported as the subscript.
    }
    \label{tab:ablation_e1}
\end{table}

\begin{table}[H]
    \centering
    \begin{tabular}{cc|ccccccc}
     \multicolumn{9}{c}{\textbf{SNLI} (Accuracy)} \\
     
    \cmidrule[1.5pt](l{\abovetopsep}r{\belowbottomsep}){1-9}
    
     & & $\mathcal{D}_{\{4\}}$ & $\mathcal{D}_{\{5\}}$ & $\mathcal{D}_{\{2,3,4\}}$ & $\mathcal{D}_{\{4,5\}}$ & $\mathcal{D}_{\{3,4,5\}}$ & $\mathcal{D}_{\{2,3,4,5\}}$ & $\mathcal{D}_{\{1,2,3,4,5\}}$
      \\
    \cmidrule[0.2pt](l{\tabcolsep}r{\tabcolsep}){1-9}
    
    \parbox[t]{5mm}{\multirow{7}{*}{\rotatebox[origin=c]{90}{\begin{tabular}[c]{@{}c@{}}RoBERTa\\Large\end{tabular}}}} & 
    \multirow{2}{*}{$E = 3$} & 
    $91.41_{\hspace{1pt}0.31}$ & $91.84_{\hspace{1pt}0.15}$ & $91.66_{\hspace{1pt}0.16}$ & $92.13_{\hspace{1pt}0.22}$ & $92.38_{\hspace{1pt}0.03}$ & $92.41_{\hspace{1pt}0.03}$ & $92.50_{\hspace{1pt}0.14}$
    \\[1pt]
    & &
    $(4.2\%)$ & $(9.27\%)$ & $(9.52\%)$ & $(13.47\%)$ & $(16.39\%)$ & $(18.8\%)$ & $(21.26\%)$
    \\[5pt]
    &
    \multirow{2}{*}{$E = 2$} & 
    $60.03_{\hspace{1pt}29.49}$ & $52.42_{\hspace{1pt}34.04}$ & $32.77_{\hspace{1pt}0.00}$ & $92.08_{\hspace{1pt}0.15}$ & $52.52_{\hspace{1pt}34.21}$ & $52.62_{\hspace{1pt}34.38}$ & $72.56_{\hspace{1pt}34.46}$
    \\[1pt]
    & &
    $(4.19\%)$ & $(9.38\%)$ & $(9.48\%)$ & $(13.58\%)$ & $(16.48\%)$ & $(18.87\%)$ & $(21.27\%)$
    \\[5pt]
    &
    \multirow{2}{*}{$E = 1$} & 
    $71.77_{\hspace{1pt}33.78}$ & $32.77_{\hspace{1pt}0.00}$ & $32.77_{\hspace{1pt}0.00}$ & $52.56_{\hspace{1pt}34.27}$ & $32.77_{\hspace{1pt}0.00}$ & $52.55_{\hspace{1pt}34.26}$ & $52.58_{\hspace{1pt}34.31}$
    \\[1pt]
    & &
    $(4.2\%)$ & $(9.29\%)$ & $(9.48\%)$ & $(13.68\%)$ & $(16.51\%)$ & $(18.76\%)$ & $(20.95\%)$
    \\

    \cmidrule[1pt](l{\abovetopsep}r{\belowbottomsep}){1-9}

    \parbox[t]{5mm}{\multirow{7}{*}{\rotatebox[origin=c]{90}{\begin{tabular}[c]{@{}c@{}}OPT\\350M\end{tabular}}}} & 
    \multirow{2}{*}{$E = 3$} & 
    $87.34_{\hspace{1pt}0.12}$ & $88.44_{\hspace{1pt}0.03}$ & $88.21_{\hspace{1pt}0.32}$ & $89.13_{\hspace{1pt}0.13}$ & $89.58_{\hspace{1pt}0.11}$ & $89.87_{\hspace{1pt}0.14}$ & $90.07_{\hspace{1pt}0.29}$
    \\[1pt]
    &&
    $(5.34\%)$ & $(10.71\%)$ & $(12.26\%)$ & $(16.06\%)$ & $(19.79\%)$ & $(22.97\%)$ & $(26.13\%)$
    \\[5pt]
    &
    \multirow{2}{*}{$E = 2$} & 
    $87.23_{\hspace{1pt}0.23}$ & $88.46_{\hspace{1pt}0.27}$ & $87.63_{\hspace{1pt}0.72}$ & $89.15_{\hspace{1pt}0.06}$ & $89.57_{\hspace{1pt}0.15}$ & $89.74_{\hspace{1pt}0.14}$ & $89.76_{\hspace{1pt}0.25}$
    \\[1pt]
    &&
    $(5.31\%)$ & $(10.75\%)$ & $(12.13\%)$ & $(16.05\%)$ & $(19.76\%)$ & $(22.88\%)$ & $(25.97\%)$
    \\[5pt]
    &
    \multirow{2}{*}{$E = 1$} & 
    $85.44_{\hspace{1pt}1.17}$ & $88.33_{\hspace{1pt}0.05}$ & $84.12_{\hspace{1pt}0.81}$ & $88.99_{\hspace{1pt}0.10}$ & $89.37_{\hspace{1pt}0.17}$ & $89.59_{\hspace{1pt}0.17}$ & $89.33_{\hspace{1pt}0.10}$
    \\[1pt]
    &&
    $(5.29\%)$ & $(10.88\%)$ & $(11.95\%)$ & $(16.17\%)$ & $(19.78\%)$ & $(22.83\%)$ & $(25.77\%)$
    \\
    
    \cmidrule[1.5pt](l{\abovetopsep}r{\belowbottomsep}){1-9}
     \multicolumn{9}{c}{\textbf{SST-2} (Accuracy)} \\
    \cmidrule[1.5pt](l{\abovetopsep}r{\belowbottomsep}){1-9}

     & & $\mathcal{D}_{\{4\}}$ & $\mathcal{D}_{\{2,3,4\}}$ & $\mathcal{D}_{\{5\}}$ & $\mathcal{D}_{\{4,5\}}$ & $\mathcal{D}_{\{3,4,5\}}$ & $\mathcal{D}_{\{2,3,4,5\}}$ & $\mathcal{D}_{\{1,2,3,4,5\}}$
      \\
    \cmidrule[0.2pt](l{\tabcolsep}r{\tabcolsep}){1-9}
    
    \parbox[t]{5mm}{\multirow{7}{*}{\rotatebox[origin=c]{90}{\begin{tabular}[c]{@{}c@{}}RoBERTa\\Large\end{tabular}}}} & 
    \multirow{2}{*}{$E = 3$} & 
    $94.15_{\hspace{1pt}0.35}$ & $94.65_{\hspace{1pt}0.46}$ & $94.99_{\hspace{1pt}0.17}$ & $95.45_{\hspace{1pt}0.24}$ & $95.53_{\hspace{1pt}0.42}$ & $95.87_{\hspace{1pt}0.23}$ & $95.80_{\hspace{1pt}0.37}$
    \\[1pt]
    & &
    $(4.6\%)$ & $(9.11\%)$ & $(13.21\%)$ & $(17.81\%)$ & $(20.41\%)$ & $(22.32\%)$ & $(23.95\%)$
    \\[5pt]
    &
    \multirow{2}{*}{$E = 2$} & 
    $63.45_{\hspace{1pt}24.90}$ & $49.04_{\hspace{1pt}0.06}$ & $64.45_{\hspace{1pt}26.62}$ & $95.10_{\hspace{1pt}0.27}$ & $80.24_{\hspace{1pt}26.98}$ & $95.49_{\hspace{1pt}0.17}$ & $95.87_{\hspace{1pt}0.46}$
    \\[1pt]
    & &
    $(4.65\%)$ & $(9.07\%)$ & $(13.3\%)$ & $(17.95\%)$ & $(20.53\%)$ & $(22.37\%)$ & $(23.91\%)$
    \\[5pt]
    &
    \multirow{2}{*}{$E = 1$} & 
    $62.08_{\hspace{1pt}22.31}$ & $49.08_{\hspace{1pt}0.00}$ & $95.26_{\hspace{1pt}0.24}$ & $80.43_{\hspace{1pt}27.15}$ & $79.89_{\hspace{1pt}26.68}$ & $79.13_{\hspace{1pt}26.08}$ & $94.04_{\hspace{1pt}3.49}$
    \\[1pt]
    & &
    $(4.44\%)$ & $(8.48\%)$ & $(13.49\%)$ & $(17.93\%)$ & $(20.4\%)$ & $(21.96\%)$ & $(23.25\%)$
    \\

    \cmidrule[1pt](l{\abovetopsep}r{\belowbottomsep}){1-9}

    \parbox[t]{5mm}{\multirow{7}{*}{\rotatebox[origin=c]{90}{\begin{tabular}[c]{@{}c@{}}OPT\\350M\end{tabular}}}} & 
    \multirow{2}{*}{$E = 3$} & 
    $91.67_{\hspace{1pt}0.33}$ & $90.98_{\hspace{1pt}0.35}$ & $93.19_{\hspace{1pt}0.18}$ & $93.81_{\hspace{1pt}0.12}$ & $93.73_{\hspace{1pt}0.07}$ & $94.19_{\hspace{1pt}0.29}$ & $94.04_{\hspace{1pt}0.12}$
    \\[1pt]
    &&
    $(4.37\%)$ & $(8.58\%)$ & $(10.71\%)$ & $(15.08\%)$ & $(17.57\%)$ & $(19.29\%)$ & $(20.82\%)$
    \\[5pt]
    &
    \multirow{2}{*}{$E = 2$} & 
    $87.27_{\hspace{1pt}5.76}$ & $88.91_{\hspace{1pt}3.15}$ & $92.55_{\hspace{1pt}0.31}$ & $93.27_{\hspace{1pt}0.48}$ & $93.04_{\hspace{1pt}0.78}$ & $93.50_{\hspace{1pt}0.57}$ & $93.20_{\hspace{1pt}0.27}$
    \\[1pt]
    &&
    $(4.4\%)$ & $(8.59\%)$ & $(10.75\%)$ & $(15.15\%)$ & $(17.62\%)$ & $(19.34\%)$ & $(20.85\%)$
    \\[5pt]
    &
    \multirow{2}{*}{$E = 1$} & 
    $90.48_{\hspace{1pt}1.43}$ & $72.59_{\hspace{1pt}13.93}$ & $93.08_{\hspace{1pt}0.29}$ & $93.08_{\hspace{1pt}0.35}$ & $93.85_{\hspace{1pt}0.24}$ & $93.27_{\hspace{1pt}0.24}$ & $92.85_{\hspace{1pt}0.35}$
    \\[1pt]
    &&
    $(4.41\%)$ & $(8.43\%)$ & $(10.75\%)$ & $(15.16\%)$ & $(17.56\%)$ & $(19.18\%)$ & $(20.5\%)$
    \\

\end{tabular}

    \caption{
    Evaluation accuracy of fine-tuning RoBERTa\textsubscript{\textrm{LARGE}} and OPT\textsubscript{\textrm{350M}} on subsets acquired with different values of $E$.
    The size of each subset is provided in parenthesis as a percentage of the original dataset size.
    Each reported accuracy is the average of 3 runs with different intialization seeds.
    The standard deviation of the runs is also reported as the subscript.
    }
    \label{tab:ablation_e2}
\end{table}

\end{document}